\pdfoutput=1
\PassOptionsToPackage{table}{xcolor}
\documentclass[10pt]{article}
\ifdefined\pdfminorversion\pdfminorversion=7\fi
\usepackage[letterpaper,textwidth=6.75in,textheight=9in,headheight=12pt,headsep=14pt,footskip=28pt,top=0.95in]{geometry}
\usepackage[utf8]{inputenc}
\usepackage[T1]{fontenc}
\usepackage{lmodern}
\usepackage{xcolor}
\usepackage{natbib}
\usepackage{booktabs}
\usepackage{amsfonts}
\usepackage{nicefrac}
\usepackage{microtype}
\usepackage{amsmath}
\usepackage{amsthm}
\usepackage{tcolorbox}
\tcbuselibrary{breakable}
\theoremstyle{definition}
\newtheorem{proposition}{Proposition}
\usepackage{bm}
\usepackage{graphicx}
\usepackage{multirow}
\usepackage{longtable}
\usepackage{wrapfig}
\usepackage{placeins}
\usepackage{ragged2e}
\usepackage{parskip}
\usepackage{titlesec}
\usepackage{caption}
\usepackage{fancyhdr}
\usepackage{marvosym}
\usepackage{hyperref}
\usepackage{url}
\usepackage{soul}
\sethlcolor{blue!25}
\newcommand{\redhl}[1]{\leavevmode{#1}}
\newcommand{\redhlon}{}
\definecolor{ceilingpink}{HTML}{FFD8D8}

\definecolor{bhacite}{HTML}{0071BC}
\definecolor{bhalink}{HTML}{9A4D92}
\definecolor{bhabg}{RGB}{247,247,247}
\hypersetup{colorlinks,linkcolor=bhalink,citecolor=bhacite,urlcolor=bhacite}
\def\bibfont{\small}
\titleformat{\section}{\large\bfseries\raggedright}{\thesection.}{0.5em}{}
\titleformat{\subsection}{\normalsize\bfseries\raggedright}{\thesubsection.}{0.5em}{}
\titleformat{\subsubsection}{\normalsize\bfseries\raggedright}{\thesubsubsection.}{0.5em}{}
\titleformat{\paragraph}[runin]{\normalsize\bfseries}{}{0pt}{}
\titlespacing*{\section}{0pt}{0.20in plus 0.04in minus 0.02in}{0.08in}
\titlespacing*{\subsection}{0pt}{0.15in plus 0.03in minus 0.02in}{0.06in}
\titlespacing*{\subsubsection}{0pt}{0.12in plus 0.02in minus 0.02in}{0.05in}
\titlespacing*{\paragraph}{0pt}{1.4ex plus 0.5ex minus 0.2ex}{0.8em}
\renewcommand{\headrulewidth}{1pt}
\fancypagestyle{bhafirst}{\fancyhf{}\rhead{\itshape\today}\renewcommand{\headrulewidth}{1pt}}
\newcommand{\fitwidth}{\ifdim\width>\linewidth \linewidth\else\width\fi}

\renewcommand{\floatpagefraction}{0.8}

\newcommand{\bhatitleblock}{%
\thispagestyle{bhafirst}
\vglue 0.12in
{\raggedright\fontsize{17.28}{21}\selectfont\bfseries Beyond Hierarchical Alignment: Distilling Teacher Diversity for Brain Visual Decoding\par}
\vskip 0.18in
{\raggedright\bfseries
\mbox{Dongyang Li$^{\,1,2,3,*}$}, \mbox{Peng Jiang$^{\,1}$}, \mbox{Jiayu Zuo$^{\,2,3}$}, \mbox{Chen Wei$^{\,2,3}$}, \mbox{Mei Wang$^{\,1}$},\\ \mbox{Liyuan Wang$^{\,1,\textrm{\Letter}}$}, \mbox{Xiaoxuan Jia$^{\,1,\textrm{\Letter}}$}, \mbox{Quanying Liu$^{\,2,3,\textrm{\Letter}}$}\par}
\vskip 0.06in
{\raggedright $^{1\,}$Tsinghua University \quad $^{2\,}$Omni-Intelligence \quad $^{3\,}$Southern University of Science and Technology\par}
\vskip 0.02in
{\raggedright\small $^{*\,}$Work done during an internship at Omni-Intelligence. \quad {\footnotesize\Letter}\,Corresponding authors.\par}
\begin{tcolorbox}[colback=bhabg,colframe=white,boxrule=0pt,arc=10pt,left=16pt,right=16pt,top=12pt,bottom=12pt,width=\textwidth,before skip=12pt,after skip=10pt]
\redhl{Brain decoding typically aligns neural recordings with features from a pretrained vision model. To enrich this target, prior work aligns brain signals to multiple layers of a model, mirroring the visual hierarchy, or fuses features from several models. Both, however, distill into a single target space, whose similarity geometry need not retain the complementary structure that each teacher captures.} We introduce \textbf{BHA (Beyond Hierarchical Alignment)} to learn shared brain--visual alignment while composing predictions across distinct visual spaces. BHA jointly learns the brain encoder and the fused visual target, allowing visual supervision to adapt to brain signals. \redhl{To retain each teacher’s structure,} teacher-specific \redhl{readouts} \redhl{additionally predict each teacher’s own space.} Learned weights compose their matching scores with the shared score, allowing complementary notions of visual similarity to contribute to retrieval. \redhl{The same brain representation also drives reconstruction through semantic and spatial adapters into a pretrained generator.} BHA improves within-subject retrieval on EEG and MEG, with gains persisting under reduced EEG training data. To examine what distinct visual spaces contribute, we analyze the learned representations and their matching behavior. Experts can assign opposing contributions to the same image regions, while learned model neurons exhibit semantic selectivity. These findings suggest that shared alignment can support distinct visual matching functions that can be composed for brain decoding.
\begin{center}
Project page: \url{https://ncclab-sustech.github.io/BHA-project-page/}
\end{center}
\end{tcolorbox}}

\begin{document}
\bhatitleblock

\begin{figure}[!h]
\centering
\includegraphics[width=\textwidth]{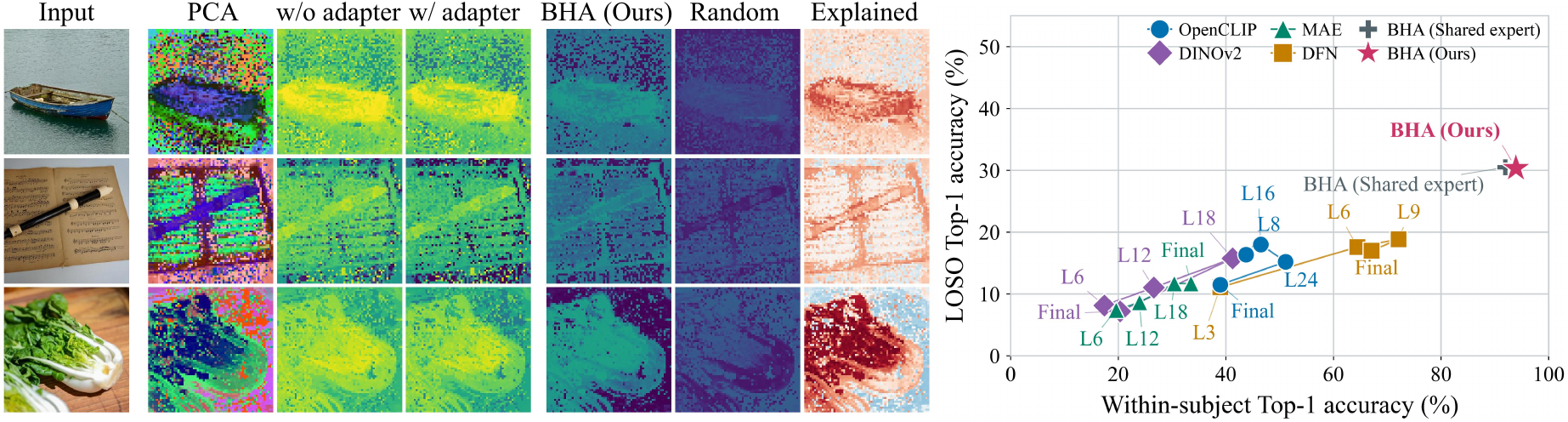}
\caption{\textbf{Visual representations shape brain--image alignment.}
\textbf{Left:} Input; visual-feature PCA; global-to-patch similarity before/after
adaptation; BHA-to-CLIP-patch similarity; random query (Random);
BHA-minus-random (Explained).
\textbf{Right:} CSLS-2 Top-1 on THINGS-EEG2 for VFM layers and BHA readouts.}
\label{fig:preview_imgs}
\end{figure}

\section{Introduction}
Decoding visual stimuli from non-invasive brain recordings requires learning from limited, noisy neural signals~\citep{simanova2010identifying,kaneshiro2015representational,horikawa2017generic,kavasidis2017brain2image,spampinato2017deep}. Pretrained vision foundation models (VFMs) provide visual targets that let brain encoders predict image representations rather than pixels. This is particularly useful for EEG and MEG, given their low signal-to-noise ratios and limited spatial resolution~\citep{benchetrit2024brain,song2024decoding,bai2024dreamdiffusion,li2024visual,fu2025brainvis}. Different models and network depths emphasize different stimulus properties, motivating complementary visual supervision~\citep{kar2024brave,tang2026aligning,du2026deep}. Figure~\ref{fig:preview_imgs} illustrates how teacher choice and adaptation affect retrieval and spatial matching.

\begin{figure}[!t]
    \centering
    \includegraphics[width=0.78\textwidth]{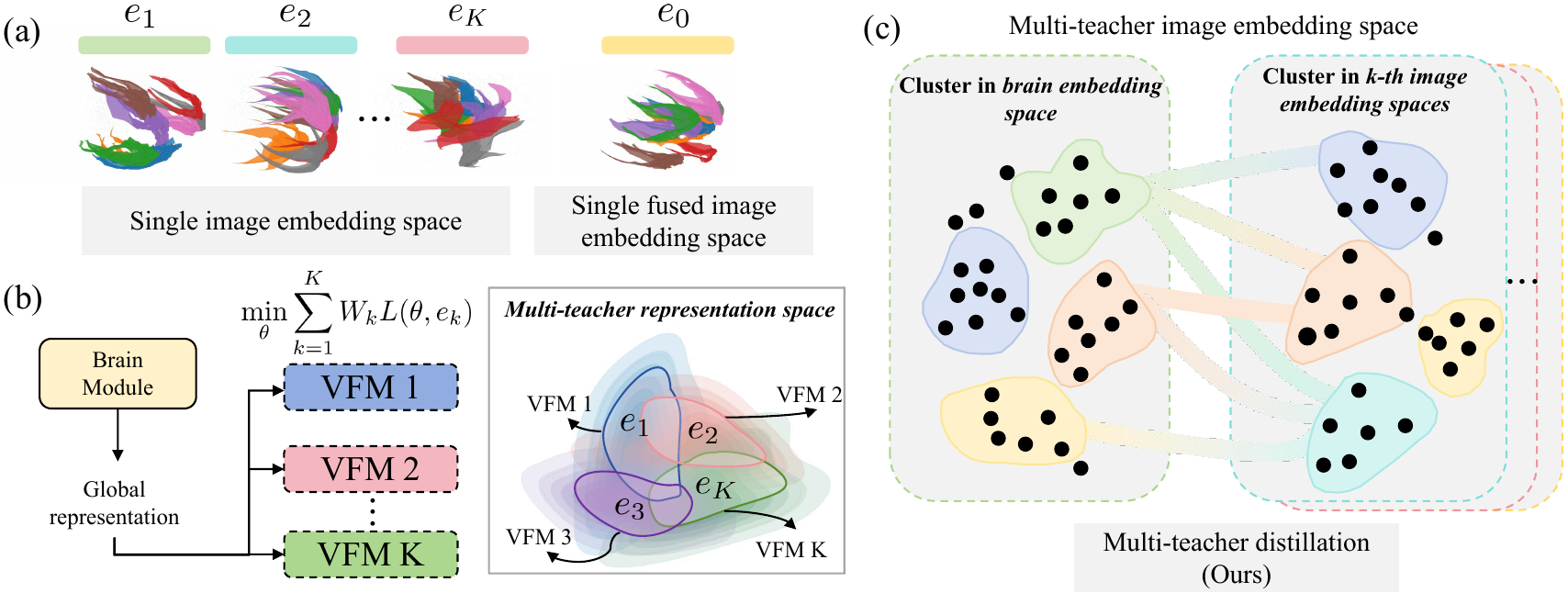}
\caption{\textbf{Shared alignment need not erase teacher-specific structure.}
(a) Different VFMs define distinct visual geometries; fusion combines them
into a single target space. (b) Multi-teacher supervision allows one brain
representation to support predictions in several visual spaces.
(c) Teacher-specific readouts retain distinct brain--image correspondences,
motivating score composition alongside shared alignment.}
    \label{fig:conceptualIzation}
\end{figure}

Recent work shows that the choice of visual supervision matters:
intermediate features benefit downstream multimodal and spatial tasks
\citep{bolya2025perception} and brain-to-image retrieval \citep{du2026deep},
while fusing global CLIP embeddings with VAE latents provides a shared
alignment target spanning semantics and structure \citep{zheng2026learning}.
These findings motivate a question beyond feature selection or fusion:
\emph{which differences between visual teachers are predictable from brain
activity and support complementary forms of image matching?}
We hypothesize that teacher diversity benefits decoding when its distinct
information is both learnable from neural signals and useful beyond a shared
alignment target.

Retrieval and reconstruction place different demands on visual supervision.
Adapting features toward brain-predictable information can improve matching
but alter the structure a frozen generator expects. This motivates
sharing a brain representation while predicting features in each task's space.

\textbf{Our approach.} We introduce \textbf{BHA (Beyond Hierarchical Alignment)}
to combine shared brain--visual alignment with task-specific prediction
spaces (Figure~\ref{fig:conceptualIzation}). BHA jointly learns a brain
encoder and a fused visual target. Teacher-specific \redhl{readouts (a Group MoE~\citep{shazeer2017outrageously} by default)}
then map the frozen brain representation into distinct visual spaces;
\redhl{learned weights combine} their matching scores with the shared
score for within-subject retrieval. Separately trained semantic and spatial
adapters predict conditions in the generator's native feature spaces,
enabling reconstruction with a frozen generator.
Across teacher families and depths, we examine which visual
distinctions remain brain-predictable and how they appear
in matching behavior and semantic selectivity.

Our main contributions are:
\begin{itemize}
  \item We formalize how complementary teacher supervision can reduce alignment estimation error, while adapting visual targets can change their compatibility with a frozen generator.
  \item We introduce BHA, which jointly learns the brain encoder and the visual fusion head, and unifies retrieval and reconstruction through a shared backbone with task-specific readouts. \redhl{Composing teacher-specific scores with the shared score substantially improves within-subject retrieval, with every readout tested and with limited training data.}
  \item Our empirical analyses reveal visual and semantic selectivity in learned brain representations, with distinct concept associations and spatial contributions to image matching across experts.
\end{itemize}

\begin{figure}[!t]
    \centering
    \includegraphics[width=1\textwidth]{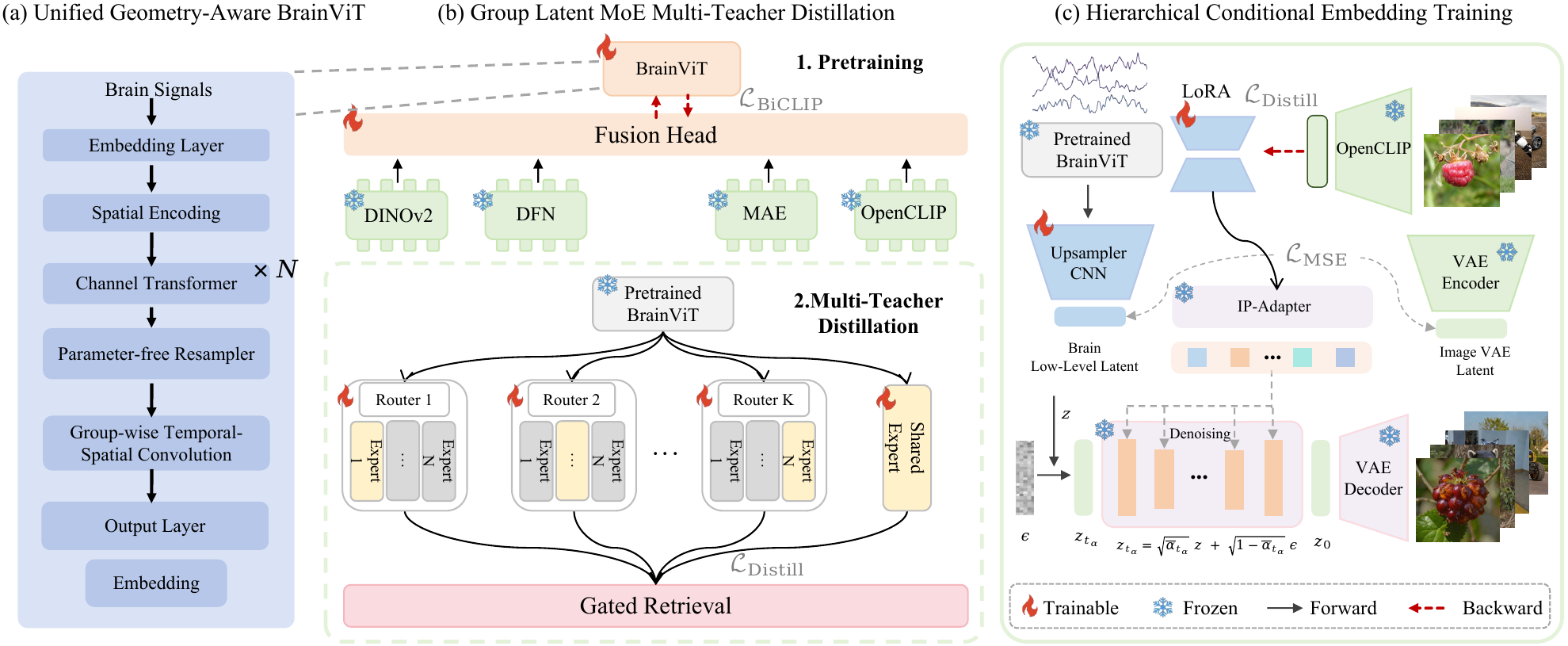}
\caption{\textbf{BHA learns shared representations and task-specific predictions.}
(a) BrainViT combines channel-wise attention and feature--sensor convolutions.
(b) \redhl{Two stage training.} Stage~1 jointly learns BrainViT and visual fusion; Stage~2 freezes both
and trains shared and teacher-specific readouts, visual adapters, and \redhl{composition weights}.
(c) Semantic and spatial predictors condition the frozen diffusion generator,
VAE, and IP-Adapter~\citep{ye2023ip}; see Appendix~\ref{sec:frozen_generator_inference}.}
    \label{fig:framework}
\end{figure}

\section{Related Work}
\paragraph{Representation Mixture for Visual Understanding}
VFM complementarity benefits visual understanding across layers and models.
MMFuser~\citep{cao2026mmfuser} queries shallow and intermediate details with deep features.
AM-RADIO~\citep{ranzinger2024amradio} and RADIOv2.5~\citep{heinrich2025radiov25}
distill VFMs; Swiss Army Knife~\citep{lu2025swissarmyknife} mixes teacher-specific
features \redhl{into one student}; and Theia~\citep{shang2025theia} and SigLino~\citep{chaybouti2026siglino}
compress heterogeneous teachers into compact students. Multi-encoder VLMs also combine visual features:
BRAVE~\citep{kar2024brave} compresses features from five encoders,
Cambrian-1~\citep{tong2024cambrian} aggregates spatial features, and
Eagle~\citep{shi2025eagle} finds simple concatenation competitive with complex mixers.
MoME~\citep{shen2024mome}, MoVA~\citep{zong2024mova}, and
PhotoEye~\citep{qi2025photoeye} condition expert fusion on tasks or instructions.
Together, these studies report downstream gains from VFM mixtures.
BHA transfers this principle to neural decoding, mixing models and depths
for retrieval while retaining generator-native targets.

\paragraph{Brain Visual Decoding and Reconstruction}
Brain-to-image reconstruction combines neural feature prediction with generative priors~\citep{lin2022mindreaderreconstructingcomplex,chen2023seeing,takagi2023high,zeng2024controllable,bai2024dreamdiffusion,scotti2024mindeye2}, including methods that keep generators frozen~\citep{Ozcelik_2023,scotti2024reconstructing,li2024visual}. Projection modules connect predicted embeddings to generation~\citep{li2024visual,dai2026autoregressive,zheng2026learning}, and fused features can require a separately trained conditioning prior~\citep{zheng2026learning}. ENIGMA~\citep{kneeland2026enigma} also predicts \redhl{a single CLIP embedding} for the IP-Adapter of SDXL-Turbo~\citep{podell2024sdxl,sauer2024adversarial}. BHA uses a multi-teacher-trained brain backbone with separate semantic and spatial readouts, and can refine noisy latents toward the predicted VAE condition at inference.

\paragraph{Hierarchical Brain--Vision Modeling}
Intermediate visual features can outperform final-layer representations~\citep{bolya2025perception}. Neural decoding studies examine feature granularity~\citep{du2026deep}, visibility-guided layer fusion~\citep{tang2026aligning}, and modality-specific routing~\citep{li2025brainflora}. In fMRI, VINDEX~\citep{xia2025exploring} compares feature organizations, while MindHier~\citep{zhang2026mindhier} aligns brain and visual hierarchies for reconstruction. MBP~\citep{zheng2026learning} \redhl{fuses several teachers into one target and} trains a Fusion Prior, including IP-Adapter, while freezing the diffusion backbone. \redhl{BHA instead keeps a brain-predicted space for each teacher alongside the fused one and composes their matching scores for retrieval, with each teacher at the depth favored by DMSA~\citep{du2026deep}. For reconstruction, it conditions the frozen generator in native CLIP and VAE spaces without a fusion prior.}

\section{Problem Setup and Preliminaries}
\label{sec:preliminaries}

Given $B$ brain--image pairs $\{(\mathbf{X}_i,\mathbf{I}_i)\}_{i=1}^{B}$,
a brain encoder and readout predict
$\hat{\mathbf{e}}_i=H(f_\theta(\mathbf{X}_i,s_i))\in\mathbb{R}^{d}$,
where $s_i$ optionally identifies the subject. A frozen visual encoder
provides targets $\mathbf{e}_i=\phi(\mathbf{I}_i)\in\mathbb{R}^{d}$.
For nonzero feature vectors, let $\operatorname{norm}(\mathbf{x})=\mathbf{x}/\|\mathbf{x}\|_2$,
$\operatorname{sim}(\mathbf{x},\mathbf{y})=
\operatorname{norm}(\mathbf{x})^\top\operatorname{norm}(\mathbf{y})$, and
$\ell_{ij}=\operatorname{sim}(\hat{\mathbf{e}}_i,\mathbf{e}_j)/\tau$
with temperature $\tau>0$. Feature regression and bidirectional contrastive
alignment (BiCLIP)~\citep{radford2021learning} use
\begin{align}
    \mathcal{L}_{\mathrm{MSE}}
    &=\frac{1}{Bd}\sum\nolimits_{i=1}^{B}
    \|\hat{\mathbf{e}}_i-\mathbf{e}_i\|_2^2,
    \label{eq:alignment_mse}\\
    \mathcal{L}_{\mathrm{BiCLIP}}(\boldsymbol{\ell})
    &=-\frac{1}{2B}\sum\nolimits_{i=1}^{B}
    \left[\log\frac{e^{\ell_{ii}}}{\sum_j e^{\ell_{ij}}}
    +\log\frac{e^{\ell_{ii}}}{\sum_j e^{\ell_{ji}}}\right].
    \label{eq:general_contrastive}
\end{align}
BiCLIP averages the brain-to-image and image-to-brain contrastive losses.
Retrieval ranks images by embedding similarity. Reconstruction predicts
semantic conditions and spatial latents in the pretrained generator's
native feature spaces. We use these losses within the corresponding
prediction spaces, with task-specific weights.

\section{Joint Brain--Visual Adaptation and Multi-Teacher Distillation}
\label{sec:methods}

\subsection{Complementary Supervision, Distinct Prediction Spaces}
\label{sec:bha_overview}

\redhl{BHA rests on two observations that a single fused target cannot
address at once: visual teachers provide complementary supervision for brain
alignment, and a target adapted for alignment need not suit a frozen
generator. We formalize each with a simple model; the first shows how
complementary teachers reduce estimation error without removing irreducible
uncertainty.}

\begin{tcolorbox}[breakable,colback=gray!8,colframe=gray!8,boxrule=0pt,
arc=0pt,left=7pt,right=7pt,top=3pt,bottom=3pt,before skip=8pt,after skip=8pt]
\begin{proposition}[Alignment risk under complementary teacher supervision]
\label{prop:multiteacher_alignment_bound}
Let $\mathbf{b}\in\mathbb{R}^{p}$ be brain features with
$\mathbb{E}[\mathbf{b}\mathbf{b}^{\top}]=I_p$, with visual target
$\mathbf{y}=\mathbf{M}_{\star}\mathbf{b}+\boldsymbol{\xi}\in\mathbb{R}^{d_0}$
and common-space teacher features
$\mathbf{t}_k=\mathbf{y}+\boldsymbol{\delta}_k$, $k=1,\ldots,K$.
The residual $\boldsymbol{\xi}$ and joint teacher-error vector have zero
means, finite second moments, and are independent of $\mathbf{b}$ and of
each other. Suppose
$\mathbb{E}\|\boldsymbol{\xi}\|_2^2/d_0=\sigma_{\mathrm{br}}^2$,
$\mathbb{E}\|\boldsymbol{\delta}_k\|_2^2/d_0=\sigma_{\mathrm{T}}^2>0$, and
$\mathbb{E}[\boldsymbol{\delta}_k^{\top}\boldsymbol{\delta}_{\ell}]/d_0
\leq\rho_{\mathrm{T}}\sigma_{\mathrm{T}}^2$ for $k\neq\ell$, where
$0\leq\rho_{\mathrm{T}}<1$.
Fit a linear predictor $\widehat g_K$ by least squares to
$K^{-1}\sum_k\mathbf{t}_k$ on $n$ i.i.d. samples with design matrix
$\mathsf{B}$ satisfying $\mathsf{B}^{\top}\mathsf{B}\succeq n\kappa I_p$,
$\kappa>0$. Define test risk
$\mathcal{R}_{\mathrm{vis}}(g)=\mathbb{E}\|g(\mathbf{b})-\mathbf{y}\|_2^2/d_0$
and $\chi_n=\operatorname{tr}[(\mathsf{B}^{\top}\mathsf{B})^{-1}]
\leq p/(n\kappa)$. Then
\begin{equation}
\sigma_{\mathrm{br}}^2
\leq\mathbb{E}_{\mathrm{train}}[\mathcal{R}_{\mathrm{vis}}(\widehat g_K)
\mid\mathsf{B}]
\leq\sigma_{\mathrm{br}}^2+
\chi_n\!\left[\sigma_{\mathrm{br}}^2+
\sigma_{\mathrm{T}}^2\!\left(\rho_{\mathrm{T}}+
\frac{1-\rho_{\mathrm{T}}}{K}\right)\right].
\label{eq:multiteacher_main_bound}
\end{equation}
For $K>1$, this upper bound is strictly below the expected risk of any
single-teacher least-squares predictor conditional on the same design.
With uncorrelated teacher errors, the teacher contribution decreases by a factor of $K$.
\end{proposition}
\end{tcolorbox}

\redhl{Proposition~\ref{prop:multiteacher_alignment_bound} motivates the
fused target of Stage~1: averaging teachers with weakly correlated errors
lowers estimation risk and, under a target-norm condition, tightens the
angular-error bound (\mbox{Appendix~\ref{sec:proof_multiteacher_alignment_bound}}).
This linear result motivates, but does not guarantee, gains from learned
nonlinear fusion. It also treats teachers as noisy views of one target,
whereas real teachers differ in what they encode
(\mbox{Sections~\ref{sec:prior_case_study}\textendash\ref{sec:teacher_geometry}});
a single fused geometry can attenuate these distinctions, so BHA additionally
keeps teacher-specific readouts and composes their scores with the shared one
(\mbox{Section~\ref{sec:gated_retrieval}}). Lower angular error, however,
concerns the adapted target, not the native features expected by a frozen
generator.}

\begin{tcolorbox}[breakable,colback=gray!8,colframe=gray!8,boxrule=0pt,
arc=0pt,left=7pt,right=7pt,top=3pt,bottom=3pt,before skip=8pt,after skip=8pt]
\begin{proposition}[Alignment to adapted targets and reconstruction discrepancy]
\label{prop:alignment_reconstruction_gap}
Let $\mathbf{e}$ be a native visual feature, $\mathbf{t}=V(\mathbf{e})$
an adapted target, and $\mathbf{h}=f(\mathbf{X})$ a brain prediction, with
$\|\mathbf{h}\|_2=\|\mathbf{t}\|_2=1$ almost surely. For sampling
noise, let $G$ be a frozen $L_G$-Lipschitz decoder on the feature domain. Define angular error
$\delta=\mathbb{E}[1-\mathbf{h}^{\top}\mathbf{t}]$, adaptation-induced
output drift $\Delta_G=\mathbb{E}\|G(\mathbf{t})-G(\mathbf{e})\|_2^2$,
and native-output discrepancy
$\mathcal{R}_G=\mathbb{E}\|G(\mathbf{h})-G(\mathbf{e})\|_2^2$.
Assuming finite second moments,
\begin{equation}
\left|\sqrt{\mathcal{R}_G}-\sqrt{\Delta_G}\right|
\leq L_G\sqrt{2\delta}.
\label{eq:alignment_native_reconstruction_bound}
\end{equation}
\end{proposition}
\end{tcolorbox}

\redhl{Together, the two results separate alignment from reconstruction:
Proposition~\ref{prop:multiteacher_alignment_bound} reduces $\delta$, yet by
Proposition~\ref{prop:alignment_reconstruction_gap}, as $\delta\to0$ the
native-output discrepancy approaches the adaptation drift $\Delta_G$ rather
than zero; even perfect alignment leaves $\mathcal{R}_G=\Delta_G$
(\mbox{Appendix~\ref{sec:proof_alignment_reconstruction_gap}}). Better
alignment to an adapted target therefore cannot by itself recover
native-conditioned reconstruction, so BHA predicts semantic and spatial
conditions directly in the generator's native spaces
(\mbox{Section~\ref{sec:hierarchical_reconstruction}}). Teacher-specific
readouts and these predictors all draw on one shared brain backbone
(\mbox{Figure~\ref{fig:framework}}).}

To preserve this common representation during specialization, we learn it
before fitting the task-specific predictors. The brain encoder and visual
fusion first adapt jointly using frozen VFM features
$\mathbf{e}_{ik}=\phi_k(\mathbf{I}_i)\in\mathbb{R}^{d_k}$.
We then freeze the backbone and fusion, allowing the prediction branches
to specialize without their objectives reshaping the shared representation.

\subsection{Two-Stage Training}
\label{sec:group_latent_moe}

Retrieval follows this separation in two stages: Stage~1 learns shared
alignment; Stage~2 learns how shared and teacher-specific predictions
contribute to matching. Reconstruction predictors are trained separately
on the pretrained backbone (Section~\ref{sec:hierarchical_reconstruction}).

\paragraph{Stage 1: Shared representation learning.}
\label{sec:fused_pretraining}
A learned fusion combines complementary visual features into a target
that adapts with the brain encoder. Let
$\mathbf{h}_i=f_\theta(\mathbf{X}_i,s_i)\in\mathbb{R}^{d_0}$.
Projections $W_k\in\mathbb{R}^{d_0\times d_k}$ map teacher features to a common
space, and $F:\mathbb{R}^{d_0}\to\mathbb{R}^{d_0}$ forms the fused target:
\begin{equation}
    \mathbf{e}_{i0}=\operatorname{norm}\!\left[
    F\!\left(\sum\nolimits_{k=1}^{K}
    \operatorname{norm}(W_k\mathbf{e}_{ik})\right)\right],
    \qquad
    \mathcal{L}_{\mathrm{pre}}=\mathcal{L}_{\mathrm{BiCLIP}}(\boldsymbol{\ell}_{\mathrm{pre}}).
    \label{eq:fused_visual_anchor}
\end{equation}
Here $\ell_{\mathrm{pre},ij}=\operatorname{sim}(\mathbf{h}_i,\mathbf{e}_{j0})/\tau_{\mathrm{pre}}$
with temperature $\tau_{\mathrm{pre}}>0$. Joint brain--visual adaptation
couples encoder learning with target learning: the brain encoder predicts
visual features, while fusion adapts the supervisory target to neural signals.

\paragraph{Stage 2: Teacher-specific prediction.}
\label{sec:multi_teacher_moe}
\label{sec:group_moe_training}
Shared brain features can support predictions in multiple visual spaces.
With BrainViT and fusion frozen, teacher-specific \redhl{readouts} learn separate
mappings into their teachers' feature spaces, allowing visual distinctions
recoverable from the backbone to contribute to matching.
A retrieval adapter yields $\mathbf{h}'_i=A(\operatorname{sg}(\mathbf{h}_i))\in\mathbb{R}^{d_0}$,
where $\operatorname{sg}$ denotes stop-gradient. \redhl{By default, each readout is a Group MoE (dense readouts perform the same; Table~\ref{table:moe_readout_ablation}):} group $k=1,\ldots,K$
combines a base projection $P_k\in\mathbb{R}^{d_k\times d_0}$, an always-active
expert $U_k$, and $R$ input-dependent residual experts, each mapping
$\mathbb{R}^{d_0}$ to $\mathbb{R}^{d_k}$:
\begin{equation}
    \hat{\mathbf{e}}_{ik}
    =P_k\mathbf{h}'_i+U_k(\mathbf{h}'_i)
    +\sum\nolimits_{r=1}^{R}w_{ikr}E_{kr}(\mathbf{h}'_i)
    \in\mathbb{R}^{d_k}.
    \label{eq:group_moe}
\end{equation}
Here $w_{ikr}\geq0$ and $\sum_r w_{ikr}=1$; the shared readout $E_0$ predicts the
fused target as $\hat{\mathbf{e}}_{i0}=E_0(\mathbf{h}'_i)$.
Groups retain their teachers' feature dimensions $d_k$, accommodating
heterogeneous visual spaces. Trainable visual adapters
$\tilde{\mathbf{e}}_{ik}=V_k(\operatorname{sg}(\mathbf{e}_{ik}))\in\mathbb{R}^{d_k}$
adjust teacher-specific targets without updating the source VFMs;
$\tilde{\mathbf{e}}_{i0}=\operatorname{sg}(\mathbf{e}_{i0})$ remains frozen.
\redhl{Each readout $k=0,\ldots,K$ is aligned on logits
$\ell_{ij}^{k}=s_{ij}^{k}/\tau_k$, where
$s_{ij}^{k}=\operatorname{sim}(\hat{\mathbf{e}}_{ik},\tilde{\mathbf{e}}_{jk})$
and $\tau_k>0$.}
Readouts, visual adapters, and \redhl{composition weights} are jointly trained with
\begin{equation}
    \redhl{\mathcal{L}_{\mathrm{distill}}
    =\sum\nolimits_{k=0}^{K}\lambda_k
      \mathcal{L}_{\mathrm{BiCLIP}}(\boldsymbol{\ell}^{k})
    +\lambda_{\mathrm{ret}}\mathcal{L}_{\mathrm{ret}}
    +\lambda_{\mathrm{aux}}\mathcal{L}_{\mathrm{aux}}
    +\lambda_{\mathrm{drift}}\mathcal{L}_{\mathrm{drift}}.}
    \label{eq:multi_teacher_distillation}
\end{equation}
The objective combines per-space alignment and retrieval-score learning
($\mathcal{L}_{\mathrm{ret}}$), with regularization for expert usage
($\mathcal{L}_{\mathrm{aux}}$) and visual drift ($\mathcal{L}_{\mathrm{drift}}$).
Appendix~\ref{sec:detail_group_latent_moe} details these objectives.

\paragraph{Gated score composition.}
\label{sec:gated_retrieval}
\redhl{As motivated in \mbox{Section~\ref{sec:bha_overview}}, we compose
matching scores to retain teacher-specific distinctions alongside shared
alignment.} For brain query $i$ and image $j$, the score is
\begin{equation}
    S_{ij}=s_{ij}^{0}+g_i\sum\nolimits_{k=1}^{K}\pi_{ik}s_{ij}^{k}.
    \label{eq:gated_retrieval_score}
\end{equation}
The shared score $s_{ij}^{0}$ supplies the common matching component.
Query-dependent weights $\pi_{ik}\geq0$, $\sum_k\pi_{ik}=1$, form a convex
combination of teacher-specific scores, while the gate $g_i\in[0,1]$
sets its overall contribution. Learning both separates the relative weighting
of teachers from the strength of their combined contribution to shared matching.
\redhl{After training, fixing $g_i$ and $\pi_{ik}$ to their averages gives the same accuracy (Table~\ref{table:moe_readout_ablation}).}

\subsection{Reconstruction without Generator Fine-Tuning}
\label{sec:hierarchical_reconstruction}

Retrieval alignment alone does not ensure reconstruction fidelity
(Proposition~\ref{prop:alignment_reconstruction_gap}). We therefore train
task-specific semantic and spatial predictors in the generator's native
feature spaces, using the pretrained BrainViT backbone with frozen base
weights and bypassing retrieval readouts.
The semantic prediction $\hat{\mathbf{e}}^{\mathrm{sem}}$ provides the
condition $\mathbf{c}_{\mathrm{sem}}$ for denoising, while the predicted
VAE latent $\hat{\mathbf{z}}$ supplies spatial initialization. Both paths
condition a pretrained latent diffusion generator~\citep{rombach2022high}
whose weights and conditioning interface remain frozen.

\paragraph{Generation from semantic and spatial predictions.}
A mixing parameter $\alpha\in[0,1]$ controls the transition from decoding
predicted spatial structure to generation under semantic conditioning.
For a schedule $t_0>\cdots>t_{N_{\mathrm{diff}}-1}$, let
$k_{\alpha}=\lfloor\alpha N_{\mathrm{diff}}\rfloor$ be the number of denoising
updates. We initialize the latent as
\begin{equation}
    \mathbf{z}^{\mathrm{init}}(\alpha)=
    \begin{cases}
        a\hat{\mathbf{z}}, & k_{\alpha}=0,\\
        a\hat{\mathbf{z}}+\sigma_{t_{N_{\mathrm{diff}}-k_{\alpha}}}\boldsymbol{\epsilon},
        & 0<k_{\alpha}<N_{\mathrm{diff}},\\
        \sigma_{\mathrm{init}}\boldsymbol{\epsilon}, & k_{\alpha}=N_{\mathrm{diff}},
    \end{cases}
    \label{eq:reconstruction_mixing_initialization}
\end{equation}
where $a$ is the VAE scaling factor, $\boldsymbol{\epsilon}$ is standard
Gaussian noise, and $\sigma_t$ and $\sigma_{\mathrm{init}}$ are scheduler noise
scales. Let $\mathcal{S}_{\alpha}$ apply the final $k_{\alpha}$ denoising
updates under $\mathbf{c}_{\mathrm{sem}}$, acting as the identity when
$k_{\alpha}=0$. \redhl{The frozen VAE decoder $D_{\mathrm{VAE}}$ yields the reconstruction
$\hat{\mathbf{I}}=D_{\mathrm{VAE}}\bigl(\mathcal{S}_{\alpha}(\mathbf{z}^{\mathrm{init}}(\alpha);\mathbf{c}_{\mathrm{sem}})/a\bigr)$.}
At $\alpha=0$, generation reduces
to decoding the spatial prediction; at $\alpha=1$, it starts from noise.
Intermediate values combine spatial initialization with denoising under
semantic conditioning, with larger $\alpha$ allowing greater departure
from the predicted structure.

\paragraph{Training-free guidance.}
Optional guidance refines the noisy latent to bring its estimated clean
latent closer to the brain-predicted VAE latent, maintaining a spatial
constraint beyond initialization. It uses no target images or
weight updates. Only the brain-to-condition predictors require training;
objectives are in Appendix~\ref{sec:detail_hierarchical_reconstruction};
conditioning interfaces and sampling settings in
Appendix~\ref{sec:reconstruction_implementation}.

\begin{table}[!htbp]
  \caption{\textbf{200-way brain-to-image retrieval (Top-1 / Top-5, \%).}
  Means over ten EEG and four MEG subjects. BHA uses Stage~2 gated scores
  within subject and Stage~1 outputs under LOSO. CSLS uses all 200 queries
  as unlabeled references; ``--'': unreported.}
  \label{table:retrieval_comparison}
  \centering
  \small
  \setlength{\tabcolsep}{8pt}
  \resizebox{\fitwidth}{!}{%
  \begin{tabular}{lcccccccc}
  \toprule
  \multirow{2}{*}{\textbf{Method}}
  & \multicolumn{4}{c}{\textbf{EEG}}
  & \multicolumn{4}{c}{\textbf{MEG}} \\
  \cmidrule(lr){2-5} \cmidrule(lr){6-9}
  & \multicolumn{2}{c}{\textbf{Within-subject}}
  & \multicolumn{2}{c}{\textbf{LOSO}}
  & \multicolumn{2}{c}{\textbf{Within-subject}}
  & \multicolumn{2}{c}{\textbf{LOSO}} \\
  \cmidrule(lr){2-3} \cmidrule(lr){4-5} \cmidrule(lr){6-7} \cmidrule(lr){8-9}
  & Top-1 & Top-5 & Top-1 & Top-5 & Top-1 & Top-5 & Top-1 & Top-5 \\
  \midrule
  NICE~\citep{song2024decoding} & 16.1 & 43.6 & 6.2 & 21.4 & 12.8 & 36.0 & -- & -- \\
  ATM-S~\citep{li2024visual} & 28.5 & 60.4 & 11.9 & 33.8 & -- & -- & -- & -- \\
  CC-All~\citep{zhang2024cognitioncapturer} & 35.6 & 80.2 & -- & -- & -- & -- & -- & -- \\
  UBP~\citep{wu2025bridging} & 50.9 & 79.7 & 12.4 & 33.4 & 26.7 & 55.2 & 2.2 & 10.4 \\
  HyFI~\citep{jo2026hyfi} & 68.2 & 91.9 & 15.1 & 37.2 & 35.8 & 64.6 & 3.2 & 11.5 \\
  NeuroBridge~\citep{zhang2026neurobridge} & 63.2 & 89.9 & 19.0 & 45.9 & 32.2 & 60.8 & 3.4 & 12.8 \\
  MBP~\citep{zheng2026learning} & 75.7 & 94.6 & 20.0 & 44.1 & 33.9 & 60.7 & \textbf{5.4} & 15.2 \\
  LVBP~\citep{liu2026leveraging} & 80.0 & 96.9 & 20.0 & 48.0 & 44.0 & 72.0 & 5.3 & 15.9 \\    
  DMSA~\citep{du2026deep} & 82.6 & 97.7 & 21.8 & 49.4 & 48.0 & 74.4 & 3.7 & 13.0 \\
    \rowcolor{gray!10}
    \textbf{BHA (Ours)} / Raw & 87.00 & 98.45 & 22.50 & 51.60 & 55.63 & 85.75 & 3.63 & 12.75 \\
    \rowcolor{gray!10}
    \textbf{BHA (Ours)} / CSLS-2 & \textbf{93.90} & \textbf{99.70} & \textbf{30.35} & \textbf{63.35} & \textbf{68.13} & \textbf{91.38} & 5.25 & \textbf{16.50} \\
    \bottomrule
  \end{tabular}
  }
\end{table}

\begin{table}[!htbp]
  \caption{\textbf{Visual reconstruction on THINGS-EEG2.}
  Ten-subject BHA means; metrics follow MindEye2~\citep{scotti2024mindeye2} and ATM~\citep{li2024visual}. \protect\colorbox{ceilingpink}{Pink}: image-conditioned and permutation controls
  (Appendix~\ref{sec:reconstruction_ceilings}). Baselines retain their reported
  generation settings (Appendix~\ref{sec:generator_comparison}).}
  \label{table:generation_comparison}
  
  \centering
  \small
  \setlength{\tabcolsep}{2pt}
  \resizebox{\fitwidth}{!}{%
  \begin{tabular}{lcccccccc}
  \toprule
  \textbf{Methods} 
  & \multicolumn{4}{c}{\textbf{Low-level Metrics}} 
  & \multicolumn{4}{c}{\textbf{High-level Metrics}} \\
  \cmidrule(lr){2-5} \cmidrule(lr){6-9}
  & PixCorr$\uparrow$ 
  & SSIM$\uparrow$ 
  & AlexNet(2)$\uparrow$ 
  & AlexNet(5)$\uparrow$ 
  & Inception$\uparrow$ 
  & CLIP$\uparrow$ 
  & EffNet-B$\downarrow$ 
  & SwAV$\downarrow$ \\
  \midrule
  \cellcolor{ceilingpink}\textit{w/ GT image feature ($\alpha=1.0$)} & 0.2877 & 0.4429 & \underline{0.9472} & \underline{0.9919} & \underline{0.9985} & \underline{0.9998} & 0.4064 & 0.2649 \\
  \cellcolor{ceilingpink}\textit{w/ GT latent ($\alpha=0.0$)} & \textbf{0.9971} & \textbf{0.9588} & \textbf{1.0000} & \textbf{1.0000} & \textbf{1.0000} & \textbf{1.0000} & \textbf{0.0168} & \textbf{0.0118} \\
  \cellcolor{ceilingpink}\textit{w/ GT feature + latent ($\alpha=0.5$)} & \underline{0.8868} & \underline{0.6840} & \textbf{1.0000} & \textbf{1.0000} & \textbf{1.0000} & \textbf{1.0000} & \underline{0.2198} & \underline{0.1310} \\
  \cellcolor{ceilingpink}\textit{Chance (non-target BHA)} & 0.1115 & 0.4004 & 0.5010 & 0.4750 & 0.4976 & 0.4910 & 0.9782 & 0.6698 \\
  \midrule
  \multicolumn{9}{c}{Within-subject (ten-subject mean)} \\
  \midrule  
  ATM-S~\citep{li2024visual} & 0.130 & 0.339 & 0.740 & 0.822 & 0.685 & 0.716 & - & 0.599 \\
  CC-All~\citep{zhang2024cognitioncapturer} & 0.143 & 0.347 & 0.754 & 0.623 & 0.669 & 0.715 & - & 0.590 \\
  AVDE~\citep{dai2026autoregressive} & 0.147 & 0.366 & 0.766 & 0.835 & 0.724 & 0.747 & - & 0.587 \\
  MBP~\citep{zheng2026learning} & 0.195 & 0.336 & \underline{0.843} & 0.905 & 0.756 & \underline{0.808} & - & 0.554 \\
  \rowcolor{gray!10}
  \textbf{BHA (Ours)} & \underline{0.277} & \underline{0.454} & \textbf{0.874} & \textbf{0.922} & \textbf{0.791} & \textbf{0.830} & \textbf{0.845} & \underline{0.516} \\
  \rowcolor{gray!10}
  \textbf{BHA + VAE-guided diffusion} & \textbf{0.283} & \textbf{0.464} & \textbf{0.874} & \underline{0.921} & \underline{0.790} & \textbf{0.830} & \textbf{0.845} & \textbf{0.515} \\

\bottomrule
  \end{tabular}
  }
  \end{table}
  
\section{Experiments}
\subsection{Experimental Setup}
We evaluate 200-way brain-to-image retrieval on THINGS-EEG2~\citep{gifford2022large}
and THINGS-MEG~\citep{hebart2023things} under within-subject and
leave-one-subject-out (LOSO) settings, and image reconstruction on THINGS-EEG2.
Results average ten EEG and four MEG subjects. BrainViT combines channel
attention with a feature--sensor convolutional readout; its teachers are
OpenCLIP ViT-H/14~\citep{cherti2023reproducible}, DFN-6~\citep{fang2024data}, MAE-18~\citep{he2022masked},
and DINOv2-12~\citep{oquab2024dinov}
(Appendices~\ref{sec:brainvit_architecture} and~\ref{sec:retrieval_implementation}).
Within-subject retrieval uses Stage~2 gated scores; LOSO matches Stage~1
outputs to fused visual features. Raw scores each query independently.
Cross-domain similarity local scaling~\citep[CSLS;][]{Conneau_2017_171004087} uses all
200 test queries as unlabeled references, making it transductive.
See Appendix~\ref{sec:implementation_details} for preprocessing and training.

\subsection{Retrieval Results}
BHA achieves within-subject Raw Top-1 accuracy of $87.00\%$ on EEG and
$55.63\%$ on MEG, rising to $93.90\%$ and $68.13\%$ with CSLS-2
(Table~\ref{table:retrieval_comparison}). These Raw scores exceed the reported
within-subject baselines in both modalities. Under LOSO, pretrained BrainViT
achieves $22.50\%$ and $3.63\%$ Raw Top-1, respectively, identifying
cross-subject transfer as a remaining limitation. CSLS rescoring uses
unchanged predictions and depends on the reference cohort; \redhl{$k=2$ is selected on
validation data (\mbox{Appendix~\ref{sec:csls_validation_k}}).} Subject-wise EEG and MEG results and
scoring sensitivity appear in Appendices~\ref{sec:comprehensive_thingseeg2},
\ref{sec:comprehensive_thingsmeg}, and~\ref{sec:retrieval_csls_ablation}.

\subsection{Reconstruction Results}
BHA supports reconstruction with separately pretrained BrainViT weights
and a frozen generator (Table~\ref{table:generation_comparison}). Across
ten EEG subjects, it achieves PixCorr $0.277$, SSIM $0.454$, and CLIP
$0.830$, improving on MBP's reported $0.195$, $0.336$, and $0.808$,
respectively~\citep{zheng2026learning}. VAE guidance further improves
pixel and structural fidelity while preserving CLIP identification.
These comparisons reflect complete systems, including differences in
generators, conditioning interfaces, and sampling settings
(Appendix~\ref{sec:generator_comparison}). Permuting the predicted conditions
reduces identification to near chance, supporting stimulus-specific
information in the reconstruction predictions. Evaluation details and
subject-wise results appear in
Appendix~\ref{sec:comprehensive_reconstruction_performance}, with examples
in Appendix~\ref{sec:add_reconst_display}.

\begin{figure}[!t]
    \centering
    \includegraphics[width=1\textwidth]{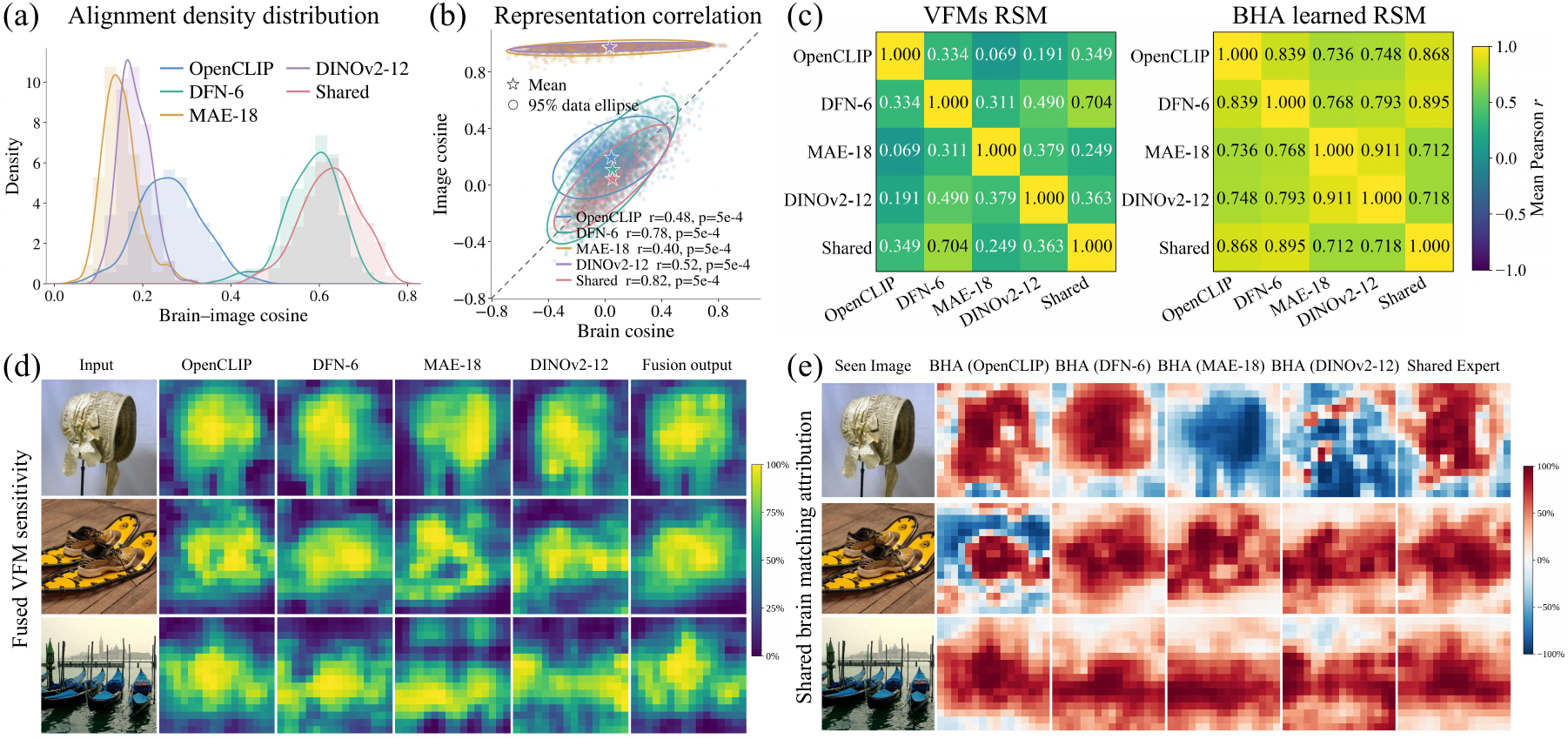}
    \caption{\textbf{Shared geometry with distinct matching contributions.}
    (a) Paired brain--image cosine similarity.
    (b) Agreement between inter-image similarities in student and visual spaces.
    (c) Correlations among representational similarity matrices for VFMs (left)
    and student outputs (right), averaged over ten subjects.
    (d) Visual sensitivity maps.
    (e) Occlusion attributions for each teacher branch and the shared expert:
    red/blue indicate positive/negative contributions to matching.
    Colors denote within-map percentiles, not comparable effect magnitudes.}
    \label{fig:rsa}
\end{figure}

\subsection{Ablation Studies}
The gated readout improves within-subject retrieval over both pretrained
BrainViT and the shared expert across training-data scales
(Figure~\ref{fig:data_scale_withinsubject}). \redhl{Under LOSO, the gated readout does not improve on pretrained BrainViT
(Table~\ref{table:stage_main}): the brain encoder itself transfers poorly
($22.50\%$ versus $80.80\%$ within-subject), leaving little teacher-specific
signal to compose. With the retrieval adapter kept, linear, MLP, and Group MoE
readouts reach the same accuracy and the same composition gain
(Appendix Table~\ref{table:moe_readout_ablation}), so the gain comes from
composing teacher-specific scores.}

\begin{table}[!t]
\centering
\begin{minipage}[c]{0.53\textwidth}
\setlength{\abovecaptionskip}{0pt}
\caption{\textbf{Retrieval readouts across training stages.}
THINGS-EEG2 Top-1 (\%, $\uparrow$), full training data.}
\label{table:stage_main}
\centering
\footnotesize
\setlength{\tabcolsep}{3pt}
\begin{tabular}{@{}lrrr@{}}
\toprule
Stage / readout & Raw & CSLS-2 & CSLS-10 \\
\midrule
\multicolumn{4}{c}{\textit{Within-subject}} \\
\addlinespace[2pt]
Stage 1 / pretrained & 80.80 & 90.95 & 87.85 \\
Stage 2 / shared expert & 82.20 & 91.95 & 88.65 \\
Stage 2 / gated & \textbf{87.00} & \textbf{93.90} & \textbf{91.35} \\
\midrule
\multicolumn{4}{c}{\textit{LOSO}} \\
\addlinespace[2pt]
Stage 1 / pretrained & \textbf{22.50} & 30.35 & \textbf{31.65} \\
Stage 2 / shared expert & 22.05 & \textbf{30.50} & 31.55 \\
Stage 2 / gated & 20.20 & 28.15 & 29.90 \\
\bottomrule
\end{tabular}
\end{minipage}\hfill
\begin{minipage}[c]{0.45\textwidth}
\centering
\includegraphics[width=1\linewidth,pagebox=cropbox]{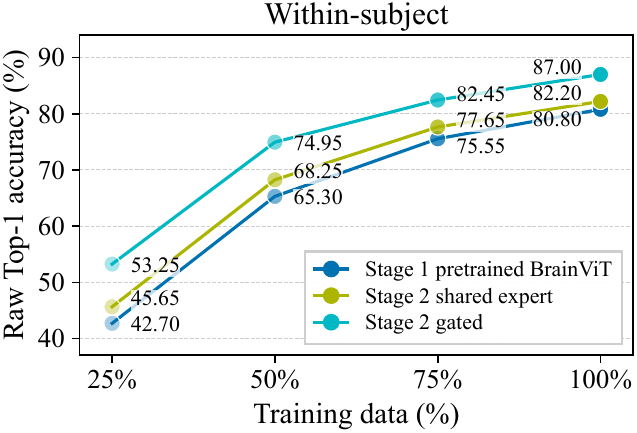}
\captionof{figure}{\textbf{Readout gains across training-data scales.}
THINGS-EEG2 within-subject Raw Top-1 (\%); ten-subject means.}
\label{fig:data_scale_withinsubject}
\end{minipage}
\end{table}

Reconstruction benefits depend on the evaluation metric. In matched
pretraining controls, fused visual targets yield higher SSIM, whereas
OpenCLIP yields higher CLIP identification at stronger generation settings
(Appendix~\ref{sec:reconstruction_pretraining_control}). These results
motivate evaluating structural fidelity and semantic identification separately.
Under BiCLIP, visual adaptation improves paired cosine ($0.217\to0.481$) and Raw Top-1 ($30.05\%\to37.80\%$), but lowers CLIP identification ($0.754\to0.706$). Under MSE--BiCLIP, adaptation lowers both retrieval and reconstruction (Appendix~\ref{sec:visual_adapter_task_control}). Joint brain--visual adaptation can therefore improve matching without preserving generator compatibility.

\begin{figure}[!t]
    \centering
    \includegraphics[width=1\textwidth]{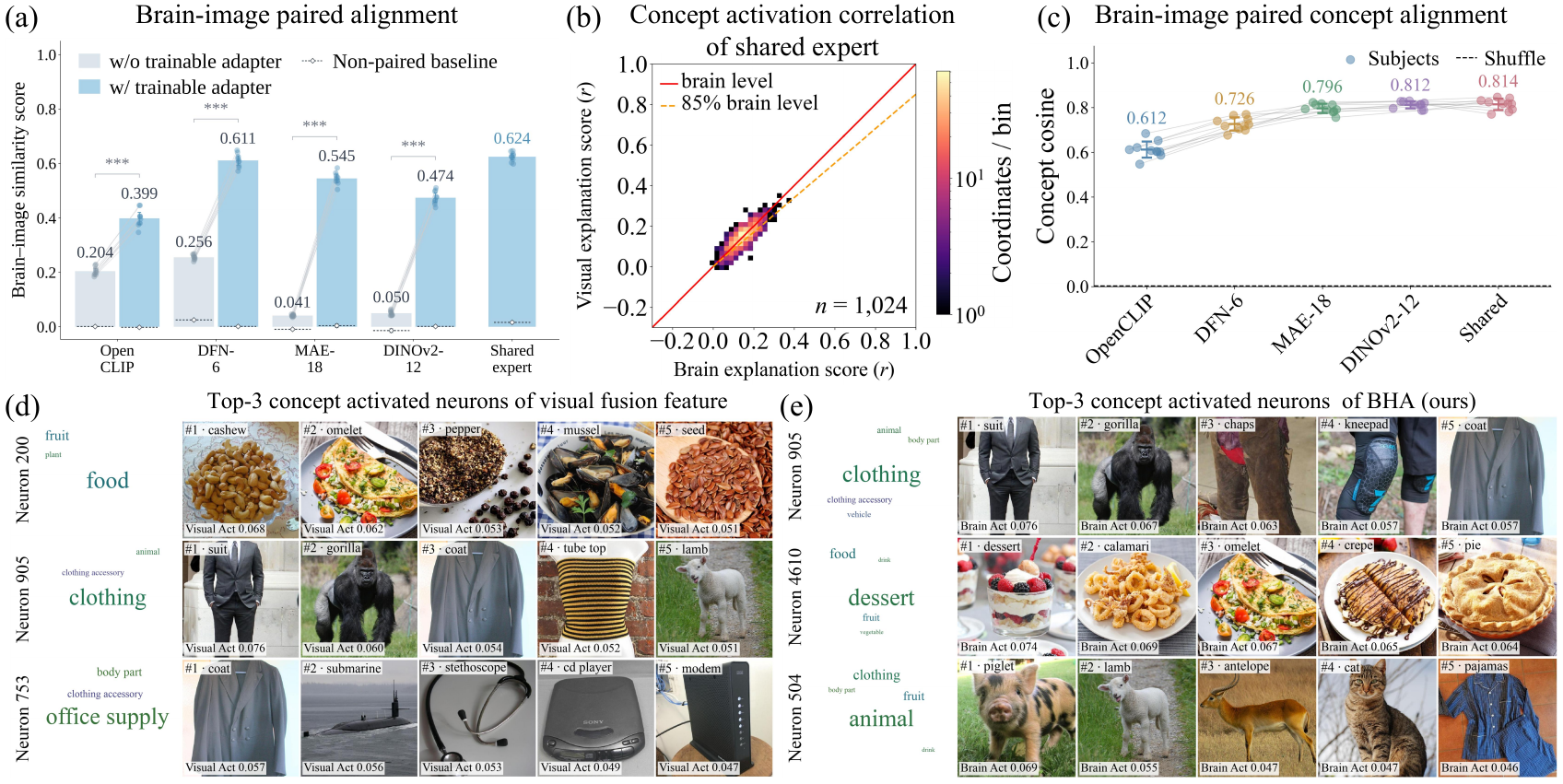}
    \caption{\textbf{Feature alignment and semantic selectivity capture different structure.}
    (a) Paired brain--image similarity before and after visual adaptation.
    (b) Shared-expert concept correlations on test images; concepts are selected
    separately for student and visual features using training images
    (Appendix Figure~\ref{fig:concept_scatter}).
    (c) Brain--visual concept-profile alignment versus neuron shuffling on
    training images. (b--c) report subject means. (d--e) Concept associations
    and activating test images for training-selected visual-fusion and BHA neurons.}
    \label{fig:concept_analysis}
\end{figure}

\subsection{Representation Analysis}
\subsubsection{Representational Alignment and Matching}
We examine whether teacher-specific readouts
learn distinct matching functions through paired brain--image similarity,
inter-image geometry, and spatial attributions. Student outputs have more similar geometries than their
visual targets, with cross-expert correlations of $0.712$--$0.911$ versus
$0.069$--$0.704$ (Figure~\ref{fig:rsa}(c)). The shared expert also best
preserves inter-image relationships ($r=0.82$;
Figure~\ref{fig:rsa}(a--b)). Yet overlapping visual sensitivity can yield
opposite signed contributions to matching across experts
(Figure~\ref{fig:rsa}(d--e)). Similar embedding geometry thus does not
imply interchangeable matching functions. The shared geometry may
reflect common inputs, backbone, and supervision; these model-level
analyses do not establish its neural origin or biological causality
(Appendix~\ref{sec:extended_representation_analysis}).

\subsubsection{Concept Associations in Model Outputs}
To test whether feature alignment preserves concept structure, we use
CLIP-Dissect~\citep{oikarinen2023clip} to associate model neurons with 27
THINGS categories, selecting concepts on training images and evaluating
correlations on held-out images (Figure~\ref{fig:concept_analysis}(b)).
On training images, the shared expert and DINOv2-12 show the strongest
brain--visual concept-profile alignment (mean cosine $0.814$ and $0.812$),
and all branches exceed neuron shuffling (Figure~\ref{fig:concept_analysis}(c)).
This ordering differs from paired feature alignment
(Figure~\ref{fig:concept_analysis}(a)), suggesting that concept correspondence
captures a distinct property of the learned space. Pooled activations
illustrate food-, clothing-, and animal-related associations
(Figure~\ref{fig:concept_analysis}(d--e)).
These results characterize semantic selectivity within learned models;
they do not establish corresponding neuron identities across subjects.
Appendices~\ref{sec:extended_representation_analysis} and~\ref{sec:concept_examples}
provide subject-wise scores and additional galleries.
\redhl{Temporal masking on four datasets probes model dependence on
response windows (\mbox{Appendix~\ref{sec:temporal_window_ablation}}).}

\begin{figure}[!t]
    \centering
    \includegraphics[width=1\textwidth]{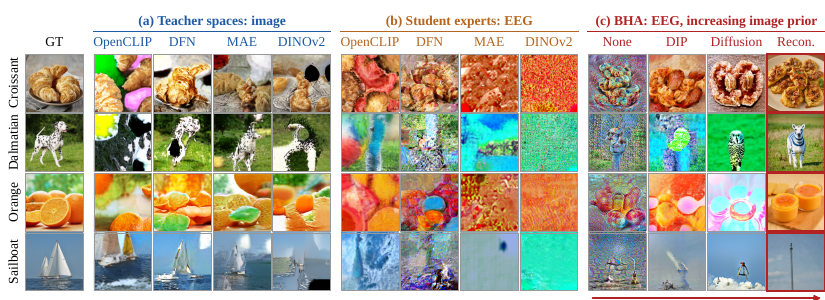}
    \caption[Visual teachers encode different features, and BHA reconstructions follow the EEG prediction.]{\redhl{\textbf{Visual teachers encode different features, and BHA
    reconstructions follow the EEG prediction (subject-08).}
    (a) Teacher image features and (b) EEG-predicted expert outputs, inverted
    with a deep image prior~\citep[DIP;][]{Ulyanov_2018_CVPR}. (c) Gated BHA under increasing priors: none, DIP,
    ADM~\citep{dhariwal2021diffusion} guidance, and our reconstruction ($\alpha=0.75$).}}
    \label{fig:prior_case_study}
\end{figure}

\subsubsection{\texorpdfstring{\protect\redhl{Case Study: Teacher Diversity and Image Priors}}{Case Study: Teacher Diversity and Image Priors}}
\label{sec:prior_case_study}
\redhl{Feature inversion under increasing priors
(\mbox{Figure~\ref{fig:prior_case_study}}; \mbox{Appendix~\ref{sec:feature_inversion}})
shows that the teachers recover different stimulus properties, from object
parts (OpenCLIP) and colour and texture (DFN, MAE) to contours (DINOv2), which
the EEG experts inherit. Stronger priors make BHA outputs more natural but keep
the EEG-predicted content (e.g., a striped dalmatian), so the generator renders
the brain prediction rather than restoring the stimulus.}

\begin{figure}[!t]
    \centering
    \includegraphics[width=0.9\textwidth]{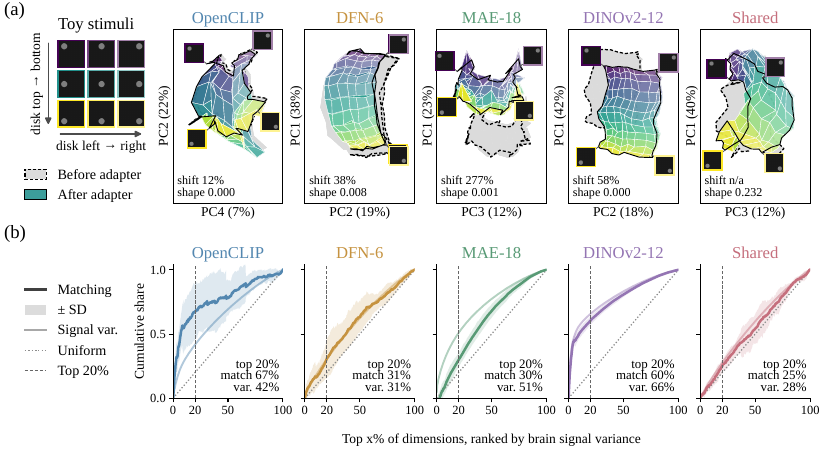}
    \caption[Adapters preserve teacher-specific geometry, and experts use their dimensions differently.]{\redhl{\textbf{Adapters preserve teacher-specific geometry, and experts use their
    dimensions differently.} (a) Toy disk images (left) in each visual space
    before (grey) and after (coloured) the adapter, on the leading PCs that best
    encode position (subject~08; grey is the aligned teacher
    concatenation). shift: displacement over grid size; shape: Procrustes
    disparity. (b)
    Cumulative share of brain--image matching (dark, $\pm$SD) and of EEG signal
    variance (light; across-image variance after removing trial noise) over the
    top $x\%$ of dimensions ranked by signal variance.}}
    \label{fig:teacher_geometry}
\end{figure}

\subsubsection{\texorpdfstring{\protect\redhl{Teacher Diversity in Geometry and Dimension Usage}}{Teacher Diversity in Geometry and Dimension Usage}}
\label{sec:teacher_geometry}
\redhl{To explore how the visual adapter shapes the visual teacher space, following \citet{lyo2025disentangled}, we map synthetic disk images
into each visual space to test whether the adapters erase teacher-specific
structure (\mbox{Figure~\ref{fig:teacher_geometry}(a)}). The teachers organize position
differently (near-Cartesian in DINOv2, curved in DFN, folded in OpenCLIP and
MAE); the adapters translate these sheets but preserve their shape
(Procrustes disparity $\leq 0.008$), and only the shared projector reshapes
the joint geometry (0.23). When dimensions are ranked by EEG signal variance
and an occlusion-based matching score is split per dimension
(\mbox{Figure~\ref{fig:teacher_geometry}(b)}), the top 20\% carry 67\% of the
matching in OpenCLIP but 30\% in MAE, while DFN and the shared
expert spread it almost uniformly; the adapters thus preserve complementary
teacher structure.}

\section{Conclusion}
\redhl{We introduced Beyond Hierarchical Alignment (BHA), a multi-teacher distillation framework that, instead of collapsing visual teachers into a single target, keeps a shared brain--visual alignment alongside teacher-specific prediction spaces.} Composing shared and teacher-specific scores improves within-subject
retrieval, while generator-native conditions enable reconstruction without
generator fine-tuning. \redhl{Analyses indicate that the two kinds of space are complementary: the adapters preserve teacher-specific geometry while the shared projector reshapes it, and experts use their dimensions and image regions differently.} The gains are task-dependent: readout specialization does not consistently
improve cross-subject transfer, and richer pretraining does not improve
every reconstruction metric. These findings motivate preserving complementary
prediction spaces while evaluating discrimination, reconstruction fidelity,
and cross-subject generalization separately.

\section*{Reproducibility statement}
Proofs appear in Appendix~\ref{sec:proof_of_propositions}; datasets, preprocessing, architecture, objectives, and inference settings are specified in Appendix~\ref{sec:implementation_details}. Appendix~\ref{sec:add_experimental_results} provides subject-wise results, ablations, and analysis protocols. Source code will be released via the project page.

\section*{Ethics statement}
This study uses only existing recordings from \redhl{six} publicly available neural
datasets: THINGS-EEG1, THINGS-EEG2, THINGS-MEG, Alljoined-1.6M\redhl{, NSD, and TVSD. NSD contains human fMRI and TVSD contains macaque intracortical recordings}.
We did not recruit participants, conduct new human-subject \redhl{or animal} experiments,
or collect new neural recordings.

\section*{AI use statement}
The authors used large language model tools to support manuscript drafting and revision, literature search and comparison, experimental planning, methodological assessment, code development, and interpretation of results. All manuscript text, including AI-assisted content, was reviewed, revised as needed, and approved by the authors. The authors retain full responsibility for the manuscript, results, and scientific claims, including the correctness of all mathematical statements and proofs.

\bibliographystyle{iclr2027_conference}
\bibliography{references}

\clearpage
{\LARGE\bfseries Appendix\par}
\vspace{0.4cm}
\appendix
\renewcommand{\floatpagefraction}{0.85}
\setcounter{topnumber}{4}
\setcounter{bottomnumber}{3}
\setcounter{totalnumber}{6}
\setcounter{figure}{0}
\renewcommand{\thefigure}{A.\arabic{figure}}
\renewcommand{\theHfigure}{appendix.A.\arabic{figure}}
\setcounter{table}{0}
\renewcommand{\thetable}{A.\arabic{table}}
\renewcommand{\theHtable}{appendix.A.\arabic{table}}
\section{Appendix Overview}
The contents below link to the appendix sections and subsections; page
numbers appear on the right.
\begingroup
\small
\setlength{\parskip}{2pt}
\newcommand{\appendixentry}[3]{%
  \par\noindent\hspace*{#1}%
  \hyperref[#2]{\ref*{#2}\quad #3}%
  \nobreak\hfill\hyperref[#2]{\pageref*{#2}}\par}
\appendixentry{0pt}{sec:proof_of_propositions}{Theoretical Analysis}
\appendixentry{1em}{sec:proof_multiteacher_alignment_bound}{Proof of Proposition~\ref*{prop:multiteacher_alignment_bound}}
\appendixentry{1em}{sec:proof_alignment_reconstruction_gap}{Proof of Proposition~\ref*{prop:alignment_reconstruction_gap}}
\appendixentry{0pt}{sec:implementation_details}{Implementation Details}
\appendixentry{1em}{sec:dataset_description}{Datasets and Settings}
\appendixentry{1em}{sec:method_details}{Detailed Objectives and Training}
\appendixentry{1em}{sec:brainvit_architecture}{Model Architecture}
\appendixentry{1em}{sec:retrieval_implementation}{Retrieval Implementation}
\appendixentry{1em}{sec:reconstruction_implementation}{Reconstruction Implementation}
\appendixentry{0pt}{sec:add_experimental_results}{Additional Experimental Results}
\appendixentry{1em}{sec:comprehensive_bha_results}{Comprehensive BHA Results}
\appendixentry{1em}{sec:extended_representation_analysis}{Extended Representation Analysis}
\appendixentry{1em}{sec:comprehensive_reconstruction_performance}{Comprehensive Reconstruction Performance}
\appendixentry{1em}{sec:nsd_results}{\redhl{Transfer to fMRI: NSD}}
\appendixentry{1em}{sec:tvsd_results}{\redhl{Transfer to Intracortical Recordings: TVSD}}
\appendixentry{0pt}{sec:supplementary_examples}{Supplementary Examples}
\appendixentry{1em}{sec:concept_examples}{Neuron Concept Associations}
\appendixentry{1em}{sec:feature_inversion_examples}{Feature Inversion Examples}
\appendixentry{1em}{sec:add_reconst_display}{Reconstruction Examples}
\endgroup

\section{Theoretical Analysis}
\label{sec:proof_of_propositions}
We first establish the alignment-risk bound of
Proposition~\ref{prop:multiteacher_alignment_bound}, then prove
Proposition~\ref{prop:alignment_reconstruction_gap} by separating retrieval
alignment to adapted targets from fidelity to native generator outputs.

\subsection{Proof of Proposition~\ref{prop:multiteacher_alignment_bound}}
\label{sec:proof_multiteacher_alignment_bound}
\paragraph{General weighted form.}
Retain the signal model and independence assumptions of
Proposition~\ref{prop:multiteacher_alignment_bound}, but allow unequal
teacher error variances. Define the error covariance matrix through
$C_{k\ell}=\mathbb{E}[\boldsymbol{\delta}_k^{\top}\boldsymbol{\delta}_{\ell}]/d_0$.
Choose weights $\mathbf{w}\in\Delta_K=\{\mathbf{w}\geq0:\sum_k w_k=1\}$
independently of the training samples and fit
\begin{equation}
\widehat{\mathbf{M}}_{\mathbf{w}}
=\arg\min_{\mathbf{M}}\frac{1}{nd_0}
\sum_{i=1}^{n}\sum_{k=1}^{K}w_k
\|\mathbf{M}\mathbf{b}_i-\mathbf{t}_{ik}\|_2^2.
\label{eq:multiteacher_distillation_estimator}
\end{equation}
With $\widehat g_{\mathbf{w}}(\mathbf{b})=
\widehat{\mathbf{M}}_{\mathbf{w}}\mathbf{b}$, we will show
\begin{equation}
\begin{aligned}
\mathbb{E}_{\mathrm{train}}[\mathcal{R}_{\mathrm{vis}}(\widehat g_{\mathbf{w}})
\mid\mathsf{B}]
&=\sigma_{\mathrm{br}}^2+
\chi_n\bigl(\sigma_{\mathrm{br}}^2+\mathbf{w}^{\top}\mathbf{C}\mathbf{w}\bigr)\\
&\leq\sigma_{\mathrm{br}}^2+
\frac{p}{n\kappa}\bigl(\sigma_{\mathrm{br}}^2+
\mathbf{w}^{\top}\mathbf{C}\mathbf{w}\bigr).
\end{aligned}
\label{eq:multiteacher_alignment_risk_bound}
\end{equation}
Uniform weights recover $\widehat g_K$ in the proposition.

\paragraph{Predictability floor and fused supervision.}
Since $\mathbb{E}[\boldsymbol{\xi}\mid\mathbf{b}]=0$, the conditional mean
$\mathbf{M}_{\star}\mathbf{b}$ minimizes squared prediction error, and
\[
\mathcal{R}_{\mathrm{vis}}(g)=\sigma_{\mathrm{br}}^2+
\frac{1}{d_0}\mathbb{E}
\|g(\mathbf{b})-\mathbf{M}_{\star}\mathbf{b}\|_2^2.
\]
This proves the lower bound. Define
$\overline{\mathbf{t}}_{i,\mathbf{w}}=\sum_k w_k\mathbf{t}_{ik}$.
For any prediction $\mathbf{v}$,
\[
\sum_k w_k\|\mathbf{v}-\mathbf{t}_{ik}\|_2^2
=\|\mathbf{v}-\overline{\mathbf{t}}_{i,\mathbf{w}}\|_2^2+
\sum_k w_k\|\mathbf{t}_{ik}-\overline{\mathbf{t}}_{i,\mathbf{w}}\|_2^2.
\]
The second term is independent of the prediction. Thus multi-teacher
least squares is equivalent to regression on the fused target, whose
residual is $\boldsymbol{\eta}_{i,\mathbf{w}}=
\boldsymbol{\xi}_i+\sum_k w_k\boldsymbol{\delta}_{ik}$.
Independence and zero means give
$\mathbb{E}\|\boldsymbol{\eta}_{i,\mathbf{w}}\|_2^2/d_0
=\sigma_{\mathrm{br}}^2+\mathbf{w}^{\top}\mathbf{C}\mathbf{w}$.

\paragraph{Finite-sample alignment error.}
Let $\mathsf{N}_{\mathbf{w}}\in\mathbb{R}^{n\times d_0}$ contain the
residuals $\boldsymbol{\eta}_{i,\mathbf{w}}^{\top}$ as rows.
The full-rank normal equations yield
\[
\widehat{\mathbf{M}}_{\mathbf{w}}-\mathbf{M}_{\star}
=\mathsf{N}_{\mathbf{w}}^{\top}\mathsf{B}
(\mathsf{B}^{\top}\mathsf{B})^{-1}.
\]
For an independent test feature, $\mathbb{E}[\mathbf{b}\mathbf{b}^{\top}]=I_p$
implies that its prediction error relative to the conditional mean is
$\|\widehat{\mathbf{M}}_{\mathbf{w}}-\mathbf{M}_{\star}\|_F^2/d_0$.
Conditional on $\mathsf{B}$, the training residuals are independent with
zero means, so
\[
\mathbb{E}[\mathsf{N}_{\mathbf{w}}\mathsf{N}_{\mathbf{w}}^{\top}
\mid\mathsf{B}]
=d_0(\sigma_{\mathrm{br}}^2+\mathbf{w}^{\top}\mathbf{C}\mathbf{w})I_n.
\]
Substitution into the squared Frobenius norm gives
\[
\frac{1}{d_0}\mathbb{E}_{\mathrm{train}}\!\left[
\|\widehat{\mathbf{M}}_{\mathbf{w}}-\mathbf{M}_{\star}\|_F^2
\mid\mathsf{B}\right]
=(\sigma_{\mathrm{br}}^2+\mathbf{w}^{\top}\mathbf{C}\mathbf{w})\chi_n.
\]
Adding the independent test residual proves the equality in
Eq.~\eqref{eq:multiteacher_alignment_risk_bound}.
Every eigenvalue of $\mathsf{B}^{\top}\mathsf{B}$ is at least $n\kappa$,
hence $\chi_n\leq p/(n\kappa)$, proving the upper bound.

\paragraph{When multiple teachers improve alignment.}
For uniform weights, the covariance assumptions give
\[
\mathbf{w}^{\top}\mathbf{C}\mathbf{w}
=\frac{1}{K^2}\sum_{k,\ell}C_{k\ell}
\leq\sigma_{\mathrm{T}}^2
\left(\rho_{\mathrm{T}}+\frac{1-\rho_{\mathrm{T}}}{K}\right)
<\sigma_{\mathrm{T}}^2.
\]
Since $\chi_n>0$, Eq.~\eqref{eq:multiteacher_alignment_risk_bound}
establishes strict improvement over every single teacher with variance
$\sigma_{\mathrm{T}}^2$, completing the proof.
For unequal teacher quality, uniform averaging need not improve on the
best teacher. Oracle weights instead satisfy
$\min_{\mathbf{w}\in\Delta_K}\mathbf{w}^{\top}\mathbf{C}\mathbf{w}
\leq\min_k C_{kk}$, because each single-teacher choice belongs to the
simplex; this does not guarantee that weights learned from finite data
attain the oracle risk.

\paragraph{Relation to angular alignment.}
For any nonzero vectors $u,v$, the triangle and reverse-triangle
inequalities give
\[
\|\operatorname{norm}(u)-\operatorname{norm}(v)\|_2
\leq\frac{2\|u-v\|_2}{\|v\|_2}.
\]
If $g(\mathbf{b})\neq0$ and $\|\mathbf{y}\|_2\geq r_{\min}>0$
almost surely, then
\begin{equation}
\mathbb{E}\bigl[1-\operatorname{sim}(g(\mathbf{b}),\mathbf{y})\bigr]
\leq\frac{2d_0}{r_{\min}^2}\mathcal{R}_{\mathrm{vis}}(g).
\label{eq:multiteacher_angular_bound}
\end{equation}
This follows by squaring the preceding inequality and using
$1-\operatorname{sim}(u,v)=
\|\operatorname{norm}(u)-\operatorname{norm}(v)\|_2^2/2$.
The same inequality holds after averaging over training samples,
conditional on the design. Thus a tighter feature-risk bound also tightens
this angular-error bound under the additional norm condition; it need not
improve the realized cosine similarity or retrieval ranking, which also
depends on similarities to unmatched images.

\paragraph{Implications for brain--vision alignment.}
Complementarity is useful here when teacher errors are not fully shared:
multiple sources reduce estimation error around the brain-predictable
visual component, especially when $n$ is small relative to $p$.
They do not reduce the irreducible term $\sigma_{\mathrm{br}}^2$; identical
teacher errors provide no averaging benefit. The result assumes unbiased,
commensurate teacher targets and a well-conditioned linear design.
It motivates learning complementary visual supervision, but does not
establish these assumptions for BHA's learned, normalized fusion or give
a guarantee for its BiCLIP objective, retrieval accuracy, or reconstruction
quality. In particular, lower alignment risk supports better estimation
of recoverable visual information, not complete recovery of stimulus content.

\subsection{Proof of Proposition~\ref{prop:alignment_reconstruction_gap}}
\label{sec:proof_alignment_reconstruction_gap}

\paragraph{Alignment error and the native reconstruction space.}
The decoder $G$ includes the frozen conditioning interface and generator;
$G(\mathbf{e})$ is its output under the original visual feature, not the
ground-truth image. The proposition compares this native output with
outputs driven by adapted visual features and brain predictions. All
three features must lie in the decoder's input domain. For a stochastic
generator, keep its random input, or average over a shared random input
independent of the features, assuming the same Lipschitz bound.

\begin{proof}
For a square-integrable random vector $\mathbf{A}$, write
$\|\mathbf{A}\|_{L^2}=(\mathbb{E}\|\mathbf{A}\|_2^2)^{1/2}$.
Applying the reverse triangle inequality in $L^2$ gives
\begin{align}
\left|\sqrt{\mathcal{R}_G}-\sqrt{\Delta_G}\right|
&=\left|\|G(\mathbf{h})-G(\mathbf{e})\|_{L^2}
       -\|G(\mathbf{t})-G(\mathbf{e})\|_{L^2}\right|\notag\\
&\leq\|G(\mathbf{h})-G(\mathbf{t})\|_{L^2}
\leq L_G\|\mathbf{h}-\mathbf{t}\|_{L^2}
=L_G\sqrt{2\delta}.
\label{eq:alignment_native_reconstruction_proof}
\end{align}
The last equality uses unit norms. If $\delta=0$, then
$\mathbf{h}=\mathbf{t}$ almost surely and $\mathcal{R}_G=\Delta_G$.
\end{proof}

For unnormalized predictions and targets, the same proof replaces
$\sqrt{2\delta}$ by $(\mathbb{E}\|\mathbf{h}-\mathbf{t}\|_2^2)^{1/2}$;
cosine similarity alone does not constrain their magnitudes.
The bound holds for each jointly learned pair $(f,V)$. Better angular
alignment brings the brain-conditioned output closer to the adapted
visual output; it does not constrain how far that output moves from the
native one. In particular,
$\mathcal{R}_G\geq(\sqrt{\Delta_G}-L_G\sqrt{2\delta})_+^2$,
where $(x)_+=\max\{x,0\}$. To relate the statement to ground-truth
reconstruction, let
$\mathcal{R}_{\mathrm{native}}=\mathbb{E}\|G(\mathbf{e})-\mathbf{I}\|_2^2$.
Another triangle inequality yields
\begin{equation}
\mathbb{E}\|G(\mathbf{h})-\mathbf{I}\|_2^2
\geq\left(\sqrt{\Delta_G}-L_G\sqrt{2\delta}
-\sqrt{\mathcal{R}_{\mathrm{native}}}\right)_+^2.
\label{eq:alignment_image_reconstruction_bound}
\end{equation}
Thus a sufficiently large adaptation-induced drift cannot be removed by
perfecting alignment to the adapted target alone.

\paragraph{Joint brain--visual adaptation and reconstruction.}
Let $C$ index an image category, $\mathbf{u}_C$ be its standard basis
vector, and $U$ an independent uniform sign representing a visual detail.
The native feature is $\mathbf{e}=(\mathbf{u}_C,U)/\sqrt{2}$.
Let the brain input contain $(C,W)$, where $W=SU$ is a noisy observation
of that detail: $S\in\{-1,+1\}$ is independent of $(C,U)$ and
$\mathbb{E}[S]=c$. Choose $1/2<\beta<c<1$. For $q\in[0,1]$, define
\begin{equation}
\mathbf{t}_q=\frac{(\mathbf{u}_C,qU)}{\sqrt{1+q^2}},
\qquad
\mathbf{h}_q=\frac{(\mathbf{u}_C,\beta qW)}{\sqrt{1+\beta^2q^2}},
\qquad G(\mathbf{z})=\mathbf{z}.
\label{eq:alignment_information_counterexample}
\end{equation}
As $q$ decreases, both sides suppress a detail represented more weakly
and noisily in the brain prediction. With the native feature itself as
the reconstruction target, the expected angular and reconstruction errors are
\begin{equation}
\delta(q)=1-\frac{1+\beta c q^2}
{\sqrt{(1+q^2)(1+\beta^2q^2)}},
\qquad
\mathcal{R}_G(q)=2-\frac{\sqrt{2}(1+\beta c q)}{\sqrt{1+\beta^2q^2}}.
\label{eq:alignment_reconstruction_counterexample}
\end{equation}
Conditional on $S=+1$ or $S=-1$, the paired angle is respectively
$\vartheta_+(q)=\arctan(q)-\arctan(\beta q)$ or
$\vartheta_-(q)=\arctan(q)+\arctan(\beta q)$.
Both increase strictly with $q$: in particular,
$\vartheta_+'(q)=(1-\beta)(1-\beta q^2)/[(1+q^2)(1+\beta^2q^2)]>0$.
Hence $\delta(q)$ strictly decreases as $q$ decreases from one to zero,
whereas
$\mathcal{R}_G'(q)=-\sqrt{2}\beta(c-\beta q)/(1+\beta^2q^2)^{3/2}<0$
shows that reconstruction error strictly increases. At $q=0$, alignment
is perfect but $\mathcal{R}_G=\Delta_G=2-\sqrt{2}>0$.

To check ranking separately, take a two-image gallery with distinct
categories and independent draws of their details and observation noise.
At $q=1$, a brain-to-image query is misranked exactly when its detail is
observed incorrectly and the distractor has the opposite detail.
An image-to-brain query is misranked exactly when its paired brain detail
is incorrect and the distractor brain detail agrees with the query image.
Since $\beta>1/2$, each event makes the unmatched score exceed the matched
score; each has probability $(1-c)/4$. Expected Top-1 accuracy in both
directions therefore increases from $(3+c)/4<1$ at $q=1$ to one at $q=0$,
while reconstruction worsens. At $q=0$, both embeddings discard $U$;
even an unrestricted decoder then has minimum expected error $1/2$ for
recovering $\mathbf{e}$, the variance of its missing last component.
For $q>0$, the corresponding minima are zero from $\mathbf{t}_q$ and
$(1-c^2)/2$ from $\mathbf{h}_q$.

\paragraph{Implications for task-specific prediction.}
The proposition concerns fidelity to the generator's native output under
an explicit regularity assumption; it does not assert that every visual
adapter destroys semantic or spatial information. Invertible adaptation
can alter generator compatibility without losing information, whereas a
many-to-one mapping can remove content altogether. The constructed family
shows that alignment improvement and reconstruction degradation can
coexist, not that an optimizer must follow this path. For BHA, this
motivates predicting native semantic conditions and VAE latents through
separate adapters on the shared backbone. These predictors bypass the
adapted retrieval spaces and use information retained in backbone features;
they cannot recover content already discarded by the backbone.

\section{Implementation Details}
\label{sec:implementation_details}
Experiments used eight NVIDIA A800 GPUs. This section describes the
datasets, training objectives, architectures, and inference procedures.
EEG retrieval hyperparameters are reported with the experimental settings
in Section~\ref{sec:comprehensive_thingseeg2}
(Table~\ref{table:current_eeg_hyperparameters}).

\subsection{Datasets and Settings}
\label{sec:dataset_description}

\paragraph{THINGS-EEG2.}
THINGS-EEG2 contains recordings from 10 participants, with 16,540 training
images repeated four times and 200 test images repeated 80 times per
participant~\citep{gifford2022large}.
We band-pass filter the EEG at 0.1--100 Hz, apply baseline correction using
the 200 ms pre-stimulus interval, and downsample from 1,000 to 250 Hz.
We retain 63 EEG channels and 250 time samples covering the first second
after stimulus onset, and apply multivariate noise
normalization~\citep{guggenmos2018multivariate}.

Training and validation inputs average four trials per image; test inputs
average 80 trials within each subject, including the held-out subject in
LOSO. Each input contains $C\times250$ values, with $C\in\{63,17\}$
depending on the prediction branch. Trial-count ablations vary the number
of averaged responses.

Pretraining samples one channel configuration per minibatch;
distillation uses all 63 channels. The 17-channel
posterior subset comprises \textbf{P7, P5, P3, P1, Pz, P2, P4, P6, P8,
PO7, PO3, POz, PO4, PO8, O1, Oz, and O2}, in the stored channel order.
The low-level reconstruction branch selects these channels before neural
encoding, whereas the high-level branch retains the full montage during
both training and inference.

\paragraph{THINGS-MEG.}
THINGS-MEG contains 271-channel recordings from four subjects over 12
sessions~\citep{hebart2023things}. The stimulus pool covers 1,854 concepts
with 12 images per concept. We exclude the 200 test concepts from training;
each test concept has one image presented 12 times. Images are shown for
500 ms with an onset interval of $1{,}000\pm200$ ms. We apply a 0.1--40 Hz
band-pass filter and baseline correction, downsample to 200 Hz, and use
the 0--1,000 ms interval. Inputs contain $271\times201$ values, including
both temporal endpoints. All retrieval models use the full sensor montage
and match each response to its presented image.

\paragraph{Additional datasets for temporal analysis.}
THINGS-EEG1 and Alljoined-1.6M extend the temporal-window comparison to
different stimulus timing and recording systems
(Appendix~\ref{sec:temporal_window_ablation}).

\paragraph{THINGS-EEG1.}
THINGS-EEG1 contains 64-channel EEG from 50 participants viewing 22,248
images from 1,854 concepts in rapid serial visual presentation at
10 Hz~\citep{grootswagers2022human}. We use the 46 participants retained
by the dataset's quality criteria. The 200 test images have 12 repetitions
each and are distinct from training images, although their concepts can
overlap. We use the released average-referenced, 0.1--100 Hz filtered
recordings at 250 Hz, apply a $[-100,0)$ ms baseline correction, and retain
$[0,1000)$ ms epochs, giving $64\times250$ inputs. Test queries average
all 12 repetitions. At this presentation rate, each epoch includes responses
to adjacent stimuli; the analysis therefore measures decoding within the
rapid presentation stream.

\paragraph{Alljoined-1.6M.}
Alljoined-1.6M contains over 1.6 million trials from 20 participants,
recorded over four sessions per participant with a 32-channel Emotiv Flex 2
system~\citep{xu2025alljoined}. It uses the THINGS-EEG2 stimulus split:
16,540 training images and 200 test images from disjoint concepts.
Images are presented for 100 ms followed by a 100 ms blank interval.
Recordings are sampled at 256 Hz and resampled to 250 Hz, with a 200 ms
pre-stimulus baseline correction and session-wise MVNN estimated from
training epochs. We retain 250 post-stimulus samples per channel
($32\times250$ inputs) and average all retained repetitions of each image;
test queries contain approximately 80 repetitions. Temporal interpretation
follows the timing limitations described in
Appendix~\ref{sec:temporal_window_ablation}.

\paragraph{\texorpdfstring{\protect\redhl{Additional datasets for transfer.}}{Additional datasets for transfer.}}
\redhl{NSD and TVSD test transfer to fMRI and intracortical recordings
(Appendices~\ref{sec:nsd_results} and~\ref{sec:tvsd_results}).}

\paragraph{\texorpdfstring{\protect\redhl{NSD.}}{NSD.}}
\redhl{The Natural Scenes Dataset contains 7T fMRI responses from eight
participants viewing COCO natural scenes, each image shown up to three
times~\citep{allen2022massive}. We use the 1.8-mm GLMsingle single-trial betas
within the \texttt{nsdgeneral} region (12{,}682--17{,}907 voxels per subject),
scaled to percent signal change and z-scored per voxel within each session. The
1{,}000 images shared across participants form the test set and are excluded
from training and validation for all subjects; the remaining images give
172{,}670 training trials of 62{,}610 subject--image pairs. As in
MindEye~\citep{scotti2024reconstructing,scotti2024mindeye2}, evaluation uses
subjects 1, 2, 5, and 7, who completed all 40 sessions, and test queries average
the three repetitions of each image.}

\paragraph{\texorpdfstring{\protect\redhl{TVSD.}}{TVSD.}}
\redhl{The THINGS Ventral Stream Spiking Dataset contains multi-unit activity
from two macaques (monkeys N and F) recorded with Utah arrays in V1, V4, and IT
while they viewed THINGS images~\citep{Papale_2025}. We use the zero-shot split
of \citet{Song_2026}: the 100 test concepts are removed from training, leaving
21{,}048 training images (1{,}754 concepts $\times$ 12 images), each presented
once, and 100 test images, each presented 30 times. Each channel contributes five time
bins. After removing channels without valid training statistics, monkey N has
960 channels (V1 448, V4 256, IT 256) and monkey F 1{,}024 (V1 512, V4 192,
IT 320). Responses are z-scored per recording day, channel, and bin with
statistics of the retained training trials.}

\subsection{Detailed Objectives and Training}
\label{sec:method_details}
This section details BHA's retrieval and reconstruction training objectives.
We use the MSE and BiCLIP definitions in
Eqs.~\eqref{eq:alignment_mse}--\eqref{eq:general_contrastive}; architecture
and numerical settings are given in
Appendices~\ref{sec:brainvit_architecture}--\ref{sec:reconstruction_implementation}.

\subsubsection{Two-stage retrieval training}
\label{sec:detail_group_latent_moe}
\label{sec:brainvit}
EEG pretraining uses paired BiCLIP on same-image trial averages.
Distillation and semantic reconstruction use image-ID positives: when an
image occurs more than once in a minibatch, its positive softmax
probabilities are summed before taking the logarithm in each direction
of Eq.~\eqref{eq:general_contrastive}. Distinct images of the same concept
remain negatives. MEG-specific pretraining is specified in
Appendix~\ref{sec:comprehensive_thingsmeg}.

\paragraph{Stage 1: Pretraining.}
\label{sec:detail_fused_pretraining}
We optimize Eq.~\eqref{eq:fused_visual_anchor} jointly over BrainViT,
source projections $W_k$, fusion map $F$, and contrastive temperature.
The source VFMs remain frozen. Both stages use the optimization settings in
Table~\ref{table:current_eeg_hyperparameters} and the shared data split
specified in Section~\ref{sec:comprehensive_thingseeg2}; the learned fused
representation serves as the distillation target without further adaptation.

\paragraph{Stage 2: Multi-Teacher Distillation.}
\label{sec:detail_multi_teacher_moe}
We keep BrainViT, source VFMs, and the pretrained visual fusion frozen,
and train teacher-specific visual adapters $V_k$ jointly with the retrieval adapter, shared expert,
Group MoEs, outer router, gate, and logit scales. The shared expert matches
the frozen fused feature; each native group matches its adapted teacher
feature. Both alignment and combined-score losses backpropagate into $V_k$.

Native alignment averages BiCLIP over teacher groups, with
$\lambda_k=0.25/K$ for $k\geq1$ and $\lambda_0=1$. The combined-score loss is
$\mathcal{L}_{\mathrm{ret}}=\mathcal{L}_{\mathrm{BiCLIP}}(\mathbf{S}/\tau_{\mathrm{ret}})$,
where $\mathbf{S}$ is defined in Eq.~\eqref{eq:gated_retrieval_score} and
$\tau_{\mathrm{ret}}>0$ is learned. To limit changes to teacher geometry,
visual adapters receive the regularizer
\begin{equation}
    \mathcal{L}_{\mathrm{drift}}
    =\frac{1}{BK}\sum_{i=1}^{B}\sum_{k=1}^{K}
    \left[1-\operatorname{sim}(\tilde{\mathbf{e}}_{ik},
    \operatorname{sg}(\mathbf{e}_{ik}))\right].
    \label{eq:visual_adapter_drift}
\end{equation}
We use $\lambda_{\mathrm{ret}}=1$, $\lambda_{\mathrm{aux}}=0.01$, and
$\lambda_{\mathrm{drift}}=0.1$. Architecture and training settings appear in
Tables~\ref{table:retrieval_architecture_parameters}
and~\ref{table:current_eeg_hyperparameters}.

For internal router logits $G_k(\mathbf{h}'_i)\in\mathbb{R}^{R}$, define
$\mathbf{w}_{ik}=(w_{ik1},\ldots,w_{ikR})^\top$; $\mathbf{1}\in\mathbb{R}^{R}$ is the all-ones vector.
The weights and usage penalty are
\begin{align}
    w_{ikr}
    &=(1-\rho)\operatorname{softmax}
      (G_k(\mathbf{h}'_i)/\tau_g)_r+\rho/R,
      \label{eq:detail_group_router}\\
    \mathcal{L}_{\mathrm{aux}}
    &=\frac{R}{K}\sum_{k=1}^{K}
      \left\|\bar{\mathbf{w}}_k-\mathbf{1}/R\right\|_2^2,
      \qquad \bar{\mathbf{w}}_k=B^{-1}\sum_i\mathbf{w}_{ik}.
      \label{eq:detail_group_balance}
\end{align}
Here $\rho=0.02$ is the uniform-mixing ratio, giving a minimum routing
weight of $\rho/R$; $\tau_g$ controls routing softness.
Only the $R$ routed experts enter this normalization and balance penalty;
the always-active internal shared expert does not. This soft-usage penalty
encourages balanced routing~\citep{shazeer2017outrageously,fedus2022switch}. Scoring and evaluation are specified in
Appendix~\ref{sec:retrieval_implementation}.

\subsubsection{Reconstruction training}
\label{sec:detail_hierarchical_reconstruction}
The semantic and spatial predictors use the pretrained backbone with
separate trainable parameters $\psi_{\mathrm{H}}$ and $\psi_{\mathrm{L}}$:
\begin{equation}
    \hat{\mathbf{e}}_i^{\mathrm{sem}}
    =F_{\mathrm{H}}(\mathbf{X}_i,s_i;\theta,\psi_{\mathrm{H}}),
    \qquad
    \hat{\mathbf{z}}_i^{(L)}
    =F_{\mathrm{L}}(\mathbf{X}_i,s_i;\theta,\psi_{\mathrm{L}}).
    \label{eq:reconstruction_predictions}
\end{equation}
Backbone weights $\theta$ and all generator components remain frozen.
Targets are OpenCLIP features
$\mathbf{e}_i^{\mathrm{sem}}=\phi_{\mathrm{sem}}(\mathbf{I}_i)$ and unscaled
VAE latents $\mathbf{z}_i^{(L)}=E_{\mathrm{VAE}}(\mathbf{I}_i)$, where
$L$ is the finest pyramid level. Outside the pyramid we write
$\hat{\mathbf{z}}_i=\hat{\mathbf{z}}_i^{(L)}$.
Cosine losses average $d_{\cos}(\mathbf{u},\mathbf{v})=1-\operatorname{sim}(\mathbf{u},\mathbf{v})$
over samples, flattening tensor-valued features.

\paragraph{High-level conditions.}
\label{sec:high_reconstruction}
The semantic predictor combines MSE and image-ID BiCLIP on OpenCLIP
features with MSE and cosine distance after the frozen IP-Adapter
projection~\citep{ye2023ip}. Architecture and loss weights are given in
Appendix~\ref{sec:reconstruction_implementation}.

\paragraph{Low-level conditions.}
\label{sec:low_reconstruction}
We first train a coarse VAE-latent predictor, then a conditional pyramid that
refines its spatial resolution. Area downsampling of the finest VAE target
defines $\mathbf{z}^{(p)}$, and bilinear upsampling $\mathcal{U}_{p}$ defines
residual targets
$\mathbf{r}^{(p)}=\mathbf{z}^{(p)}-\mathcal{U}_{p}(\mathbf{z}^{(p-1)})$.
At each refinement level $p=1,\ldots,L$, the predictor produces
\begin{equation}
    \hat{\mathbf{z}}^{(p)}
    =\mathcal{U}_{p}(\hat{\mathbf{z}}^{(p-1)})
     +\gamma_{p}\hat{\mathbf{r}}^{(p)},
    \label{eq:latent_pyramid_prediction}
\end{equation}
where $p=0$ denotes the coarse level and $\gamma_{p}$ is a learned scalar.
Residual supervision is applied
to $\hat{\mathbf{r}}^{(p)}$ before this scaling. Loss coefficients,
target normalization, and the two-phase training schedule are listed in
Appendix~\ref{sec:reconstruction_implementation}.

\paragraph{Frozen-generator inference.}
\label{sec:frozen_generator_inference}
An Euler ancestral sampler uses the semantic condition
$\mathbf{c}_{\mathrm{sem}}=P_{\mathrm{IP}}(\hat{\mathbf{e}}^{\mathrm{sem}})$
and spatial initialization in Eq.~\eqref{eq:reconstruction_mixing_initialization}
\redhl{and the decoding step in \mbox{Section~\ref{sec:hierarchical_reconstruction}}}.
Its noise scale satisfies $\sigma_t^2=(1-\bar\alpha_t)/\bar\alpha_t$,
where $\bar\alpha_t$ is the cumulative signal-retention factor; at
interpolated inference timesteps, $\bar\alpha_t=(1+\sigma_t^2)^{-1}$.
The denoiser receives the normalized Euler latent, giving
\[
    \frac{a\hat{\mathbf{z}}_i+\sigma_t\boldsymbol{\epsilon}_i}
    {\sqrt{1+\sigma_t^2}}
    =\sqrt{\bar\alpha_t}\,a\hat{\mathbf{z}}_i
    +\sqrt{1-\bar\alpha_t}\,\boldsymbol{\epsilon}_i,
    \qquad \boldsymbol{\epsilon}_i\sim\mathcal{N}(\mathbf{0},\mathrm{I}_{d_z}).
\]
Here $d_z$ is the latent dimension. Sampling settings appear in
Appendix~\ref{sec:reconstruction_implementation}.

\subsection{Model Architecture}
\label{sec:brainvit_architecture}
BrainViT combines channel attention with a convolutional readout;
nearest-neighbor mapping places the sensor tokens on a common reference
sensor grid, so one model accepts different input montages.
Table~\ref{table:brainvit_architecture} lists the stages, output shapes,
and parameter counts for a 63-sensor reference grid, $T=250$ time
samples, ten subject embeddings, and a 1,024-dimensional output. The
token count is $N_{\mathrm{tok}}=C+1$ with a subject token and $C$
without one. MEG uses the same architecture with a 271-sensor reference
grid. Subject tokens are enabled only for within-subject models.
\redhl{Module~1 is a per-sensor linear map $\mathbb{R}^{250}\to\mathbb{R}^{250}$
initialized with the DCT-II basis, plus learned sensor-ID embeddings
and, for within-subject models, a learned subject token.
Module~2 is one Transformer layer with four heads, width 250,
feed-forward width 256, dropout 0.25, and 2D RoPE~\citep{su2024roformer} computed from each
subject's montage coordinates.
Module~4 applies four parallel temporal convolutions with 16 filters each
and kernel lengths 5, 15, 25, and 51 samples ($\approx$20--204\,ms at
250\,Hz), concatenated to 64 channels with BatchNorm and ELU, then
adaptive average pooling to 36 time bins.
Module~5 applies a $9\times1$ spatial convolution over neighboring grid
sensors (64 channels), a full-grid spatial convolution that collapses the
remaining sensor axis to one position (40 channels), each with BatchNorm
and ELU, dropout 0.5, and a $1\times1$ convolution.
This temporal-then-spatial convolutional design follows EEG ConvNets~\citep{schirrmeister2017deep}.
Module~6 flattens the $40\times36$ map and applies a linear layer to 1,024
dimensions, a residual GELU--linear block with dropout 0.3, and LayerNorm.}

\begin{table}[!htbp]
\caption{\textbf{BrainViT architecture for THINGS-EEG2.}
Modules 4--6 act on the Transformer output after mapping onto the
reference sensor grid.}
\label{table:brainvit_architecture}
\centering
\small
\renewcommand{\arraystretch}{1.12}
\setlength{\tabcolsep}{5pt}
\begin{tabular}{c l c r}
\toprule
\textbf{Module} & \textbf{Component} & \textbf{Output shape} & \textbf{Params} \\
\midrule
1 & Temporal projection + sensor/subject embeddings & $N_{\mathrm{tok}}\times250$ & 81,500 \\
2 & Channel Transformer + montage RoPE & $N_{\mathrm{tok}}\times250$ & 379,000 \\
3 & Nearest-neighbor sensor mapping & $63\times250$ & 0 \\
4 & Multi-branch temporal conv + average pooling & $64\times63\times36$ & 1,664 \\
5 & Two-step spatial 2D conv + pointwise conv & $40\times1\times36$ & 179,512 \\
6 & Flatten + linear projection + LayerNorm & $1{,}024$ & 2,527,232 \\
7 & Logit scale & scalar & 1 \\
\midrule
\multicolumn{3}{r}{\textbf{Total BrainViT parameters}} & \textbf{3,168,909} \\
\bottomrule
\end{tabular}
\end{table}

\paragraph{Teacher-specific experts and routing.}
The retrieval adapter $A$ uses two residual MLP blocks with bottleneck
width 512; the shared readout $E_0$ uses width 256. Each teacher group has
one always-active expert and four densely routed experts, each an MLP with
bottleneck width 64 and dropout $0.1$. All experts receive the same adapted
brain representation and predict the teacher's native feature space
(1,024 dimensions for OpenCLIP, DINOv2, and MAE; 768 for DFN). The base
projection is the identity except for DFN's learned linear map. Internal
routers share a width-256 stem with separate teacher-specific heads, with
weights defined in Eq.~\eqref{eq:detail_group_router}. Visual adapters are
residual MLPs with bottleneck width 512.
Tables~\ref{table:retrieval_architecture_parameters}
and~\ref{table:current_eeg_hyperparameters} summarize parameter counts
and optimization settings.

\paragraph{Outer teacher weights and residual gate.}
The outer router combines scores across teacher spaces, whereas the
internal routers above combine residual predictions within each space.
Both the outer weights and the gate depend only on the adapted brain query
$\mathbf{h}'_i\in\mathbb{R}^{1024}$:
\begin{align}
\mathbf{a}_i
&=W_{\pi}\operatorname{LN}_{\pi}(\mathbf{h}'_i+\mathbf{t}_0)
+\mathbf{b}_{\pi},
&\pi_{ik}&=\frac{\exp(a_{ik})}{\sum_{m=1}^{K}\exp(a_{im})},
\label{eq:outer_teacher_router}\\
g_i&=\operatorname{sigmoid}(\mathbf{w}_g^\top\mathbf{h}'_i+b_g).
\label{eq:outer_residual_gate}
\end{align}
Here $K=4$, $W_{\pi}\in\mathbb{R}^{4\times1024}$,
$\mathbf{b}_{\pi}\in\mathbb{R}^{4}$, and
$\mathbf{w}_g,\mathbf{t}_0\in\mathbb{R}^{1024}$.
The learned embedding $\mathbf{t}_0$ is shared across retrieval queries.
Both are trained through $\mathcal{L}_{\mathrm{ret}}$ with the backbone and
fused visual target frozen.

\begin{table}[!htbp]
\caption{\textbf{Retrieval architecture and parameter counts.}
Counts are for the THINGS-EEG2 within-subject model.
Frozen source VFMs and the image generator are excluded. Stage~2
freezes BrainViT and visual fusion and trains the remaining modules.}
\label{table:retrieval_architecture_parameters}
\centering\small
\setlength{\tabcolsep}{5pt}
\begin{tabular}{lrr}
\toprule
Component & Trainable in Stage~2 & Frozen in Stage~2 \\
\midrule
BrainViT & 0 & 3,168,909 \\
Visual fusion & 0 & 6,035,456 \\
Retrieval adapter $A$ & 2,101,250 & 0 \\
Shared readout $E_0$ & 528,385 & 0 \\
Group MoEs: $4\times(1+4)$ rank-64 experts & 2,580,500 & 0 \\
Group MoEs: internal routing & 268,560 & 0 \\
Group MoEs: DFN base projection & 786,432 & 0 \\
Four visual adapters & 3,939,844 & 0 \\
Outer teacher router and scalar gate & 9,221 & 0 \\
Retrieval logit scales & 6 & 0 \\
\midrule
Total & 10,214,198 & 9,204,365 \\
\bottomrule
\end{tabular}
\end{table}

\subsection{Architecture Ablations}
\label{sec:readout_ablations}

\paragraph{BrainViT readout search.}
We treat the readout (Modules~4--6) as a design choice and compare five
candidates on the same Modules~1--3: flattening all sensor tokens into the
Module~6 head; the Module~4 temporal convolution followed by flattening the
full sensor--time map into the head; 16 learnable queries that
cross-attend to the sensor tokens (five heads), flattened into the head;
the Module~5 spatial convolution alone, with a pointwise $1\times1$
convolution in place of the temporal branches; and the full Module~4 plus
Module~5 readout (BrainViT).
Each encoder and its visual fusion are jointly trained under the Stage~1
settings and selected by validation loss; retrieval compares their
embeddings directly, without Stage~2 or CSLS. Evaluation uses ten subjects
or LOSO folds, 63 sensors, 80-trial averages, and 200 candidates; LOSO
excludes the target subject from training and selection and disables
subject tokens.
Mean Top-1 spans $16.30$ percentage points within subjects and $4.95$
points under LOSO (Table~\ref{table:brainvit_readout_comparison}), so
retrieval is more sensitive to the readout within subjects.
The spatial convolution gives the highest mean Top-1 in both settings;
the temporal convolution adds no further gain.

\begin{table}[!htbp]
\caption{\textbf{BrainViT readout search on THINGS-EEG2.}
Five readouts on the same channel Transformer under a common training
protocol. Raw retrieval accuracy (\%), subject mean $\pm$ SD;
encoder parameters in millions.}
\label{table:brainvit_readout_comparison}
\centering\small\setlength{\tabcolsep}{4pt}
\resizebox{\fitwidth}{!}{%
\begin{tabular}{lrcccc}
\toprule
& & \multicolumn{2}{c}{Within-subject (Raw)} & \multicolumn{2}{c}{LOSO (Raw)} \\
\cmidrule(lr){3-4}\cmidrule(lr){5-6}
Encoder readout & Params & Top-1 & Top-5 & Top-1 & Top-5 \\
\midrule
Flatten + MLP & 17.641 & $66.80\pm7.28$ & $92.30\pm4.36$ & $\underline{23.35\pm8.32}$ & $\boldsymbol{53.50\pm11.58}$ \\
Temporal conv  & 150.150 & $70.30\pm10.78$ & $92.75\pm5.56$ & $19.60\pm4.85$ & $47.35\pm8.17$ \\
Learnable-query ($Q=16$) & 6.117 & $79.10\pm8.31$ & $96.15\pm2.39$ & $23.25\pm7.36$ & $51.25\pm11.14$ \\
Spatial conv & 3.167 & $\boldsymbol{83.10\pm5.37}$ & $\boldsymbol{97.35\pm2.19}$ & $\boldsymbol{24.55\pm7.88}$ & $\underline{52.30\pm10.49}$ \\
Temporal + spatial conv & 3.169 & $\underline{80.90\pm7.03}$ & $\underline{96.25\pm2.67}$ & $22.50\pm7.44$ & $51.60\pm8.60$ \\
\bottomrule
\end{tabular}}
\end{table}

\paragraph{\texorpdfstring{\protect\redhl{Teacher-specific readout search.}}{Teacher-specific readout search.}}
\redhl{BHA treats the teacher-specific readout as an interchangeable component.
To test whether the Group MoE, its routing, or its capacity is responsible
for the gain, we train alternative readouts on the same frozen BrainViT and
visual fusion, keeping the retrieval adapter $A$, the shared readout $E_0$,
the visual adapters, and the Stage~2 settings, and score each trained model
under four rules: the shared space alone, uniform weights over the teacher
spaces ($g_i=1$, $\pi_{ik}=1/K$), learned weights (the query averages
of $g_i$ and $\pi_{ik}$), and BHA's query-dependent weights
(Table~\ref{table:moe_readout_ablation}a). Readout size is varied through
the expert rank $r$, with MLP and low-rank linear readouts matched to the
Group MoE's parameter count at every $r$
(Table~\ref{table:moe_readout_ablation}b). All readouts reach the same
accuracy to within one point, and learned weights recover the full
gain over the shared space, so the gain comes from composing
teacher-specific scores rather than from the MoE, its routing, or its size.
The Group MoE remains the default. The learned weights are consistent across
subjects and seeds: the gate stays at 0.88--0.97, and DFN-B receives
0.89--0.98 of the teacher weight, matching its contribution, since removing its
score lowers Top-1 by 4.0--11.2 points while removing any other costs at most
0.3 (Table~\ref{table:moe_readout_ablation}c). The weights vary little across
queries (per-query SD of the DFN-B weight 0.03--0.11), so they can be fixed
after training at little cost.}

\begin{table}[!htbp]\redhlon
\caption{\textbf{Teacher-specific readout search within BHA on THINGS-EEG2.}
Within-subject Raw Top-1 (\%); means over three seeds, ten subjects each. All readouts share the frozen BrainViT,
visual fusion, retrieval adapter, shared readout, and Stage~2 settings.
(a) Scoring rules applied to the same trained model; \emph{Params}:
teacher-specific readout parameters. (b) Query-dependent Raw Top-1 as
readout size varies with the expert rank $r$; MLP and low-rank linear
readouts are matched to the Group MoE's parameter count at each $r$.
(c) Learned composition weights of the default Group MoE for each subject:
gate $g$ and teacher weights $\pi_k$, averaged over test queries;
\emph{w/o DFN-B}: change in Top-1 when the DFN-B score is removed at test time.
Gray marks the default.}
\label{table:moe_readout_ablation}
\centering\footnotesize\setlength{\tabcolsep}{4pt}
(a) Scoring rules\\[2pt]
\begin{tabular}{@{}lrcccc@{}}
\toprule
Readout & Params & Shared only & Uniform & Learned & Query-dependent \\
\midrule
\rowcolor{gray!10}
Group MoE ($r=64$, default) & 2.85M & 83.3 & 83.6 & 86.4 & 86.5 \\
Group MoE, sparse top-1 routing & 2.85M & 83.3 & 83.7 & 86.5 & 86.5 \\
MLP, parameter-matched & 2.85M & 83.4 & 83.6 & 86.7 & 86.6 \\
Linear & 3.93M & 83.4 & 84.6 & 86.6 & 86.6 \\
\bottomrule
\end{tabular}\\[6pt]
(b) Readout size\\[2pt]
\begin{tabular}{@{}lrccc@{}}
\toprule
Expert rank $r$ & Params & Group MoE & MLP & Low-rank linear \\
\midrule
4 & 0.47M & 86.8 & 86.5 & 86.7 \\
16 & 0.94M & 86.5 & 86.6 & 86.6 \\
\rowcolor{gray!10}
64 (default) & 2.85M & 86.5 & 86.6 & 86.5 \\
256 & 10.5M & 86.7 & 85.9 & 86.4 \\
\bottomrule
\end{tabular}\\[6pt]
(c) Learned composition weights\\[2pt]
\begin{tabular}{@{}lcccccccc@{}}
\toprule
Subject & Shared & BHA & $g$ & \multicolumn{4}{c}{Teacher weight $\pi_k$} & w/o \\
\cmidrule(lr){5-8}
 & & & & OpenCLIP-H & DINOv2-L & MAE-L & DFN-B & DFN-B \\
\midrule
1 & 87.0 & 92.3 & 0.94 & 0.02 & 0.00 & 0.00 & 0.98 & $-10.5$ \\
2 & 82.7 & 86.0 & 0.91 & 0.02 & 0.00 & 0.00 & 0.97 & $-9.2$ \\
3 & 85.7 & 87.8 & 0.95 & 0.03 & 0.00 & 0.00 & 0.96 & $-8.0$ \\
4 & 79.8 & 84.2 & 0.93 & 0.10 & 0.00 & 0.00 & 0.89 & $-11.2$ \\
5 & 73.8 & 77.0 & 0.90 & 0.03 & 0.00 & 0.00 & 0.97 & $-10.8$ \\
6 & 86.7 & 89.5 & 0.93 & 0.02 & 0.00 & 0.00 & 0.98 & $-9.8$ \\
7 & 85.8 & 88.2 & 0.88 & 0.04 & 0.02 & 0.02 & 0.93 & $-7.2$ \\
8 & 86.5 & 90.2 & 0.97 & 0.04 & 0.00 & 0.00 & 0.96 & $-8.3$ \\
9 & 78.5 & 82.8 & 0.91 & 0.02 & 0.00 & 0.00 & 0.98 & $-9.5$ \\
10 & 86.3 & 86.8 & 0.97 & 0.03 & 0.00 & 0.00 & 0.96 & $-4.0$ \\
\midrule
Mean & 83.3 & 86.5 & 0.93 & 0.03 & 0.00 & 0.00 & 0.96 & $-8.9$ \\
\bottomrule
\end{tabular}
\end{table}

\subsection{Retrieval Implementation}
\label{sec:retrieval_implementation}
EEG visual features are extracted with deterministic OpenCLIP preprocessing
and frozen VFMs in evaluation mode.

\paragraph{Model selection and evaluation.}
Stage~1 selects the minimum validation loss; Stage~2 selects the maximum
validation gated Raw Top-1, with Top-5 as the tie-breaker. Both use the
image-disjoint split in Section~\ref{sec:comprehensive_thingseeg2}.
We evaluate 200-way retrieval and average accuracies across ten EEG or
four MEG subjects, using 80- and 12-trial test averages, respectively.
Within-subject retrieval uses Stage~2 gated scores. EEG and MEG LOSO
train on the remaining subjects and directly use
$S_{ij}^{\mathrm{LOSO}}=\operatorname{sim}(\mathbf{h}_i,\mathbf{e}_{j0})$,
bypassing all Stage~2 modules. Raw scores queries independently;
CSLS rescores the same predictions using other test queries.
The extended EEG LOSO table also reports image-to-brain retrieval.

\paragraph{CSLS and reference information.}
For neural queries indexed by a reference set $\mathcal{R}$ and an image
gallery $\mathcal{G}$, CSLS~\citep{Conneau_2017_171004087} adjusts the raw score
matrix $\mathbf{Q}$, where $Q_{ij}=S_{ij}$ for within-subject retrieval and
$Q_{ij}=S_{ij}^{\mathrm{LOSO}}$ for LOSO. For a neighborhood size $k\geq1$, we use $k_{\mathrm{eff}}=\min(k,|\mathcal{R}|,|\mathcal{G}|)$.
Let $\mathcal{N}^{\mathcal{G}}_{k_{\mathrm{eff}}}(i)$ index the
$k_{\mathrm{eff}}$ highest-scoring images for query $i$, and
$\mathcal{N}^{\mathcal{R}}_{k_{\mathrm{eff}}}(j)$ index the
$k_{\mathrm{eff}}$ highest-scoring neural references for image $j$. Then
\begin{equation}
    \begin{aligned}
    r_i(k)&=\frac{1}{k_{\mathrm{eff}}}
        \sum_{j'\in\mathcal{N}^{\mathcal{G}}_{k_{\mathrm{eff}}}(i)}Q_{ij'},
    & c_j(k)&=\frac{1}{k_{\mathrm{eff}}}
        \sum_{i'\in\mathcal{N}^{\mathcal{R}}_{k_{\mathrm{eff}}}(j)}Q_{i'j},\\
    \widetilde Q_{ij}(k)&=2Q_{ij}-r_i(k)-c_j(k).
    \end{aligned}
    \label{eq:csls_retrieval_score}
\end{equation}

Neighborhoods use uncorrected scores. Main CSLS results use all 200 test
queries and 200 gallery images, so $k_{\mathrm{eff}}=k$. The reference-size
ablation truncates both neighborhoods when fewer queries are available.
For neural-to-image retrieval, $r_i(k)$ is constant across candidates;
for the reverse direction, $c_j(k)$ is constant. CSLS therefore changes
rankings through the opposite-side correction. It uses no labels or
parameter updates, but access to other test queries makes it transductive.

\paragraph{Visual targets.}
Our four retrieval teachers are OpenCLIP ViT-H/14 global features~\citep{cherti2023reproducible},
MAE ViT-L~\citep{he2022masked}, DINOv2-L~\citep{oquab2024dinov}, and
DFN ViT-B~\citep{fang2024data}. The latter three use intermediate features
MAE-18, DINOv2-12, and DFN-6, whose suffixes denote zero-based Transformer
block indices. DFN-9 is used only for the layer comparison; the main
fusion model uses DFN-6.
OpenCLIP, MAE-18, and DINOv2-12 provide 1,024-dimensional features;
DFN-6 provides 768-dimensional features. Their projections, normalization,
and fusion follow Eq.~\eqref{eq:fused_visual_anchor}, producing
$\mathbf{e}_{i0}\in\mathbb{R}^{d_0}$ with $d_0=1024$.
The individual teacher targets are adapted only during distillation.
Reconstruction uses the native OpenCLIP and VAE spaces described in
Appendix~\ref{sec:reconstruction_implementation}.

\subsection{Reconstruction Implementation}
\label{sec:reconstruction_implementation}
Reconstruction uses the same BrainViT encoder and convolutional readout
architecture as primary retrieval.
Its base weights and all generator components remain frozen; only the
reconstruction predictors are trained.

\paragraph{Semantic and spatial prediction.}
The semantic branch uses rank-64 LoRA~\citep{hu2022lora} and a two-block residual calibrator
with bottleneck width 256 to predict a 1,024-dimensional OpenCLIP ViT-H/14
image embedding.
The frozen IP-Adapter~\citep{ye2023ip} maps it to four 2,048-dimensional
conditioning tokens, following the CLIP-conditioning approach of
ENIGMA~\citep{kneeland2026enigma}.

The spatial branch uses the 17 posterior EEG sensor tokens to predict
an unscaled $4\times16\times16$ VAE latent with a convolutional readout.
A residual convolutional adapter maps this prediction to the pyramid's
coarse target. Cross-attention from 256 learned spatial queries to the
EEG tokens supplies a 64-channel conditioning map. Two conditioned
residual stages then refine the latent to $4\times32\times32$ and
$4\times64\times64$, following Eq.~\eqref{eq:latent_pyramid_prediction}.
The coarse predictor targets the VAE posterior mean of a $128\times128$
image; pyramid targets are downsampled from its $512\times512$ encoding.

\paragraph{Training.}
Semantic losses weight global-feature MSE and BiCLIP by $0.9$ and $0.1$,
and IP-token MSE and cosine distance by $0.1$ and $0.01$; BiCLIP temperature
is $0.07$. Coarse MSE and cosine weights are $1$ and $0.1$.
Pyramid MSE weights are $(0.5,0.5,1)$ for spatial sizes $(16,32,64)$,
with $0.2$ for each unscaled residual and $0.05$ for finest-level cosine
distance. Pyramid MSE errors are divided by target-channel training
standard deviations. We first train the pyramid with the coarse predictor
frozen, then jointly refine the spatial path with an additional
unstandardized coarse MSE term (weight $0.5$). AdamW uses learning rates $3\times10^{-4}$ for
the pyramid and $10^{-5}$ for the coarse predictor, with weight decay
$10^{-4}$. Training and validation use four-trial EEG averages.

\paragraph{Generation.}
We generate $512\times512$ images with frozen SDXL-Turbo~\citep{podell2024sdxl,sauer2024adversarial}
and a ViT-H IP-Adapter, using Euler ancestral sampling, IP-Adapter scale
$1.5$, and classifier-free guidance scale $0$. The main setting is
$\alpha=0.75$: three denoising updates from a five-step schedule.
Semantic embeddings enter the IP-Adapter without normalization; spatial
latents use VAE scale $a=0.13025$ before noise addition. Every stimulus
uses the same generic quality prompt, without image-specific text.
Results average three generations per image. The guided variant uses
the same models and sampling settings with the additional updates below.

\paragraph{Training-free guidance.}
\label{sec:vae_guidance_details}
For noisy latent $\mathbf{z}_t$, Euler noise level $\sigma_t$, and denoiser
condition $\mathbf{c}_{\mathrm{sem}}$ (the same text conditioning omitted), let
$\tilde{\mathbf{z}}_t=\mathbf{z}_t/\sqrt{1+\sigma_t^2}$. The estimated clean
latent and its consistency loss are
\begin{equation}
    \hat{\mathbf{z}}_{0\mid t}=\mathbf{z}_t-\sigma_t\epsilon_{\mathrm{den}}(\tilde{\mathbf{z}}_t,t,\mathbf{c}_{\mathrm{sem}}),
    \qquad
    \mathcal{L}_{\mathrm{guide}}(\mathbf{z}_t)
    =\frac{1}{d_z}\left\|\frac{\hat{\mathbf{z}}_{0\mid t}}{a}-\hat{\mathbf{z}}\right\|_2^2.
\label{eq:vae_gradient_guidance_loss}
\end{equation}
Here $\epsilon_{\mathrm{den}}$ is the frozen denoiser, $a=0.13025$ is the VAE scaling
factor, and $d_z=4\cdot64\cdot64$; latent tensors are vectorized in the loss
and gradient normalization. The reference $\hat{\mathbf{z}}$ is the
unscaled spatial prediction. An RMS-normalized gradient step refines the
noisy latent:
\begin{equation}
\begin{aligned}
    \mathbf{g}_t&=\nabla_{\mathbf{z}_t}\mathcal{L}_{\mathrm{guide}}(\mathbf{z}_t),
    &\operatorname{RMS}(\mathbf{g}_t)&=\sqrt{\frac{1}{d_z}\sum_{j=1}^{d_z}(\mathbf{g}_t)_j^2},\\
    \mathbf{z}_t^{+}&=\mathbf{z}_t-\eta_{\mathrm{g}}\sigma_t
        \frac{\mathbf{g}_t}{\max\{\operatorname{RMS}(\mathbf{g}_t),\varepsilon\}}.
\end{aligned}
\label{eq:vae_gradient_guidance_update}
\end{equation}
We use $\eta_{\mathrm{g}}=0.02$ and $\varepsilon=10^{-12}$, applying one refinement
before each of the final two denoising updates, followed by a new noise
prediction and an Euler ancestral step. Model weights and predicted
conditions remain unchanged. The remaining generation settings are those above.

Image-conditioned and permuted-condition controls are described in
Appendix~\ref{sec:reconstruction_ceilings}.

\section{Additional Experimental Results}
\label{sec:add_experimental_results}

\subsection{Comprehensive BHA Results}
\label{sec:comprehensive_bha_results}
We examine how retrieval varies across subjects and evaluation settings.
The subject-wise tables report individual accuracies and aggregate scores (Avg); ``--'' denotes unreported results. Baseline aggregates are reproduced from their reporting sources, while BHA averages subjects equally.
Across Raw and CSLS, BHA uses identical model outputs and trial averages.
Raw scores queries independently, whereas CSLS-2/10 use all 200 test
queries as references (Appendix~\ref{sec:retrieval_implementation}).

\paragraph{Baseline aggregation discrepancies.}
A dagger marks published aggregates that differ from the arithmetic means of the displayed subject scores beyond rounding. For within-subject EEG, the ATM-S row reproduced from MBP's comparison reports $28.5/60.4$, whereas its subject entries average $27.10/58.08$; MBP reports Top-5 $94.6$ versus $94.91$ from its entries~\citep{zheng2026learning}. For MEG LOSO, HyFI reports $3.2/11.5$ versus $2.875/10.775$ from its entries~\citep{jo2026hyfi}. We retain these published aggregates in the main and supplementary comparisons; without the underlying runs, we cannot resolve the source inconsistencies. CC-All subject values follow its original report~\citep{zhang2024cognitioncapturer}.

\subsubsection{Retrieval Performance in THINGS-EEG2}
\label{sec:comprehensive_thingseeg2}

\paragraph{Training settings.}
For within-subject retrieval, we fit the two-stage model separately for
each EEG participant. We reserve one image per concept (1,654 images)
for validation and use 14,886 images for fitting, keeping all four repeats
of each image together. Both stages use this split.
Table~\ref{table:current_eeg_hyperparameters} lists the key training and
inference hyperparameters.

\begin{table}[!hbp]
\centering
\caption{\textbf{THINGS-EEG2 retrieval hyperparameters.}
BS: batch size; GAS: gradient accumulation steps.}
\label{table:current_eeg_hyperparameters}
\small
\setlength{\tabcolsep}{4pt}
\renewcommand{\arraystretch}{1.0}
\begin{tabular}{@{}>{\raggedright\arraybackslash}p{0.40\linewidth}
                  >{\raggedright\arraybackslash}p{0.19\linewidth}
                  >{\raggedright\arraybackslash}p{0.34\linewidth}@{}}
\toprule
\textbf{Hyperparameter} & \textbf{Stage 1} & \textbf{Stage 2} \\
\midrule
Optimizer & \multicolumn{2}{l@{}}{AdamW; $\beta_1=0.9$, $\beta_2=0.999$, $\epsilon=10^{-8}$} \\
BS / GAS & \multicolumn{2}{l@{}}{$1024$ / $1$} \\
Learning rate / schedule & \multicolumn{2}{l@{}}{$3\times10^{-4}$; constant, no warmup} \\
Weight decay & $10^{-2}$ & $10^{-4}$ \\
Maximum epochs / patience & $100$ / $10$ & $160$ / $20$ \\
\midrule
Channel sampling & $63$ or $17$; $p=0.5$ each & -- \\
Expert dropout & -- & $0.1$ \\
Router uniform-mixing ratio & -- & $0.02$ \\
Internal routing schedule & -- & Uniform in epoch 1; temperature $1.5\!\to\!0.8$ over 40 epochs \\
Native / shared / gated loss weights & -- & $0.25$ (teacher mean) / $1$ / $1$ \\
Visual drift / router balance weights & -- & $0.1$ / $0.01$ \\
\midrule
CSLS: neighbors / reference queries & \multicolumn{2}{l@{}}{$k\in\{2,10\}$ / $M=200$} \\
\bottomrule
\end{tabular}
\end{table}

\paragraph{Within-subject retrieval.}
Within-subject retrieval uses the Stage-2 gated readout. The subject-wise
comparison shows strong within-subject decoding, with CSLS further
improving retrieval from the same predictions
(Table~\ref{table:retrieval_in_subject}).

\begin{table}[!htbp]
    \caption{\textbf{Within-subject retrieval on THINGS-EEG2.} 200-way Top-1 and Top-5 accuracy (\%). Light gray marks BHA rows.}
    \label{table:retrieval_in_subject}
    \centering
    \resizebox{\fitwidth}{!}{%
    \setlength{\tabcolsep}{2pt}
    \begin{tabular}{l *{22}{c}}
    \toprule
    \multirow{2}{*}{\textbf{Method}}
      & \multicolumn{2}{c}{\textbf{Sub1}}
      & \multicolumn{2}{c}{\textbf{Sub2}}
      & \multicolumn{2}{c}{\textbf{Sub3}}
      & \multicolumn{2}{c}{\textbf{Sub4}}
      & \multicolumn{2}{c}{\textbf{Sub5}}
      & \multicolumn{2}{c}{\textbf{Sub6}}
      & \multicolumn{2}{c}{\textbf{Sub7}}
      & \multicolumn{2}{c}{\textbf{Sub8}}
      & \multicolumn{2}{c}{\textbf{Sub9}}
      & \multicolumn{2}{c}{\textbf{Sub10}}
      & \multicolumn{2}{c}{\textbf{Avg}} \\
    \cmidrule(lr){2-3}  \cmidrule(lr){4-5}  \cmidrule(lr){6-7}
    \cmidrule(lr){8-9}  \cmidrule(lr){10-11} \cmidrule(lr){12-13}
    \cmidrule(lr){14-15} \cmidrule(lr){16-17} \cmidrule(lr){18-19}
    \cmidrule(lr){20-21} \cmidrule(lr){22-23}
    & \footnotesize{Top-1} & \footnotesize{Top-5}
    & \footnotesize{Top-1} & \footnotesize{Top-5}
    & \footnotesize{Top-1} & \footnotesize{Top-5}
    & \footnotesize{Top-1} & \footnotesize{Top-5}
    & \footnotesize{Top-1} & \footnotesize{Top-5}
    & \footnotesize{Top-1} & \footnotesize{Top-5}
    & \footnotesize{Top-1} & \footnotesize{Top-5}
    & \footnotesize{Top-1} & \footnotesize{Top-5}
    & \footnotesize{Top-1} & \footnotesize{Top-5}
    & \footnotesize{Top-1} & \footnotesize{Top-5}
    & \footnotesize{Top-1} & \footnotesize{Top-5} \\
    \midrule
    BraVL~\citep{du2023decoding}
      & 6.1 & 17.9 & 4.9 & 14.9 & 5.6 & 17.4 & 5.0 & 15.1 & 4.0 & 13.4
      & 6.0 & 18.2 & 6.5 & 20.4 & 8.8 & 23.7 & 4.3 & 14.0 & 7.0 & 19.7
      & 5.8 & 17.5 \\
    NICE~\citep{song2024decoding}
      & 13.2 & 39.5 & 13.5 & 40.3 & 14.5 & 42.7 & 20.6 & 52.7 & 10.1 & 31.5
      & 16.5 & 44.0 & 17.0 & 42.1 & 22.9 & 56.1 & 15.4 & 41.6 & 17.4 & 45.8
      & 16.1 & 43.6 \\
    ATM-S~\citep{li2024visual}
      & 25.6 & 60.4 & 22.0 & 54.5 & 25.0 & 62.4 & 31.4 & 60.9 & 12.9 & 43.0
      & 21.3 & 51.1 & 30.5 & 61.5 & 38.8 & 72.0 & 34.4 & 51.5 & 29.1 & 63.5
      & $28.5^\dagger$ & $60.4^\dagger$ \\
    CC-All~\citep{zhang2024cognitioncapturer}
      & 31.4 & 79.7 & 31.4 & 77.8 & 38.2 & 85.7 & 40.4 & 85.8 & 24.4 & 66.3
      & 34.8 & 78.8 & 34.7 & 81.0 & 48.1 & 88.6 & 37.4 & 79.4 & 35.6 & 79.3
      & 35.6 & 80.2 \\
    UBP~\citep{wu2025bridging}
      & 41.2 & 70.5 & 51.2 & 80.9 & 51.2 & 82.0 & 51.1 & 76.9 & 42.2 & 72.8
      & 57.5 & 83.5 & 49.0 & 79.9 & 58.6 & 85.8 & 45.1 & 76.2 & 61.5 & 88.2
      & 50.9 & 79.7 \\
    NeuroBridge~\citep{zhang2026neurobridge}
      & 50.0 & 77.6 & 63.2 & 90.6 & 61.6 & 91.1 & 61.4 & 90.0 & 54.8 & 85.0
      & 69.7 & 92.9 & 62.7 & 88.8 & 71.2 & 95.1 & 64.0 & 91.0 & 73.6 & 97.1
      & 63.2 & 89.9 \\
    MBP~\citep{zheng2026learning}
      & 64.3 & 88.8 & 76.3 & 95.3 & 74.0 & 95.0 & 67.0 & 91.8 & 68.0 & 91.5
      & 81.5 & 96.3 & 76.8 & 96.8 & 84.8 & 98.5 & 76.8 & 95.8 & 87.3 & 99.3
      & 75.7 & $94.6^\dagger$ \\
    DMSA~\citep{du2026deep}
      & 75.0 & 94.3 & 87.5 & 98.9 & 83.2 & 98.2 & 79.5 & 96.1 & 74.6 & 96.4
      & 89.9 & \underline{99.3} & 78.5 & 97.3 & 86.9 & 99.4 & 81.3 & 97.8 & 89.3 & 99.1
      & 82.6 & 97.7 \\      

    \midrule
    \rowcolor{gray!10}
    \textbf{BHA (Ours)} / Raw & 90.0 & \underline{99.5} & 85.5 & 97.5 & 88.0 & \underline{99.5} & 83.5 & 97.5 & 77.0 & 96.5 & 92.5 & 98.5 & 88.5 & 98.5 & 91.0 & \underline{99.5} & 85.0 & \underline{98.0} & 89.0 & \underline{99.5} & 87.00 & 98.45 \\
    \rowcolor{gray!10}
    \textbf{BHA (Ours)} / CSLS-2 & \textbf{98.0} & \textbf{100.0} & \textbf{92.5} & \textbf{100.0} & \textbf{95.0} & \textbf{100.0} & \textbf{91.0} & \textbf{99.5} & \textbf{85.0} & \textbf{99.5} & \textbf{96.5} & \textbf{99.5} & \textbf{96.5} & \textbf{99.5} & \textbf{96.5} & \textbf{100.0} & \textbf{91.0} & \textbf{99.0} & \textbf{97.0} & \textbf{100.0} & \textbf{93.90} & \textbf{99.70} \\
    \rowcolor{gray!10}
    \textbf{BHA (Ours)} / CSLS-10 & \underline{93.0} & \textbf{100.0} & \underline{90.5} & \underline{99.5} & \underline{93.0} & \underline{99.5} & \underline{88.5} & \underline{98.5} & \underline{82.5} & \underline{99.0} & \underline{94.0} & 99.0 & \underline{94.5} & \underline{99.0} & \underline{95.0} & \textbf{100.0} & \underline{87.5} & \textbf{99.0} & \underline{95.0} & \underline{99.5} & \underline{91.35} & \underline{99.30} \\
    \bottomrule
    \end{tabular}%
    }
\end{table}

\paragraph{LOSO retrieval comparison.}
To test whether shared visual supervision supports transfer without
target-subject fitting, we train on nine subjects and evaluate the held-out subject in each of ten
folds, excluding target-subject data from training and model selection.
Retrieval uses Stage-1 BrainViT without subject tokens
(Appendix~\ref{sec:retrieval_implementation}). The shared representation supports transfer
to unseen subjects, although accuracy remains below
within-subject retrieval. Table~\ref{table:retrieval_loso_comparison} compares
methods; Table~\ref{table:retrieval_loso_persubject} reports both retrieval
directions at Top-1 and Top-5. ATM-S subject-wise values follow the comparison reported by NeuroBridge~\citep{zhang2026neurobridge}; their means ($11.9/33.8$) are also used in Table~\ref{table:retrieval_comparison}.

\begin{table}[!htbp]
  \caption{\textbf{LOSO retrieval on THINGS-EEG2.} 200-way Top-1 and Top-5 accuracy (\%). Light gray marks BHA rows.}
  \label{table:retrieval_loso_comparison}
  \centering
  \resizebox{\fitwidth}{!}{%
  \setlength{\tabcolsep}{2pt}
  \begin{tabular}{l *{22}{c}}
  \toprule
  \multirow{2}{*}{\textbf{Method}}
    & \multicolumn{2}{c}{\textbf{Sub1}}
    & \multicolumn{2}{c}{\textbf{Sub2}}
    & \multicolumn{2}{c}{\textbf{Sub3}}
    & \multicolumn{2}{c}{\textbf{Sub4}}
    & \multicolumn{2}{c}{\textbf{Sub5}}
    & \multicolumn{2}{c}{\textbf{Sub6}}
    & \multicolumn{2}{c}{\textbf{Sub7}}
    & \multicolumn{2}{c}{\textbf{Sub8}}
    & \multicolumn{2}{c}{\textbf{Sub9}}
    & \multicolumn{2}{c}{\textbf{Sub10}}
    & \multicolumn{2}{c}{\textbf{Avg}} \\
  \cmidrule(lr){2-3}  \cmidrule(lr){4-5}  \cmidrule(lr){6-7}
  \cmidrule(lr){8-9}  \cmidrule(lr){10-11} \cmidrule(lr){12-13}
  \cmidrule(lr){14-15} \cmidrule(lr){16-17} \cmidrule(lr){18-19}
  \cmidrule(lr){20-21} \cmidrule(lr){22-23}
  & \footnotesize{Top-1} & \footnotesize{Top-5}
  & \footnotesize{Top-1} & \footnotesize{Top-5}
  & \footnotesize{Top-1} & \footnotesize{Top-5}
  & \footnotesize{Top-1} & \footnotesize{Top-5}
  & \footnotesize{Top-1} & \footnotesize{Top-5}
  & \footnotesize{Top-1} & \footnotesize{Top-5}
  & \footnotesize{Top-1} & \footnotesize{Top-5}
  & \footnotesize{Top-1} & \footnotesize{Top-5}
  & \footnotesize{Top-1} & \footnotesize{Top-5}
  & \footnotesize{Top-1} & \footnotesize{Top-5}
  & \footnotesize{Top-1} & \footnotesize{Top-5} \\
  \midrule
  NICE~\citep{song2024decoding}
    & 7.6 & 22.8 & 5.9 & 20.5 & 6.0 & 22.3 & 6.3 & 20.7 & 4.4 & 18.3
    & 5.6 & 22.2 & 5.6 & 19.7 & 6.3 & 22.0 & 5.7 & 17.6 & 8.4 & 28.3
    & 6.2 & 21.4 \\
  ATM-S~\citep{li2024visual}
    & 10.5 & 26.8 & 7.1 & 24.8 & 11.9 & 33.8 & 14.7 & 39.4 & 7.0 & 23.9
    & 11.1 & 35.8 & 16.1 & 43.5 & 15.0 & 40.3 & 4.9 & 22.7 & 20.5 & 46.5
    & 11.9 & 33.8 \\
  Neural-MCRL~\citep{li2024neuralmcrl}
    & 13.0 & 31.5 & 12.0 & 30.5 & 14.5 & 35.5 & 12.5 & 35.5 & 11.5 & 29.0
    & 13.5 & 35.5 & 14.0 & 36.0 & 18.5 & 38.5 & 13.5 & 32.5 & 17.0 & 39.0
    & 14.0 & 34.3 \\
  UBP~\citep{wu2025bridging}
    & 11.5 & 29.7 & 15.5 & 40.0 & 9.8 & 27.0 & 13.0 & 32.3 & 8.8 & 33.8
    & 11.7 & 31.0 & 10.2 & 23.8 & 12.2 & 32.2 & 15.5 & 40.5 & 16.0 & 43.5
    & 12.4 & 33.4 \\
  NeuroBridge~\citep{zhang2026neurobridge}
    & 23.2 & 52.4 & 21.2 & 49.3 & 13.2 & 36.5 & 17.0 & 45.3 & 14.5 & 37.7
    & 25.0 & 55.0 & 15.3 & 45.1 & 20.1 & 44.9 & 13.7 & 36.5 & 27.2 & 56.3
    & 19.0 & 45.9 \\
  DMSA~\citep{du2026deep}
    & 23.5 & 53.2 & 30.6 & 60.4 & 10.0 & 28.1 & 19.5 & 48.3 & 18.1 & 45.2
    & 22.7 & 49.8 & 18.6 & 46.0 & 17.3 & 46.1 & 23.0 & 54.8 & \underline{34.4} & 62.0
    & 21.8 & 49.4 \\
  \midrule
    \rowcolor{gray!10}
    \textbf{BHA (Ours)} / Raw & 35.0 & 64.5 & 29.0 & 57.5 & 12.5 & 38.0 & 21.0 & 48.5 & 14.0 & 44.5 & 22.5 & 51.0 & 21.0 & 52.5 & 17.0 & 42.5 & 21.0 & 54.0 & 32.0 & 63.0 & 22.50 & 51.60 \\
    \rowcolor{gray!10}
    \textbf{BHA (Ours)} / CSLS-2 & \underline{43.5} & \textbf{80.0} & \textbf{40.0} & \underline{66.0} & \underline{22.0} & \underline{48.5} & \underline{24.0} & \underline{57.0} & \underline{20.5} & \underline{50.5} & \underline{33.0} & \underline{62.0} & \underline{28.0} & \textbf{66.5} & \underline{21.5} & \textbf{57.5} & \underline{27.5} & \underline{64.0} & \textbf{43.5} & \textbf{81.5} & \underline{30.35} & \underline{63.35} \\
    \rowcolor{gray!10}
    \textbf{BHA (Ours)} / CSLS-10 & \textbf{47.5} & \underline{77.5} & \underline{39.5} & \textbf{68.0} & \textbf{23.5} & \textbf{49.5} & \textbf{25.0} & \textbf{58.5} & \textbf{21.5} & \textbf{52.0} & \textbf{34.5} & \textbf{64.0} & \textbf{30.5} & \underline{66.0} & \textbf{22.0} & \underline{56.0} & \textbf{29.0} & \textbf{65.5} & \textbf{43.5} & \underline{77.0} & \textbf{31.65} & \textbf{63.40} \\
    \bottomrule
  \end{tabular}%
  }
\end{table}

\begin{table}[!htbp]
    \caption{\textbf{Bidirectional LOSO retrieval on THINGS-EEG2.} 200-way accuracy (\%); columns 01--10 denote subjects. All rows report BHA.}
    \label{table:retrieval_loso_persubject}
    \centering
    \small
    \setlength{\tabcolsep}{8pt}
    \renewcommand{\arraystretch}{0.9}
    \resizebox{\fitwidth}{!}{%
    \begin{tabular}{@{}ll*{11}{r}@{}}
    \toprule
    \textbf{Metric} & \textbf{Scoring} & 01 & 02 & 03 & 04 & 05 & 06 & 07 & 08 & 09 & 10 & \textbf{Avg} \\
    \midrule
    \multicolumn{13}{@{}l}{\textbf{EEG$\rightarrow$Image}} \\
     & Raw & 35.0 & 29.0 & 12.5 & 21.0 & 14.0 & 22.5 & 21.0 & 17.0 & 21.0 & 32.0 & 22.50 \\
     & CSLS-2 & 43.5 & 40.0 & 22.0 & 24.0 & 20.5 & 33.0 & 28.0 & 21.5 & 27.5 & 43.5 & \underline{30.35} \\
     \multirow{-3}{*}{Top-1} & CSLS-10 & 47.5 & 39.5 & 23.5 & 25.0 & 21.5 & 34.5 & 30.5 & 22.0 & 29.0 & 43.5 & \textbf{31.65} \\
    \addlinespace[0.5pt]
     & Raw & 64.5 & 57.5 & 38.0 & 48.5 & 44.5 & 51.0 & 52.5 & 42.5 & 54.0 & 63.0 & 51.60 \\
     & CSLS-2 & 80.0 & 66.0 & 48.5 & 57.0 & 50.5 & 62.0 & 66.5 & 57.5 & 64.0 & 81.5 & \underline{63.35} \\
     \multirow{-3}{*}{Top-5} & CSLS-10 & 77.5 & 68.0 & 49.5 & 58.5 & 52.0 & 64.0 & 66.0 & 56.0 & 65.5 & 77.0 & \textbf{63.40} \\
    \midrule
    \multicolumn{13}{@{}l}{\textbf{Image$\rightarrow$EEG}} \\
     & Raw & 40.0 & 36.0 & 25.0 & 30.0 & 25.5 & 38.0 & 33.5 & 21.5 & 31.5 & 49.0 & 33.00 \\
     & CSLS-2 & 46.5 & 38.0 & 25.5 & 30.5 & 26.5 & 39.0 & 36.0 & 26.0 & 34.5 & 53.0 & \underline{35.55} \\
     \multirow{-3}{*}{Top-1} & CSLS-10 & 47.0 & 40.0 & 27.5 & 32.5 & 28.0 & 40.5 & 39.0 & 26.0 & 36.5 & 53.5 & \textbf{37.05} \\
    \addlinespace[0.5pt]
     & Raw & 76.5 & 67.0 & 58.5 & 57.0 & 49.5 & 70.5 & 65.5 & 59.0 & 63.5 & 77.0 & 64.40 \\
     & CSLS-2 & 79.5 & 72.0 & 60.0 & 59.0 & 52.0 & 71.5 & 68.0 & 60.0 & 67.5 & 81.0 & \underline{67.05} \\
     \multirow{-3}{*}{Top-5} & CSLS-10 & 80.0 & 71.5 & 62.5 & 60.5 & 55.5 & 73.5 & 69.0 & 59.0 & 69.5 & 81.0 & \textbf{68.20} \\
    \bottomrule
    \end{tabular}%
    }
\end{table}

\subsubsection{Retrieval Performance in THINGS-MEG}
\label{sec:comprehensive_thingsmeg}

\paragraph{Within-subject and LOSO retrieval comparison.}
We evaluate four within-subject and four LOSO MEG models, using the
readouts in Appendix~\ref{sec:retrieval_implementation}. Batch sizes are
256 for pretraining and 1,024 for readout training. MEG pretraining equally
weights paired and same-concept BiCLIP, with a four-example concept-average
loss weighted by $0.25$. LOSO disables subject tokens and uses source-training
normalization statistics without target-subject fitting. CSLS improves
retrieval in both settings, but the substantial transfer gap persists
across modalities (Tables~\ref{table:retrieval_in_subject_meg}
and~\ref{table:retrieval_loso_comparison_meg}). MBP values follow its published subject-wise table ($33.9/60.7$), also used in Table~\ref{table:retrieval_comparison}; the source paper's summary table instead reports $33.7/60.5$~\citep{zheng2026learning}.

\begin{table}[!htbp]
    \caption{\textbf{Within-subject retrieval on THINGS-MEG.} 200-way Top-1 and Top-5 accuracy (\%). Light gray marks BHA rows.}
    \label{table:retrieval_in_subject_meg}
    \centering
    \resizebox{\fitwidth}{!}{%
    \setlength{\tabcolsep}{8pt}
    \begin{tabular}{l *{10}{c}}
    \toprule
    \multirow{2}{*}{\textbf{Method}}
      & \multicolumn{2}{c}{\textbf{Sub1}}
      & \multicolumn{2}{c}{\textbf{Sub2}}
      & \multicolumn{2}{c}{\textbf{Sub3}}
      & \multicolumn{2}{c}{\textbf{Sub4}}
      & \multicolumn{2}{c}{\textbf{Avg}} \\
    \cmidrule(lr){2-3}  \cmidrule(lr){4-5}  \cmidrule(lr){6-7}  \cmidrule(lr){8-9}  \cmidrule(lr){10-11}
    & \footnotesize{Top-1} & \footnotesize{Top-5}
    & \footnotesize{Top-1} & \footnotesize{Top-5}
    & \footnotesize{Top-1} & \footnotesize{Top-5}
    & \footnotesize{Top-1} & \footnotesize{Top-5}
    & \footnotesize{Top-1} & \footnotesize{Top-5} \\
    \midrule
    NICE~\citep{song2024decoding}
      & 9.6 & 27.8 & 18.5 & 47.8 & 14.2 & 41.6 & 9.0 & 26.6 & 12.8 & 36.0 \\
    NICE-SA~\citep{song2024decoding}
      & 9.8 & 27.8 & 18.6 & 46.4 & 10.5 & 38.4 & 11.7 & 27.2 & 12.7 & 35.0 \\
    NICE-GA~\citep{song2024decoding}
      & 8.7 & 30.5 & 21.8 & 56.6 & 16.5 & 49.7 & 10.3 & 32.3 & 14.3 & 42.3 \\
    UBP~\citep{wu2025bridging}
      & 15.0 & 38.0 & 46.0 & 80.5 & 27.3 & 59.0 & 18.5 & 43.5 & 26.7 & 55.2 \\
    MBP~\citep{zheng2026learning}
      & 14.0 & 31.8 & 63.8 & 91.8 & 41.0 & 78.3 & 17.0 & 41.0 & 33.9 & 60.7 \\
    HyFI~\citep{jo2026hyfi}
      & 17.6 & 40.1 & 63.8 & 91.1 & 38.0 & 76.9 & 23.7 & 50.2 & 35.8 & 64.6 \\
    NeuroBridge~\citep{zhang2026neurobridge}
      & 16.5 & 41.6 & 53.7 & 85.3 & 40.4 & 73.2 & 18.1 & 43.1 & 32.2 & 60.8 \\
    LVBP~\citep{liu2026leveraging}
      & 25.4 & -- & 78.1 & -- & 37.3 & -- & 35.3 & -- & 44.0 & 72.0 \\
        DMSA~\citep{du2026deep}
          & 25.5 & 54.5 & 81.9 & 97.4 & 56.0 & 87.5 & 28.6 & 58.3 & 48.0 & 74.4 \\
    \midrule
    \rowcolor{gray!10}
    \textbf{BHA (Ours)} / Raw & 38.0 & 74.5 & 78.0 & \underline{99.5} & 63.0 & 89.0 & 43.5 & 80.0 & 55.63 & 85.75 \\
    \rowcolor{gray!10}
    \textbf{BHA (Ours)} / CSLS-2 & \textbf{53.0} & \textbf{86.0} & \textbf{92.5} & \textbf{100.0} & \textbf{71.5} & \textbf{94.5} & \textbf{55.5} & \textbf{85.0} & \textbf{68.13} & \textbf{91.38} \\
    \rowcolor{gray!10}
    \textbf{BHA (Ours)} / CSLS-10 & \underline{51.0} & \underline{83.5} & \underline{89.0} & \textbf{100.0} & \underline{68.5} & \underline{93.5} & \underline{52.0} & \underline{84.0} & \underline{65.13} & \underline{90.25} \\
    \bottomrule
    \end{tabular}%
    }
\end{table}

\begin{table}[!htb]
    \caption{\textbf{LOSO retrieval on THINGS-MEG.} 200-way Top-1 and Top-5 accuracy (\%). Light gray marks BHA rows.}
    \label{table:retrieval_loso_comparison_meg}
    \centering
    \resizebox{\fitwidth}{!}{%
    \setlength{\tabcolsep}{8pt}
    \begin{tabular}{l *{10}{c}}
    \toprule
    \multirow{2}{*}{\textbf{Method}}
      & \multicolumn{2}{c}{\textbf{Sub1}}
      & \multicolumn{2}{c}{\textbf{Sub2}}
      & \multicolumn{2}{c}{\textbf{Sub3}}
      & \multicolumn{2}{c}{\textbf{Sub4}}
      & \multicolumn{2}{c}{\textbf{Avg}} \\
    \cmidrule(lr){2-3}  \cmidrule(lr){4-5}  \cmidrule(lr){6-7}  \cmidrule(lr){8-9}  \cmidrule(lr){10-11}
    & \footnotesize{Top-1} & \footnotesize{Top-5}
    & \footnotesize{Top-1} & \footnotesize{Top-5}
    & \footnotesize{Top-1} & \footnotesize{Top-5}
    & \footnotesize{Top-1} & \footnotesize{Top-5}
    & \footnotesize{Top-1} & \footnotesize{Top-5} \\
    \midrule
    UBP~\citep{wu2025bridging}
      & 2.0 & 5.7 & 1.5 & 17.2 & 2.7 & 10.5 & \underline{2.5} & 8.0 & 2.2 & 10.4 \\
    HyFI~\citep{jo2026hyfi}
      & 2.7 & 6.0 & 4.3 & 17.0 & 3.5 & 12.5 & 1.0 & 7.6 & $3.2^\dagger$ & $11.5^\dagger$ \\
    NeuroBridge~\citep{zhang2026neurobridge}
      & \textbf{4.3} & \textbf{13.1} & 3.6 & 15.6 & 3.0 & 11.2 & \underline{2.5} & \underline{11.3} & 3.4 & 12.8 \\
        LVBP~\citep{liu2026leveraging}
          & \underline{2.9} & -- & 7.7 & -- & \underline{5.8} & -- & \textbf{4.7} & -- & \underline{5.3} & 15.9 \\
    DMSA~\citep{du2026deep}
      & 1.3 & \underline{6.9} & 6.6 & 18.5 & 5.4 & 18.5 & 1.5 & 7.9 & 3.7 & 13.0 \\

    \midrule
    \rowcolor{gray!10}
    \textbf{BHA (Ours)} / Raw & 0.5 & 5.0 & 7.5 & 18.5 & 5.0 & 19.5 & 1.5 & 8.0 & 3.63 & 12.75 \\
    \rowcolor{gray!10}
    \textbf{BHA (Ours)} / CSLS-2 & 1.0 & 6.5 & \underline{10.0} & \underline{24.5} & \textbf{7.5} & \underline{21.0} & \underline{2.5} & \textbf{14.0} & 5.25 & \underline{16.50} \\
    \rowcolor{gray!10}
    \textbf{BHA (Ours)} / CSLS-10 & 0.5 & 6.5 & \textbf{11.5} & \textbf{26.0} & \textbf{7.5} & \textbf{22.5} & \underline{2.5} & \textbf{14.0} & \textbf{5.50} & \textbf{17.25} \\
    \bottomrule
    \end{tabular}%
    }
\end{table}

\subsubsection{Training-Stage and CSLS Ablations}
\label{sec:retrieval_csls_ablation}
We examine two sources of retrieval gains: learning task-specific readouts
and correcting candidate scores at evaluation. Separate comparisons assess
readout performance, training-data requirements, and dependence on the
CSLS reference cohort.

\paragraph{Two-stage training and retrieval readouts.}
To determine whether readout specialization improves both discrimination
and transfer, we compare Stage-1 BrainViT with the Stage-2 shared and gated
outputs using Top-1 and Top-5 accuracy
(Table~\ref{table:retrieval_stage_ablation}). The shared and gated readouts are evaluated from the same Stage-2 model.
All rows use 63 channels, 80-repeat queries, and 200 candidates; results
are averaged over ten subjects or LOSO folds. Gating improves both metrics
within subjects, whereas pretrained BrainViT generally performs better
under LOSO, which motivates evaluating specialization and cross-subject transfer separately.

\begin{table}[!htbp]
\centering
\caption{\textbf{Two-stage retrieval comparison on THINGS-EEG2.} Mean Top-1 / Top-5 accuracy (\%).}
\label{table:retrieval_stage_ablation}
\footnotesize
\setlength{\tabcolsep}{1pt}
\renewcommand{\arraystretch}{0.95}
\setlength{\aboverulesep}{1.5pt}
\setlength{\belowrulesep}{1.5pt}
\begin{tabular}{@{}lrrrrrr@{}}
\toprule
& \multicolumn{3}{c}{Within-subject} & \multicolumn{3}{c}{LOSO} \\
\cmidrule(lr){2-4}\cmidrule(lr){5-7}
Stage / readout & Raw & CSLS-2 & CSLS-10 & Raw & CSLS-2 & CSLS-10 \\
\midrule
Stage 1 / pretrained & 80.80 / 96.25 & 90.95 / \underline{99.05} & 87.85 / 98.55 & \textbf{22.50} / \textbf{51.60} & \underline{30.35} / \textbf{63.35} & \textbf{31.65} / \textbf{63.40} \\
Stage 2 / shared expert & \underline{82.20} / \underline{96.80} & \underline{91.95} / \underline{99.05} & \underline{88.65} / \underline{98.60} & \underline{22.05} / \underline{50.15} & \textbf{30.50} / \underline{61.15} & \underline{31.55} / \underline{61.40} \\
Stage 2 / gated & \textbf{87.00} / \textbf{98.45} & \textbf{93.90} / \textbf{99.70} & \textbf{91.35} / \textbf{99.30} & 20.20 / 48.35 & 28.15 / 60.25 & 29.90 / 60.50 \\
\bottomrule
\end{tabular}
\end{table}

\paragraph{Training-data scale.}
\label{sec:data_scale_protocol}
We train separate backbones on nested $25\%$, $50\%$, $75\%$, and $100\%$
subsets of the 14,886 EEG training images, keeping validation data and
Stage~2 settings unchanged. Shared and gated scores come from the same
trained readouts; the full-data model supplies the main results.
Gated retrieval outperforms both pretrained BrainViT and its shared
readout at every fraction (Figure~\ref{fig:data_scale_withinsubject}),
showing that the advantage persists with less training data.

\paragraph{\texorpdfstring{\protect\redhl{Teacher diversity.}}{Teacher diversity.}}
\redhl{To separate teacher diversity from the number of branches, we keep four
branches and the parameter count and change one teacher at a time
(Table~\ref{table:teacher_diversity}). The OpenCLIP-H final branch stays unchanged in
every control, because it is the base of the fused Stage~1 target and the
generator's conditioning space. Replacing DFN-B with other vision
foundation models (Qwen3-VL~\citep{bai2025qwen3vl}, SigLIP~\citep{zhai2023sigmoid},
InternViT~\citep{chen2024internvl}, EVA-02~\citep{fang2024eva02},
DINOv3~\citep{simeoni2025dinov3}, and V-JEPA~2~\citep{assran2025vjepa2}), each at the depth reported as best by DMSA~\citep{du2026deep}
or at half depth otherwise, or with intermediate OpenCLIP-H blocks, yields a
composition gain whenever EEG predicts the new branch well, whereas
self-supervised teachers with poorly predicted branches add nothing. Across the 15 replacements in panels (a)--(b), the gain tracks how well EEG
predicts the new branch (Spearman $\rho=0.88$). Extra
branches alone add nothing (panel c): a copy of another teacher in the DINOv2-L
or MAE branch leaves BHA at 85.2--86.6, even a second DFN-B that EEG predicts as
well as the first (73.5 alone), and any copy in place of DFN-B removes the gain
($-0.4$ to $-0.8$). The gain therefore requires a branch that is both distinct
and predictable from EEG. The composition weights follow these changes: two
DFN-B copies share the teacher weight (about 0.66/0.32), and when DFN-B is
replaced by a copy, the gate falls from 0.93 to 0.68--0.81.}

\begin{table}[!htbp]\redhlon
    \centering
    \caption{\textbf{Teacher diversity controls.} THINGS-EEG2 within-subject Raw
    Top-1 (\%). BHA uses OpenCLIP-H (final), DFN-B (block 6), MAE-L (block 18)
    and DINOv2-L (block 12); every row keeps four branches, changes one, and
    retrains all stages. (a) The DFN-B branch is replaced by another model.
    (b) The DFN-B branch is replaced by an intermediate OpenCLIP-H block (CLS or
    mean patch token); the OpenCLIP-H final branch stays. (c) One branch is
    replaced by a copy of a teacher already in the model.
    \emph{Branch}: the changed branch alone. \emph{Stage~1}: fused space.
    \emph{Shared}: shared space only. \emph{Gain}: BHA minus Shared.
    Means $\pm$ SD over three seeds, ten subjects each.}
    \label{table:teacher_diversity}
    \footnotesize
    \providecommand{\tdsd}[1]{{\scriptsize$\pm$#1}}
    \setlength{\tabcolsep}{4pt}
    \begin{tabular}{@{}llccccc@{}}
    \toprule
    Changed branch & Replaced by & Branch & Stage~1 & Shared & BHA & Gain \\
    \midrule
    \multicolumn{7}{@{}l}{\textit{(a) DFN-B replaced by another model}} \\
    \multicolumn{2}{@{}l}{Default (DFN-B)} & 76.5\tdsd{0.9} & 81.1\tdsd{0.9} & 83.3\tdsd{0.7} & 86.5\tdsd{0.4} & $+3.2$\tdsd{1.0} \\
    DFN-B & Qwen3-VL vision tower & 70.0\tdsd{1.0} & 78.5\tdsd{0.3} & 80.5\tdsd{0.5} & 83.3\tdsd{0.5} & $+2.8$\tdsd{0.4} \\
    DFN-B & SigLIP-so400m & 67.4\tdsd{0.4} & 78.8\tdsd{1.0} & 81.2\tdsd{1.3} & 83.2\tdsd{1.1} & $+2.0$\tdsd{0.2} \\
    DFN-B & InternViT-6B & 57.8\tdsd{1.6} & 81.8\tdsd{0.3} & 84.1\tdsd{0.6} & 86.4\tdsd{0.1} & $+2.3$\tdsd{0.6} \\
    DFN-B & EVA-02-L & 56.4\tdsd{1.6} & 74.1\tdsd{0.8} & 77.1\tdsd{0.3} & 78.1\tdsd{0.6} & $+1.0$\tdsd{0.7} \\
    DFN-B & DINOv3-L, block 9 & 47.7\tdsd{3.3} & 76.1\tdsd{0.6} & 79.6\tdsd{0.7} & 79.4\tdsd{0.4} & $-0.2$\tdsd{0.5} \\
    DFN-B & V-JEPA 2-L & 10.8\tdsd{1.3} & 77.3\tdsd{0.6} & 79.6\tdsd{0.4} & 79.4\tdsd{0.2} & $-0.2$\tdsd{0.5} \\
    DFN-B & DINOv2-L, block 6 & 9.4\tdsd{5.1} & 76.4\tdsd{0.7} & 79.0\tdsd{0.5} & 78.0\tdsd{0.6} & $-1.0$\tdsd{0.1} \\
    \midrule
    \multicolumn{7}{@{}l}{\textit{(b) DFN-B replaced by an intermediate OpenCLIP-H block}} \\
    DFN-B & block 8, CLS & 58.3\tdsd{1.2} & 78.3\tdsd{0.4} & 80.6\tdsd{0.2} & 83.4\tdsd{0.3} & $+2.8$\tdsd{0.1} \\
    DFN-B & block 9, CLS & 55.7\tdsd{2.3} & 77.7\tdsd{0.6} & 80.9\tdsd{0.7} & 82.5\tdsd{0.1} & $+1.6$\tdsd{0.7} \\
    DFN-B & block 10, CLS & 57.9\tdsd{1.1} & 77.7\tdsd{0.4} & 80.0\tdsd{0.4} & 82.2\tdsd{0.7} & $+2.2$\tdsd{1.1} \\
    DFN-B & block 11, CLS & 58.9\tdsd{0.4} & 77.3\tdsd{0.7} & 79.8\tdsd{0.7} & 81.8\tdsd{0.5} & $+2.0$\tdsd{0.3} \\
    DFN-B & block 8, mean & 64.6\tdsd{0.7} & 78.3\tdsd{0.8} & 80.9\tdsd{0.4} & 83.8\tdsd{0.3} & $+2.9$\tdsd{0.5} \\
    DFN-B & block 9, mean & 67.3\tdsd{1.2} & 78.9\tdsd{0.6} & 81.6\tdsd{0.5} & 84.7\tdsd{0.1} & $+3.1$\tdsd{0.4} \\
    DFN-B & block 10, mean & 71.2\tdsd{1.6} & 79.8\tdsd{0.4} & 82.0\tdsd{0.2} & 85.6\tdsd{0.5} & $+3.6$\tdsd{0.6} \\
    DFN-B & block 11, mean & 73.8\tdsd{0.7} & 80.1\tdsd{0.7} & 82.4\tdsd{0.3} & 86.2\tdsd{0.2} & $+3.8$\tdsd{0.4} \\
    \midrule
    \multicolumn{7}{@{}l}{\textit{(c) One branch replaced by a copy of another teacher}} \\
    DINOv2-L & MAE-L & 30.7\tdsd{5.3} & 80.4\tdsd{0.8} & 82.9\tdsd{0.5} & 86.6\tdsd{0.4} & $+3.7$\tdsd{0.6} \\
    DINOv2-L & DFN-B & 73.5\tdsd{1.4} & 80.8\tdsd{0.4} & 82.9\tdsd{0.2} & 86.3\tdsd{0.6} & $+3.4$\tdsd{0.8} \\
    DINOv2-L & OpenCLIP-H final & 37.2\tdsd{1.2} & 79.6\tdsd{1.0} & 82.3\tdsd{0.4} & 86.5\tdsd{0.2} & $+4.2$\tdsd{0.2} \\
    MAE-L & DINOv2-L & 22.9\tdsd{3.6} & 79.9\tdsd{0.6} & 82.0\tdsd{0.1} & 85.2\tdsd{0.5} & $+3.2$\tdsd{0.6} \\
    MAE-L & DFN-B & 74.1\tdsd{1.5} & 80.2\tdsd{0.5} & 82.3\tdsd{0.5} & 85.8\tdsd{0.5} & $+3.5$\tdsd{0.7} \\
    MAE-L & OpenCLIP-H final & 37.4\tdsd{0.4} & 79.6\tdsd{0.7} & 82.1\tdsd{0.3} & 86.0\tdsd{0.2} & $+3.9$\tdsd{0.2} \\
    DFN-B & DINOv2-L & 27.0\tdsd{7.4} & 75.7\tdsd{0.3} & 77.8\tdsd{0.3} & 77.4\tdsd{0.5} & $-0.4$\tdsd{0.4} \\
    DFN-B & MAE-L & 25.7\tdsd{9.3} & 76.0\tdsd{0.8} & 78.5\tdsd{0.4} & 77.7\tdsd{0.6} & $-0.8$\tdsd{0.3} \\
    DFN-B & OpenCLIP-H final & 24.7\tdsd{3.0} & 75.0\tdsd{0.6} & 77.3\tdsd{0.4} & 76.7\tdsd{0.5} & $-0.6$\tdsd{0.5} \\
    \bottomrule
    \end{tabular}
\end{table}

\paragraph{CSLS neighborhood size.}
We test how retrieval depends on the neighborhood size used to correct
candidate scores. Each $k$ uses the same raw score matrix, readouts,
trial-averaged queries, and 200-image gallery; all 200 queries supply the
reference statistics. Results average ten EEG or four MEG subjects.
Smaller neighborhoods favor within-subject retrieval, whereas LOSO
benefits from larger neighborhoods in several metrics
(Table~\ref{table:retrieval_csls_ablation}). The preferred correction thus
depends on the evaluation setting. \redhl{The within-subject $k$ is selected on validation data, as described next.}

\begin{table}[!htbp]
    \centering
    \caption{\textbf{CSLS neighborhood-size ablation.} Mean 200-way Top-1 / Top-5 accuracy (\%).}
    \label{table:retrieval_csls_ablation}
    \footnotesize
    \setlength{\tabcolsep}{2pt}
\renewcommand{\arraystretch}{0.95}
\setlength{\aboverulesep}{1.5pt}
\setlength{\belowrulesep}{1.5pt}
    \begin{tabular}{@{}llccccc@{}}
    \toprule
    \multirow{2}{*}{Method} & \multirow{2}{*}{Setting}
      & \multirow{2}{*}{Raw} & \multicolumn{4}{c}{CSLS} \\
    \cmidrule(lr){4-7}
      & & & $k=2$ & $k=4$ & $k=10$ & $k=20$ \\
    \midrule
    \multicolumn{7}{@{}l}{\textit{THINGS-EEG2}} \\
    \textbf{BHA (Ours)} & Within-subject & 87.00 / 98.45 & \textbf{93.90} / \textbf{99.70} & \underline{92.95} / \underline{99.55} & 91.35 / 99.30 & 90.45 / 99.15 \\
    \textbf{BHA (Ours)} & LOSO & 22.50 / 51.60 & 30.35 / 63.35 & 31.40 / \underline{63.45} & \textbf{31.65} / 63.40 & \underline{31.50} / \textbf{63.55} \\
    \midrule
    \multicolumn{7}{@{}l}{\textit{THINGS-MEG}} \\
    \textbf{BHA (Ours)} & Within-subject & 55.63 / 85.75 & \textbf{68.13} / \textbf{91.38} & \underline{66.63} / \underline{90.75} & 65.13 / 90.25 & 63.25 / 89.25 \\
    \textbf{BHA (Ours)} & LOSO & 3.63 / 12.75 & 5.25 / 16.50 & \underline{5.63} / 16.63 & 5.50 / \underline{17.25} & \textbf{5.88} / \textbf{17.38} \\
    \bottomrule
    \end{tabular}
\end{table}

\paragraph{\texorpdfstring{\protect\redhl{Selecting $k$ on validation data.}}{Selecting k on validation data.}}
\label{sec:csls_validation_k}
\redhl{To choose $k$ without test data, we draw repeated 200-way retrieval problems from each subject's 1,654 held-out validation images, mirroring the test protocol: each problem's 200 brain queries supply the CSLS statistics. We score the Stage-2 gated readout for $k\in\{1,2,4,10,20\}$ and select the $k$ with the highest mean validation Top-1 (Table~\ref{table:csls_validation_k}). Averaged over ten subjects and three training seeds, $k=2$ performs best. It is selected for eight of ten subjects, and for the other two, $k=1$ is ahead by at most $0.04$ points. All CSLS results in the paper therefore use $k=2$.}

\begin{table}[!htbp]
\centering
\caption[CSLS neighborhood size selected on validation data.]{\redhl{\textbf{CSLS neighborhood size selected on validation data.} THINGS-EEG2 Stage-2 gated readout, mean 200-way validation Top-1 (\%) over ten subjects. Validation queries average four repetitions, so accuracy is lower than on the test set.}}
\label{table:csls_validation_k}
\footnotesize
\begin{tabular}{@{}lcccccc@{}}
\toprule
& \redhl{Raw} & \redhl{$k=1$} & \redhl{$k=2$} & \redhl{$k=4$} & \redhl{$k=10$} & \redhl{$k=20$} \\
\midrule
\redhl{Validation Top-1} & \redhl{60.76} & \redhl{66.29} & \redhl{\textbf{66.51}} & \redhl{65.54} & \redhl{63.99} & \redhl{62.96} \\
\bottomrule
\end{tabular}
\end{table}

\paragraph{\texorpdfstring{\protect\redhl{CSLS for baselines.}}{CSLS for baselines.}}
\redhl{We apply the same procedure to open-source baselines from
Table~\ref{table:retrieval_comparison}. Each is run with its official code on
our split, with checkpoints selected on validation, and scored with the
same CSLS code and the same 200 reference queries as BHA
(Table~\ref{table:csls_baselines}); the inductive variant draws its references
from validation images only. CSLS lifts the baselines more than BHA, narrowing
BHA's lead rather than creating it, and the ranking holds under every protocol.
DMSA's reported $82.6\%$ follows its released code's test-loss epoch selection.}

\begin{table}[!htbp]\redhlon
    \centering
    \caption{\textbf{CSLS applied to baselines.} THINGS-EEG2 within-subject
    Top-1 (\%). For every method, $k$ is selected on each subject's validation
    images ($k=2$ for BHA). Means over ten subjects and three training seeds.}
    \label{table:csls_baselines}
    \footnotesize
    \begin{tabular}{@{}lccc@{}}
    \toprule
    Method & Raw & CSLS & Inductive CSLS \\
    \midrule
    NICE~\citep{song2024decoding} & 27.5 & 38.6 & 32.9 \\
    ATM-S~\citep{li2024visual} & 28.7 & 40.7 & 36.1 \\
    UBP~\citep{wu2025bridging} & 46.4 & 57.1 & 49.8 \\
    MBP~\citep{zheng2026learning} & 75.7 & 86.3 & 78.5 \\
    DMSA~\citep{du2026deep} & 78.9 & 91.2 & 82.0 \\
    \rowcolor{gray!10}
    BHA & \textbf{86.5} & \textbf{93.9} & \textbf{88.6} \\
    \bottomrule
    \end{tabular}
\end{table}

\begin{figure}[!t]
    \centering
    \includegraphics[width=0.9\textwidth,pagebox=cropbox]{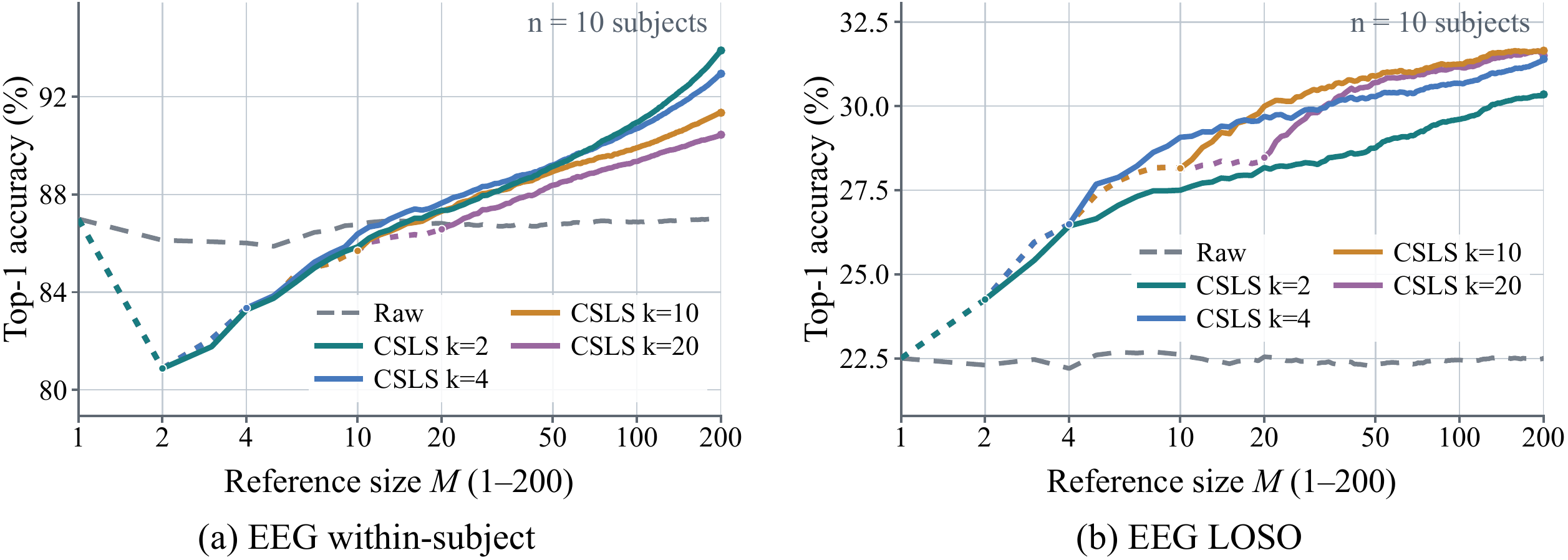}
    \caption[CSLS reference-size ablation: THINGS-EEG2.]{\textbf{CSLS reference-size ablation: THINGS-EEG2.}
(a) Within-subject Stage-2 gated and (b) LOSO Stage-1 pretrained Top-1.
\redhl{Lines average 200 subset repetitions per subject, then ten subjects.} The gallery has 200 images; dotted
segments use $k_{\mathrm{eff}}=M<k$.}
    \label{fig:eeg_csls}

    \vspace{12pt}
    \includegraphics[width=0.9\textwidth,pagebox=cropbox]{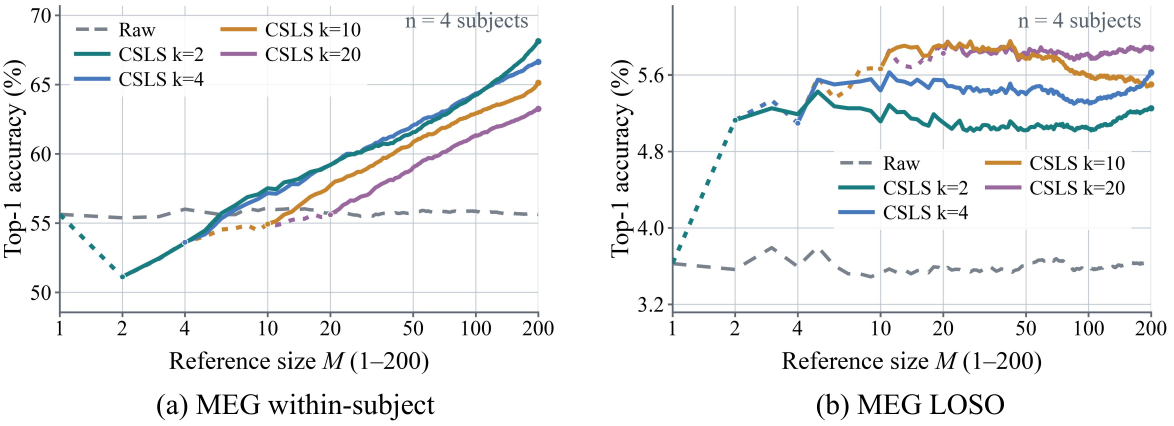}
    \caption[CSLS reference-size ablation: THINGS-MEG.]{\textbf{CSLS reference-size ablation: THINGS-MEG.}
(a) Within-subject Stage-2 gated and (b) LOSO Stage-1 pretrained Top-1.
\redhl{Sampling follows \mbox{Figure~\ref{fig:eeg_csls}}, averaged over four subjects.}
Raw and CSLS use the same query subsets and 200-image gallery.}
    \label{fig:meg_csls}
\end{figure}

\paragraph{CSLS reference-set size.}
We vary the available query cohort from $M=1$ to $200$ while retaining
the 200-image gallery (Figures~\ref{fig:eeg_csls} and~\ref{fig:meg_csls}).
Each subject uses nested subsets from 200 random orderings; the same
queries supply CSLS statistics and evaluation, and Raw uses identical
subsets. Trial averages and readouts are unchanged, with
$k_{\mathrm{eff}}=\min(k,M)$. CSLS preserves Raw rankings at $M=1$ and
recovers the main results at $M=200$, quantifying its dependence on other
test queries.

\subsubsection{Temporal Window Ablation}
\label{sec:temporal_window_ablation}

\paragraph{Evaluation details.}
We measure temporal dependence by deleting or retaining short windows in
THINGS-EEG2, THINGS-EEG1, THINGS-MEG, and Alljoined-1.6M
(Figure~\ref{fig:eeg_meg_time}; datasets in Appendix~\ref{sec:dataset_description}).
Each subject's frozen model evaluates its teacher
branches, shared expert, and composed output with Raw 200-way retrieval.
Trial averages follow the dataset settings: 80 for THINGS-EEG2,
12 for THINGS-EEG1 and MEG, and approximately 80 for Alljoined.

\paragraph{Delete and Keep masks.}
For $\mathbf{X}\in\mathbb{R}^{C\times T}$, let $\tau_n$ be the time of
sample $n$ and define $m_w(\tau_n)=\mathbf{1}[a_w\leq\tau_n<a_w+50\,\mathrm{ms}]$.
The mask is shared across channels. We compare
\begin{equation}
    \mathbf{X}^{\mathrm{Delete},w}_{c,n}=\mathbf{X}_{c,n}\bigl(1-m_w(\tau_n)\bigr),
    \qquad
    \mathbf{X}^{\mathrm{Keep},w}_{c,n}=\mathbf{X}_{c,n}m_w(\tau_n).
\end{equation}
Delete zeros only the selected window; Keep zeros everything outside it.
We evaluate each masked input with the frozen model, preserving sequence
length and sample times without rescaling or renormalization. We scan 96
windows with starts $a_w=0,10,\ldots,950$ ms; the figure shows window centers
up to 795 ms on a 0--800 ms axis.

For subject $s$, let $\operatorname{Acc}_s$ denote Top-1 accuracy as a fraction. The Delete curves in Figure~\ref{fig:eeg_meg_time} report the paired decrease
$100[\operatorname{Acc}_s(\mathbf{X})-\operatorname{Acc}_s(\mathbf{X}^{\mathrm{Delete},w})]$ in percentage points; the Keep curves in Figure~\ref{fig:eeg_meg_time} report $100\operatorname{Acc}_s(\mathbf{X}^{\mathrm{Keep},w})$ in percent, with chance at
$0.5\%$. We compute these quantities separately for each subject before
reporting their mean and sample standard deviation.

\paragraph{Temporal sensitivity.}
THINGS-EEG1, THINGS-EEG2, and THINGS-MEG show early sensitivity around
100--125 ms and additional sensitivity later in the response. Later peaks
occur near image offset or subsequent stimulus onset, so the masking
experiment cannot distinguish these overlapping contributions. Keep and
Delete curves capture complementary aspects of model dependence on time;
their peaks need not coincide.

\paragraph{Temporal interpretation.}
Alljoined timestamps contain sample-clock drift and gaps, whereas the
preprocessing maps events onto a nominal uniform sample axis. Its plotted
latencies are therefore approximate and should not be used for precise
cross-dataset timing comparisons. Across datasets, these curves measure
the sensitivity of frozen models to temporal masking; overlapping responses
and acquisition differences preclude identifying ERP components or causal
effects of stimulus duration.

\begin{figure}[!t]
    \centering
    \includegraphics[width=\textwidth,height=0.72\textheight,keepaspectratio,pagebox=cropbox]{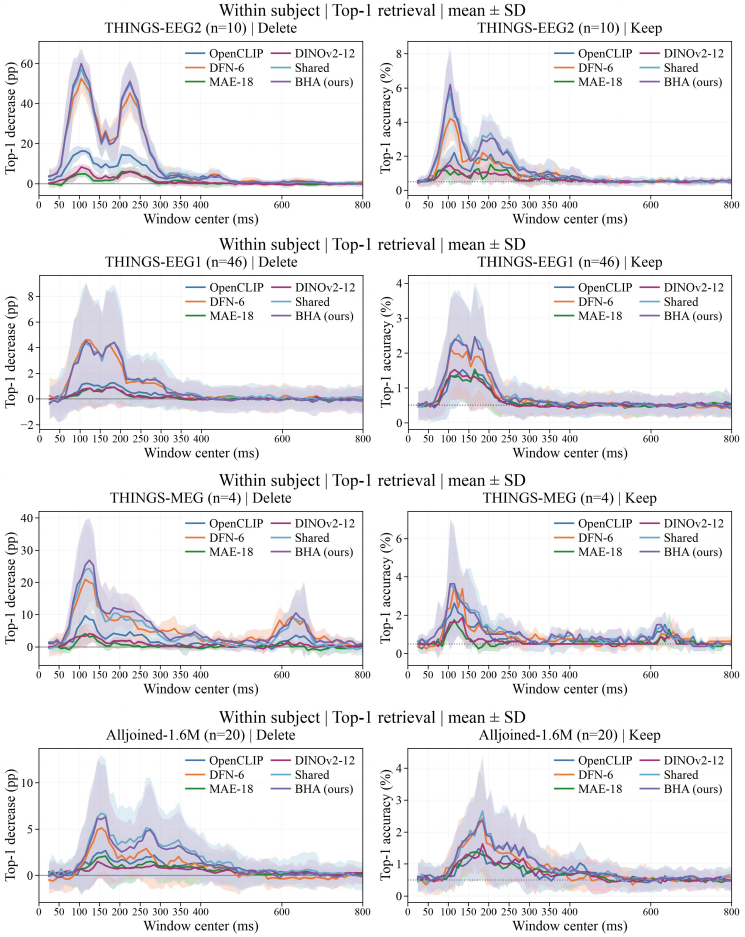}
    \caption{\textbf{Temporal masking of frozen models.}
Rows: THINGS-EEG2 ($n=10$), THINGS-EEG1 ($n=46$), THINGS-MEG ($n=4$),
and Alljoined ($n=20$). Left: Full minus Delete Top-1 (percentage points).
Right: Keep Top-1 (\%). Windows span 50 ms with 10 ms stride;
lines/bands show subject mean/$\pm1$ SD. Reference lines mark zero
decrease and $0.5\%$ chance. Alljoined timing is provisional
(Section~\ref{sec:temporal_window_ablation}).}
    \label{fig:eeg_meg_time}
\end{figure}

\subsection{Extended Representation Analysis}
\label{sec:extended_representation_analysis}
We examine shared geometry, concept associations, and recoverable image
content to characterize what the learned embeddings retain beyond
retrieval accuracy. The shared expert denotes the outer expert $E_0$ defined in
Section~\ref{sec:multi_teacher_moe}.

\subsubsection{Representational Alignment and Matching}
We examine spatial correspondence between brain predictions and image
features by comparing OpenCLIP patch-similarity maps for visual and brain
queries, alongside a random-query reference
(Figure~\ref{fig:openclip_query_map}).

\begingroup
\setlength{\intextsep}{8pt}
\begin{figure}[!t]
    \centering
    \includegraphics[width=0.9\textwidth,pagebox=cropbox]{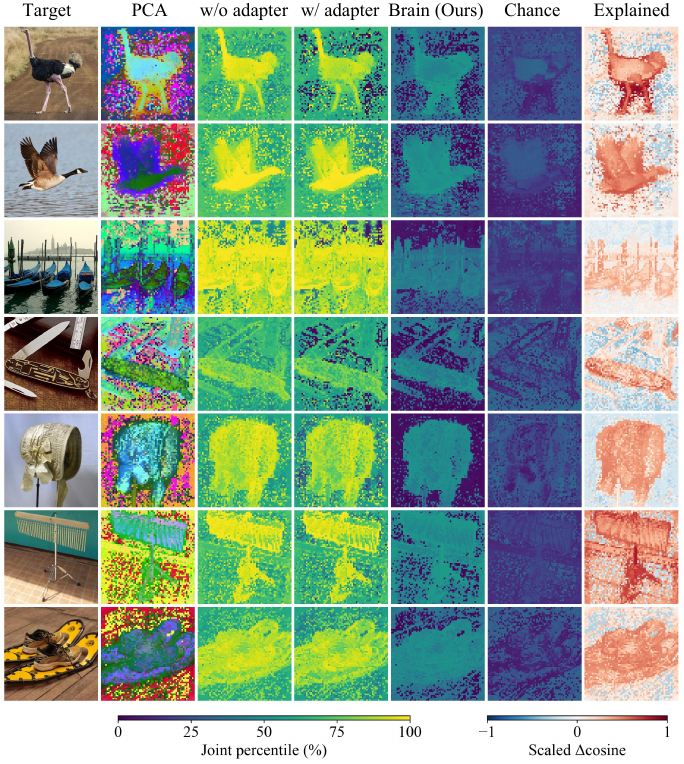}
    \caption{\textbf{Additional OpenCLIP dense-feature query maps.}
    Supplementing Figure~\ref{fig:preview_imgs}, columns show the target image,
    PCA projections of visual features, global-to-patch cosine similarity
    without and with the adapter, patch similarity to the matched brain
    features, a chance-query baseline, and the residual (brain query minus
    chance). Similarity maps use percentile coloring; the residual preserves
    the sign of the cosine difference.}
    \label{fig:openclip_query_map}
\end{figure}
\endgroup

The shared expert combines strong paired alignment with close agreement
in inter-image geometry. Student outputs also agree more strongly across
experts than their visual targets do, consistent with their shared inputs,
backbone, and supervision. This comparison does not isolate the contribution
of neural information to that agreement. Spatial analyses show a further
distinction: experts can respond to similar image regions while assigning
opposite signed contributions to a brain--image match.

\begin{figure}[!t]
    \centering
    \includegraphics[width=0.9\textwidth,pagebox=cropbox]{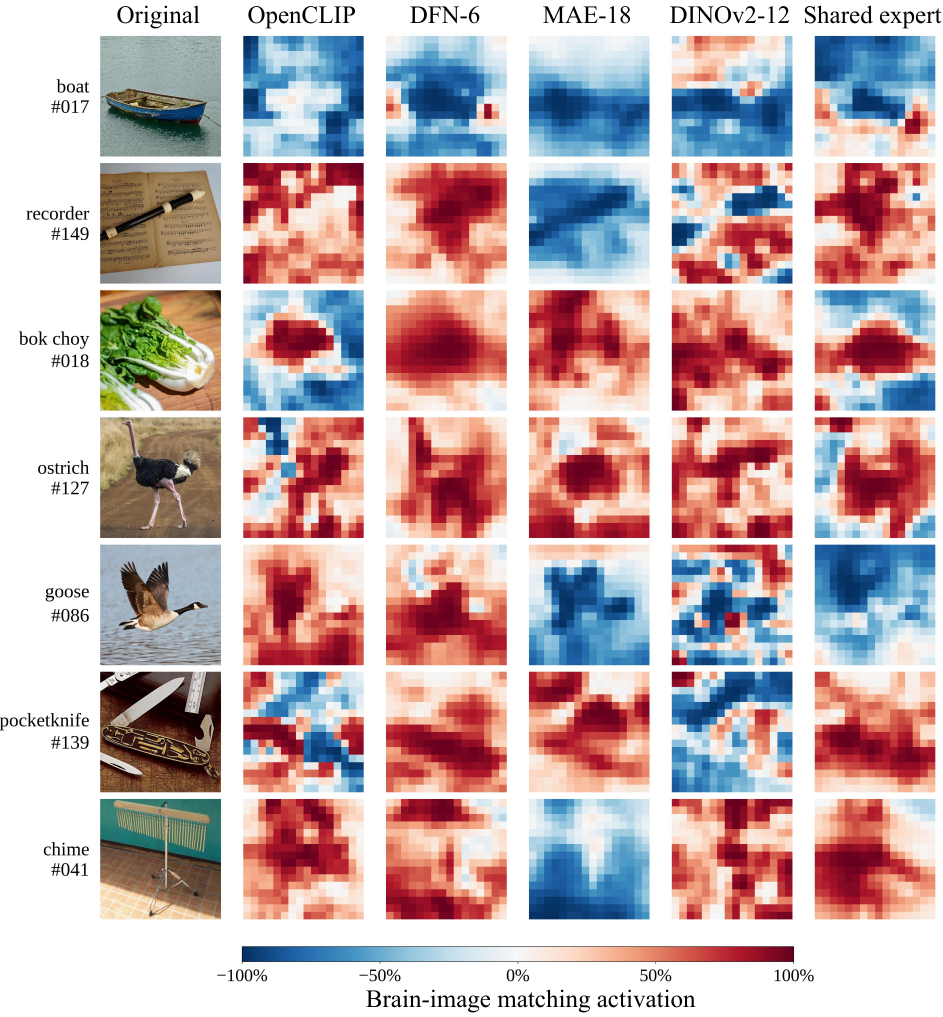}
    \caption{\textbf{Additional brain--image matching attributions.}
Each row shows one test image and ten-subject mean maps for four teacher
branches and the shared expert. Red/blue indicate positive/negative matching
contributions; intensity is a within-map percentile.
Masking and averaging follow Section~\ref{sec:spatial_attribution_protocol}.}
    \label{fig:match_attribution}
\end{figure}

\paragraph{Spatial attribution.}
\label{sec:spatial_attribution_protocol}
We test whether overlapping visual sensitivity yields similar matching
contributions by masking $2\times2$ regions on a $16\times16$ patch grid,
with stride one patch. Images are resized to 224, 512, 256, and 224 pixels
for OpenCLIP, DFN, MAE, and DINOv2, respectively; masks replace pixels with
the image-wide mean color. With the brain query unchanged, attribution is
the original minus masked-image cosine similarity. Overlapping scores
are averaged spatially, then across ten subjects. Colors show signed
within-map percentiles, so intensities are not comparable absolute effect
sizes. These full-image attributions use different preprocessing from the
training crops. Figure~\ref{fig:match_attribution} extends the three examples
in Figure~\ref{fig:rsa}(e) with seven additional images.

\subsubsection{Concept Associations in Model Outputs}

Strong feature alignment need not imply the same concept organization.
We therefore compare concept associations within each subject and inspect
representative neurons in pooled activations. The concept galleries are
collected in Appendix~\ref{sec:concept_examples}; the selection procedure
and quantitative comparisons are given below. Visual adaptation improves
paired alignment, especially for MAE and DINOv2, and the shared expert has
the strongest paired alignment (Figure~\ref{fig:concept_analysis}(a));
the analyses below ask how this ordering relates to semantic correspondence.

\begingroup
\setlength{\intextsep}{10pt}
\begin{figure}[!t]
    \centering
    \includegraphics[width=0.8\textwidth,pagebox=cropbox]{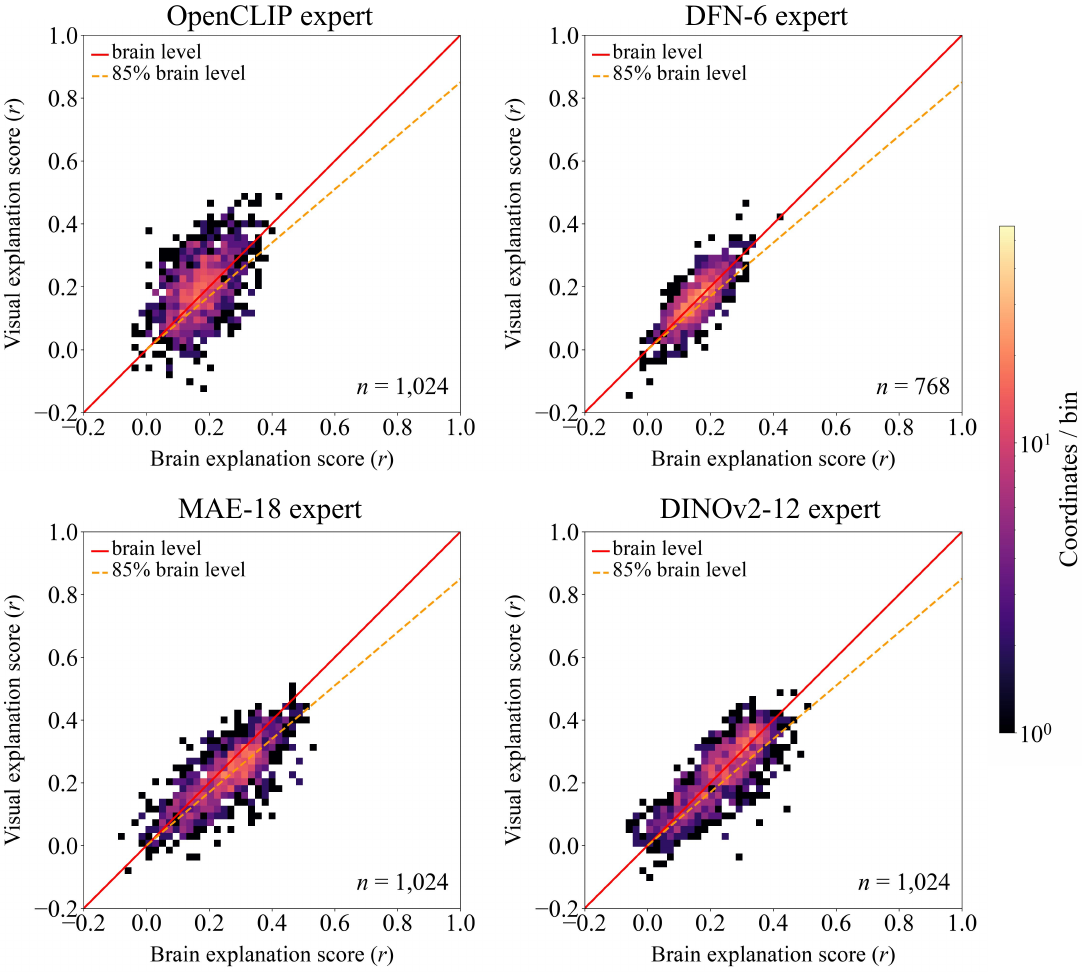}
    \caption{\textbf{Held-out concept explanation scores.}
Student-output (horizontal) versus visual-feature (vertical) Pearson scores,
averaged over ten subjects after separate training-only concept selection
(Section~\ref{sec:concept_score_protocol}). Each branch has 1,024 neurons
except DFN-6 (768). Colors show log bin counts; red: $y=x$;
orange dashed: $y=0.85x$.}
    \label{fig:concept_scatter}
\end{figure}
\endgroup

\paragraph{Subject-wise concept scores.}
\label{sec:concept_score_protocol}
For each subject and shared-expert neuron, we select the category with
the highest signed Pearson correlation on 16,540 training images and
evaluate it on 200 test images. Selection is independent for student and
visual features, with test scores averaged across subjects
(Figure~\ref{fig:concept_analysis}(b)); Figure~\ref{fig:concept_scatter}
extends this comparison to all branches.
We also compare training-image SoftWPMI concept profiles, centering across
neurons within each subject and L2-normalizing before computing paired
brain--visual cosine similarity. The shared expert and DINOv2 show the
closest correspondence, and all branches exceed the neuron-shuffle
reference (Figure~\ref{fig:concept_analysis}(c)). This ordering differs
from paired feature alignment, distinguishing concept structure from
feature similarity.

\paragraph{Concept-gallery construction.}
\label{sec:concept_gallery_protocol}
Figures~\ref{fig:brain_cosine_openclip}--\ref{fig:brain_cosine_bha_gated}
show six neurons per output, selected by mean activation over their
100 strongest training responses after averaging across ten subjects.
For gated outputs, each subject's normalization and gating precede pooling.
Using the Cos variant of CLIP-Dissect~\citep{oikarinen2023clip}, we match
neuron activation vectors $\mathbf{q}_j^{\mathrm{tr}}$ to CLIP ViT-L/14
image--text response vectors $\mathbf{p}_c^{\mathrm{tr}}$ over the same
16,540 training images:
\begin{equation}
 c_j^\star\in\operatorname*{arg\,max}_{c\in\mathcal{C}}
 \operatorname{sim}(\mathbf{q}_j^{\mathrm{tr}},\mathbf{p}_c^{\mathrm{tr}}),
 \qquad |\mathcal{C}|=27.
 \label{eq:concept_cosine_selection}
\end{equation}
Responses remain signed and uncentered. The probing set includes the
internal validation split; the 200 test images have disjoint object
concepts. Each gallery shows the five highest-scoring categories and
five most activating test images. Test images do not select neurons or
categories, and word size represents relative association strength.

The 27 labels are \textit{animal}, \textit{bird}, \textit{body part}, \textit{clothing}, \textit{clothing accessory},
\textit{container}, \textit{dessert}, \textit{drink}, \textit{electronic device}, \textit{food}, \textit{fruit}, \textit{furniture},
\textit{home decor}, \textit{insect}, \textit{kitchen appliance}, \textit{kitchen tool}, \textit{medical equipment},
\textit{musical instrument}, \textit{office supply}, \textit{part of car}, \textit{plant}, \textit{sports equipment},
\textit{tool}, \textit{toy}, \textit{vegetable}, \textit{vehicle}, and \textit{weapon}.
Pooled galleries do not imply that corresponding neurons share selectivity
across subjects. Quantitative scores and category proportions are computed
within subject before averaging (Section~\ref{sec:concept_score_protocol}).

\subsubsection{Feature Inversion under Increasing Image Priors}
\label{sec:feature_inversion}
We examine which visual properties can be recovered from the retrieval
representations using direct pixel optimization, a deep image prior (DIP),
and guided diffusion, following DreamSim~\citep{fu2023dreamsim}.
Inversion seeks an image whose features match a EEG prediction or
image-derived representation.
Appendix~\ref{sec:feature_inversion_examples}
(Figures~\ref{fig:feature_inversion_optimization}--\ref{fig:feature_inversion_diffusion})
compares subject~08 predictions, obtained by averaging 80 EEG trials before
encoding, with image-derived features for the same four stimuli.
The figures cover the four teacher spaces, shared fusion, and gated BHA,
with image-derived DreamSim features as an independent reference.
Reconstruction accuracy is evaluated separately in
Section~\ref{sec:comprehensive_reconstruction_performance}.

\paragraph{Feature matching in the retrieval spaces.}
\redhl{Each inversion maintains a target $\mathbf{z}$ and a frozen feature map $f$, and
searches for an image $\mathbf{I}$ that minimizes
$\mathcal{D}_{i,d}(\mathbf{I})=d\bigl(\mathbf{z},f(\mathbf{I})\bigr)$ for a feature distance $d$:}
\begin{itemize}
\item \emph{EEG branch $k$}: $\mathbf{z}=\hat{\mathbf{e}}_{ik}$ and
$f(\mathbf{I})=\tilde{\mathbf{e}}_k(\mathbf{I})=V_k(\phi_k(\mathbf{I}))$, the adapted space used for
retrieval; for the shared space, $\tilde{\mathbf{e}}_0(\mathbf{I})$ is the fused feature of
Eq.~\eqref{eq:fused_visual_anchor}.
\item \emph{Image branch $k$}: $\mathbf{z}=\mathbf{e}_{ik}=\phi_k(\mathbf{I}_i)$ and
$f=\phi_k$, the original teacher space.
\item \emph{Gated BHA}: $\mathbf{z}$ and $f(\mathbf{I})$ are the weighted concatenations
in Eq.~\eqref{eq:feature_inversion_composition}.
\item \emph{DreamSim reference}: $\mathbf{z}=\phi_{\mathrm{DS}}(\mathbf{I}_i)$ and
$f=\phi_{\mathrm{DS}}$, an independent image-derived reference.
\end{itemize}
For gated BHA, we preserve the score composition in
Eq.~\eqref{eq:gated_retrieval_score} by concatenating weighted feature blocks:
\begin{equation}
\begin{aligned}
\hat{\mathbf{e}}_i^{\mathrm{BHA}}
&=\operatorname{norm}(\hat{\mathbf{e}}_{i0})
 \oplus\bigoplus_{k=1}^{K}\sqrt{g_i\pi_{ik}}\,
 \operatorname{norm}(\hat{\mathbf{e}}_{ik}),\\
\tilde{\mathbf{e}}_i^{\mathrm{BHA}}(\mathbf{I})
&=\operatorname{norm}(\tilde{\mathbf{e}}_0(\mathbf{I}))
 \oplus\bigoplus_{k=1}^{K}\sqrt{g_i\pi_{ik}}\,
 \operatorname{norm}(\tilde{\mathbf{e}}_k(\mathbf{I})).
\end{aligned}
\label{eq:feature_inversion_composition}
\end{equation}
Here $\oplus$ denotes concatenation, and the EEG-dependent $g_i$ and
$\pi_{ik}$ remain changeless. Both vectors have squared norm $1+g_i$;
hence their cosine similarity at $\mathbf{I}=\mathbf{I}_j$ is
$S_{ij}/(1+g_i)$, preserving the retrieval ranking for query $i$.

The first argument of each distance is a constant; only the candidate image
changes during inversion. Brain encoders, retrieval readouts, and visual mappings remain
frozen; gradients through the visual mappings are retained to optimize the image.
All methods produce
$256\times256$ images.

\redhl{Table~\ref{table:feature_inversion_settings} summarizes the three image priors.}

\begin{table}[!htbp]
\centering
\caption{\redhl{\textbf{Feature inversion settings.}}}
\label{table:feature_inversion_settings}
\footnotesize
\setlength{\tabcolsep}{4pt}
\begin{tabular}{@{}llll@{}}
\toprule
Prior & Optimized variable & Matching loss & Schedule \\
\midrule
None & RGB pixels & $d_{\cos}$ on random crops & 2,500 Adam steps, lr $0.01$ \\
Deep image prior & generator weights $\omega$ & $d_{\cos}$ on the full image & 3,000 Adam steps, lr $0.01$ \\
Diffusion & reverse-process mean & $d_{\mathrm{sph}}$ on 64 crops, + TV and range & 1,000 ancestral steps \\
\bottomrule
\end{tabular}
\end{table}

\paragraph{Direct pixel optimization.}
Starting from random RGB pixels, we minimize cosine distance using
$\operatorname{sim}$ from Section~\ref{sec:preliminaries}:
\begin{equation}
\min_{\mathbf{I}}\;
\mathbb{E}_{\mathcal{T}}\!\left[\mathcal{D}_{i,d_{\cos}}(\mathcal{T}\mathbf{I})\right],
\qquad d_{\cos}(\mathbf{a},\mathbf{b})=1-\operatorname{sim}(\mathbf{a},\mathbf{b}).
\label{eq:feature_inversion_pixels}
\end{equation}
Here $\mathcal{T}$ samples a $224\times224$ candidate crop; there is no explicit image regularizer.

\paragraph{Deep image prior.}
We instead parameterize the candidate as $\mathbf{I}=G_{\omega}(\boldsymbol{\xi})$,
where $G_{\omega}$ is a randomly initialized convolutional encoder--decoder
with skip connections and $\boldsymbol{\xi}$ is certain random
noise~\citep{Ulyanov_2018_CVPR}. Its architecture provides the image prior;
only the generator parameters are optimized:
\begin{equation}
\min_{\omega}\;\mathcal{D}_{i,d_{\cos}}(G_{\omega}(\boldsymbol{\xi})).
\label{eq:feature_inversion_dip}
\end{equation}
The feature encoder receives the full generated image with its standard
preprocessing. The generator is not pretrained on images.

\paragraph{Guided diffusion.}
A frozen unconditional ImageNet ADM~\citep{dhariwal2021diffusion} supplies
a learned image prior. Let $\mathbf{x}_t$ denote its noisy image state and
$\hat{\mathbf{x}}_0(\mathbf{x}_t,t)$ its clean-image prediction.
Following CLIP-guided diffusion,\footnote{\url{https://github.com/afiaka87/clip-guided-diffusion/tree/a631a06b51ac5c6636136fab27833c68862eaa24}.}
we evaluate feature matching on a clean/noisy blend:
\begin{equation}
\mathbf{u}_t=\rho_t\hat{\mathbf{x}}_0+(1-\rho_t)\mathbf{x}_t,
\qquad \rho_t=\sqrt{1-\bar\alpha_t^{\mathrm{ADM}}},
\qquad \mathbf{r}_t=(\mathbf{u}_t+1)/2,
\label{eq:feature_inversion_blend}
\end{equation}
where $\bar\alpha_t^{\mathrm{ADM}}$ is the cumulative signal-retention factor
and $\mathbf{r}_t$ is the corresponding RGB candidate.
Using the same $\operatorname{norm}$ as in the main text, define
$d_{\mathrm{sph}}(\mathbf{a},\mathbf{b})=
2\arcsin^2(\|\operatorname{norm}(\mathbf{a})-\operatorname{norm}(\mathbf{b})\|_2/2)$.
The guidance objective combines feature matching with image regularization:
\begin{equation}
\begin{aligned}
\mathcal{L}_{\mathrm{inv},t}(\mathbf{x}_t)
&=\frac{\lambda_f}{M_c}\sum_{m=1}^{M_c}
 \mathcal{D}_{i,d_{\mathrm{sph}}}(\mathcal{T}_m\mathbf{r}_t)\\
&\quad+\lambda_{\mathrm{TV}}\operatorname{TV}(\mathbf{u}_t)
 +\lambda_{\mathrm{range}}\mathcal{R}(\hat{\mathbf{x}}_0).
\end{aligned}
\label{eq:feature_inversion_diffusion_loss}
\end{equation}
Here $\operatorname{TV}$ averages squared horizontal and vertical pixel
differences, and
$\mathcal{R}(\mathbf{v})=\operatorname{mean}[(\mathbf{v}-\operatorname{clip}(\mathbf{v},-1,1))^2]$.
We use $M_c=64$ random candidate crops, resized to $224\times224$, and
$(\lambda_f,\lambda_{\mathrm{TV}},\lambda_{\mathrm{range}})=(1000,150,50)$.
The gradient passes through the frozen visual mappings and denoiser to
shift the reverse-process mean:
\begin{equation}
\boldsymbol{\mu}_t^{\mathrm{guided}}
=\boldsymbol{\mu}_t-\Sigma_t\nabla_{\mathbf{x}_t}
 \mathcal{L}_{\mathrm{inv},t}(\mathbf{x}_t).
\label{eq:feature_inversion_guidance}
\end{equation}
Ancestral sampling retains the pretrained reverse covariance $\Sigma_t$
and the linear noise schedule.

\paragraph{Comparison and interpretation.}
All branches share hyperparameters and sampling noise within each method.
Each inversion matches one global feature vector; crops, when used, come
only from the candidate image. This adapts DreamSim's analysis to global
features rather than reproducing its target-image crop conditioning.
The displayed stimuli were selected from eight correctly retrieved images
based on the visual quality of their image-conditioned DreamSim inversions;
failed EEG inversions remain included. These qualitative examples show
that stronger image priors can improve plausibility while changing
identity, pose, or layout
(Figures~\ref{fig:feature_inversion_optimization}--\ref{fig:feature_inversion_diffusion}).
They do not estimate average reconstruction performance or establish a
ranking of the information preserved by different representations.

\FloatBarrier

\subsection{Comprehensive Reconstruction Performance}
\label{sec:comprehensive_reconstruction_performance}

\subsubsection{Evaluation and controls}
\label{sec:reconstruction_ceilings}
We assess structural fidelity and semantic identification for brain-to-image reconstruction.
Because strong generative priors can yield plausible reconstructions from little neural information~\citep{mayo2024brainbits}, we report a permutation-based chance level and image-conditioned references alongside the brain-conditioned results.
Scores are averaged over 200 test images, three generations per image,
and ten subjects; image-conditioned controls are evaluated once on the
shared stimulus set. Deterministic VAE decoding produces one image per stimulus.
The CLIP identification metric uses ViT-L/14, distinct from the ViT-H/14
conditioning encoder.

\paragraph{Generator differences.}
\label{sec:generator_comparison}
Table~\ref{table:reconstruction_generator_protocols} summarizes the systems
compared in Table~\ref{table:generation_comparison}. Fine-tuning includes
the image-synthesis backbone or its pretrained conditioning interface.
AVDE~\citep{dai2026autoregressive} fine-tunes VAR~\citep{tian2024var}; MBP keeps the SDXL
backbone and VAE frozen while training visual fusion and fine-tuning the
IP-Adapter projection and attention modules on ImageNet~\citep{zheng2026learning}. ATM-S and CC-All learn embedding
priors~\citep{li2024visual,zhang2024cognitioncapturer}, whereas BHA trains
brain-side predictors while keeping SDXL-Turbo and IP-Adapter frozen.

\begin{table}[!htbp]
\centering
\caption{\textbf{Generator and sampling comparison.}
Fine-tuning includes the generation backbone or pretrained conditioning interface.}
\label{table:reconstruction_generator_protocols}
\footnotesize
\setlength{\tabcolsep}{3pt}
\renewcommand{\arraystretch}{1.0}
\begin{tabular}{@{}>{\raggedright\arraybackslash}p{0.08\linewidth}
                    >{\raggedright\arraybackslash}p{0.29\linewidth}
                    >{\centering\arraybackslash}p{0.09\linewidth}
                    >{\centering\arraybackslash}p{0.13\linewidth}
                    >{\raggedright\arraybackslash}p{0.35\linewidth}@{}}
\toprule
Method & Generator / interface & Fine-tuned? & Resolution & Synthesis steps \\
\midrule
ATM-S
& SDXL-Turbo + IP-Adapter & No & $512\times512$ & 4 denoising steps (released example) \\
CC-All
& SDXL-Turbo + IP-Adapters & No & $512\times512$\, & 5 denoising steps \\
AVDE
& VAR + VQ-VAE & Yes & $256\times256$\, & 10 autoregressive scales \\
MBP
& SDXL-Turbo + IP-Adapter & Yes & $512\times512$\, & 4 denoising steps \\
BHA
& SDXL-Turbo + IP-Adapter & No & $512\times512$ & 3 of 5 scheduled steps ($\alpha=0.75$) \\
\bottomrule
\end{tabular}
\end{table}

Code-derived settings follow the released implementations.\footnote{
\href{https://github.com/ncclab-sustech/EEG_Image_decode/blob/49cb3a20a0c1fd35fd2955f78d0e4d0f2939c70d/Generation/Generation_metrics_sub8.ipynb}{ATM-S},
\href{https://github.com/XiaoZhangYES/CognitionCapturer/blob/810fb904e2c2545de21e2202d4042b5f4a4b3106/src/Scripts/generation/multiadapter/main.py}{CC-All},
\href{https://github.com/ddicee/avde/blob/3be06a90ae69932c203bf4ae45d90803d8e752ed/utils/arg_util.py}{AVDE}, and
MBP (\href{https://github.com/ssshamiii/Brain-HIVE/blob/f06db69fcc4a029439e369c5c3dba6099e57560e/train_prior.py}{training},
\href{https://github.com/ssshamiii/Brain-HIVE/blob/f06db69fcc4a029439e369c5c3dba6099e57560e/build_reconstruction.py}{sampling}).}

\paragraph{Metrics.}
\label{sec:reconstruction_metrics}
Following ATM~\citep{li2024visual}, PixCorr is RGB pixel Pearson correlation and SSIM measures structural fidelity. AlexNet(2)/(5) use post-ReLU features from convolutional layers 2/5; these, InceptionV3, and CLIP report two-way identification: the fraction of distractors with lower feature correlation than the matched image (chance $0.5$). EffNet-B and SwAV report feature correlation distance, $1-r$. Higher is better except for these two distances.

\paragraph{Image-conditioned references.}
To assess generation with accurate conditioning, we replace brain
predictions with ground-truth OpenCLIP ViT-H/14 image features, VAE latents,
or both (Table~\ref{table:generation_comparison}). Semantic-only generation starts from noise;
spatial-only generation directly decodes the latent; their combination
uses $\alpha=0.5$. All use the frozen generator with IP-Adapter scale
$1.5$, classifier-free guidance scale $0$, and $512\times512$ output.
Denoising follows a five-step schedule.

\paragraph{Chance level.}
To test whether reconstruction depends on stimulus-specific predictions,
we jointly permute semantic and spatial conditions across the 200 test
images within each subject, so that no image retains its own conditions. Reconstructions are evaluated against the original targets,
using the same models, sampling settings, and random seeds as the main BHA row.

\subsubsection{Reconstruction ablations}

\paragraph{\texorpdfstring{\protect\redhl{Retrieval and reconstruction favor different training targets.}}{Retrieval and reconstruction favor different training targets.}}
\label{sec:visual_adapter_task_control}
\redhl{We test
whether one training target can serve both with a deliberately minimal model,
so that only two factors vary: the training objective, and whether the visual
target may adapt. For each of ten THINGS-EEG2 subjects, BrainViT is trained
from scratch to predict a single teacher, OpenCLIP ViT-H/14, without fusion or
teacher-specific readouts. The target is either the native feature
$\mathbf{e}_i=\phi(\mathbf{I}_i)$ or the adapted feature
$\tilde{\mathbf{e}}_i=V(\mathbf{e}_i)$, where
$V(\mathbf{e})=\mathbf{e}+\gamma W_{\mathrm{up}}\operatorname{GELU}(W_{\mathrm{down}}\operatorname{LN}(\mathbf{e}))$
is the visual adapter with rank 512 and 1.05M parameters, initialized near
identity and optimized jointly with BrainViT. The objective
is either contrastive, $\mathcal{L}_{\mathrm{BiCLIP}}$, or
regression-weighted, $0.9\mathcal{L}_{\mathrm{MSE}}+0.1\mathcal{L}_{\mathrm{BiCLIP}}$.
Paired models share initialization, data split, minibatch order, and
optimizer settings: AdamW, learning rate $3\times10^{-4}$, weight decay
$10^{-4}$, batch size 1,024, and 100 epochs; the final model is evaluated.
Retrieval ranks the 200 test images in each model's own target space. Reconstruction feeds the same prediction directly to the frozen
SDXL-Turbo and IP-Adapter, with no spatial condition and no learned mapping:
five denoising steps, IP-Adapter scale $1.5$, guidance scale $0$, and three
noise draws per image.}

\begin{table}[!htbp]
\centering
\caption{{\textbf{Retrieval and reconstruction favor different training targets.}
Single-teacher BrainViT predicting OpenCLIP ViT-H/14 features, trained from
scratch. Cos.: paired brain--target cosine similarity. Retrieval: Raw 200-way
accuracy (\%) in each model's target space. Reconstruction: the same
predictions passed to the frozen generator. THINGS-EEG2 within-subject means
over ten subjects; bold marks the best value in each column.}}
\label{table:visual_adapter_task_control}
\footnotesize
\setlength{\tabcolsep}{1.5pt}
\renewcommand{\arraystretch}{1.05}
\begin{tabular*}{\linewidth}{@{\extracolsep{\fill}}l*{11}{c}@{}}
\toprule
& & \multicolumn{2}{c}{Retrieval (\%)} & \multicolumn{8}{c}{Reconstruction} \\
\cmidrule(lr){3-4}\cmidrule(lr){5-12}
Visual target & Cos.$\uparrow$ & Top-1$\uparrow$ & Top-5$\uparrow$ & PixCorr$\uparrow$ & SSIM$\uparrow$ & \shortstack{AlexNet\\(2)$\uparrow$} & \shortstack{AlexNet\\(5)$\uparrow$} & \shortstack{Inception\\V3$\uparrow$} & CLIP$\uparrow$ & EffNet-B$\downarrow$ & SwAV$\downarrow$ \\
\midrule
\multicolumn{12}{@{}l}{\textit{Contrastive objective,} $\mathcal{L}_{\mathrm{BiCLIP}}$} \\
\quad Native   & 0.217 & 30.05 & 60.70 & \textbf{0.168} & 0.391 & 0.779 & 0.823 & 0.725 & 0.754 & 0.900 & 0.612 \\
\quad Adapted  & 0.481 & \textbf{37.80} & \textbf{69.50} & 0.160 & 0.389 & 0.764 & 0.809 & 0.685 & 0.706 & 0.917 & 0.635 \\
\addlinespace[2pt]
\multicolumn{12}{@{}l}{\textit{Regression-weighted objective,} $0.9\mathcal{L}_{\mathrm{MSE}}+0.1\mathcal{L}_{\mathrm{BiCLIP}}$} \\
\quad Native   & 0.583 & 21.10 & 50.30 & 0.167 & \textbf{0.414} & \textbf{0.830} & \textbf{0.902} & \textbf{0.793} & \textbf{0.838} & \textbf{0.858} & \textbf{0.533} \\
\quad Adapted  & 0.598 & 19.00 & 44.45 & 0.099 & 0.370 & 0.657 & 0.665 & 0.550 & 0.541 & 0.968 & 0.683 \\
\bottomrule
\end{tabular*}
\end{table}

\redhl{Table~\ref{table:visual_adapter_task_control} shows two effects.}

\redhl{\emph{The objective trades one task for the other.} With native targets, the
contrastive objective raises Raw Top-1 from $21.10\%$ to $30.05\%$ but
reconstructs worse on seven of eight metrics, with CLIP identification
falling from $0.838$ to $0.754$. Regression keeps predictions in the
generator's coordinates, whereas the contrastive objective constrains only
their direction relative to other images.}

\redhl{\emph{Adaptation helps retrieval and hurts reconstruction.} Under the
contrastive objective, letting the target adapt raises paired cosine from
$0.217$ to $0.481$ and Raw Top-1 from $30.05\%$ to $37.80\%$, but lowers every
reconstruction metric, with CLIP identification falling from $0.754$ to
$0.706$. This is the pattern described by
Proposition~\ref{prop:alignment_reconstruction_gap}: adaptation reduces the
angular error $\delta$ to the adapted target but adds the drift $\Delta_G$
away from the native features the generator expects. When regression
dominates, adaptation hurts both tasks, because the regression target itself
moves away from the native feature.}

\redhl{The best retrieval thus comes from a contrastive objective with an adapted
target, and the best reconstruction from regression to the native target; no
single configuration is best at both.}

\begin{tcolorbox}[colback=gray!8,colframe=gray!8,boxrule=0pt,arc=0pt,
left=7pt,right=7pt,top=4pt,bottom=4pt,before skip=6pt,after skip=8pt]
\redhl{\textbf{Takeaway.} Retrieval rewards a discriminative geometry that is free
to adapt to the brain; reconstruction rewards staying in the frozen
generator's native coordinates. One target and one objective cannot serve
both. BHA therefore gives each task its own parameters and objective:
adapted, composed multi-teacher spaces trained contrastively for retrieval,
and native-space semantic and spatial predictors for reconstruction, all on
one shared brain backbone.}
\end{tcolorbox}
\FloatBarrier

\paragraph{\texorpdfstring{\protect\redhl{Fusion-prior control.}}{Fusion-prior control.}}
\redhl{To isolate native-space conditioning from a learned fusion prior, we train
an MBP-style prior~\citep{zheng2026learning} on the fused targets of BHA's four
teachers for the 14{,}886 training images, following MBP's released code: BHA’s frozen fused target feeds a new IP-Adapter projection and cross-attention layers (349M trained parameters), initialized from the SDXL ViT-H IP-Adapter and trained on
SDXL-base~\citep{podell2024sdxl} with the UNet and VAE frozen. We use MBP's optimizer settings (AdamW,
learning rate $10^{-4}$, weight decay 0.01, condition dropout 0.1, batch size 8)
for six epochs, after which the validation denoising loss is flat. Native
conditioning wins all six mid- and high-level metrics without training any
generator-side parameters (Table~\ref{table:fusion_prior_control}). The prior wins only the two pixel-level metrics and stays behind even when the
images' own fused features replace its semantic input, so the gap is not
brain-prediction error: a fusion prior has to learn a new conditioning
interface, here from the 14{,}886 training images of a brain dataset, whereas
native conditioning reuses the interface the generator was trained with.}

\begin{table}[!htbp]\redhlon
    \centering
    \caption{\textbf{Native-space conditioning versus an MBP-style fusion prior.}
    THINGS-EEG2, ten-subject means. All rows share the brain predictions, the
    SDXL-Turbo sampler ($\alpha=0.75$, 5 steps, IP-Adapter scale 1.5), prompts,
    seeds, and evaluator; only the high-level condition differs. \emph{GT}: the images' own fused features replace the predicted semantic
    condition; all rows start from the same brain-predicted VAE latent. \emph{Params}: trained
    generator-side parameters.}
    \label{table:fusion_prior_control}
    \small\setlength{\tabcolsep}{2pt}
    \resizebox{\fitwidth}{!}{%
    \begin{tabular}{lccccccccc}
    \toprule
    Condition & Params & PixCorr$\uparrow$ & SSIM$\uparrow$ & AlexNet(2)$\uparrow$ & AlexNet(5)$\uparrow$ & Inception$\uparrow$ & CLIP$\uparrow$ & EffNet-B$\downarrow$ & SwAV$\downarrow$ \\
    \midrule
    \rowcolor{gray!10}
    BHA (native) & 0 & 0.274 & 0.459 & \textbf{0.872} & \textbf{0.921} & \textbf{0.794} & \textbf{0.832} & \textbf{0.843} & \textbf{0.515} \\
    Fusion prior & 349M & 0.301 & 0.489 & 0.801 & 0.835 & 0.702 & 0.740 & 0.898 & 0.596 \\
    Fusion prior (GT) & 349M & \textbf{0.324} & \textbf{0.508} & 0.787 & 0.831 & 0.737 & 0.777 & 0.873 & 0.588 \\
    \bottomrule
    \end{tabular}}
\end{table}

\paragraph{VAE guidance.}
We test whether the predicted spatial latent can preserve structure
throughout denoising by comparing standard sampling with the training-free
guidance in Appendix~\ref{sec:vae_guidance_details}. Guidance settings were selected
on Sub-01 and applied unchanged to the other nine subjects. Guidance
modestly improves pixel and structural fidelity while leaving CLIP
identification largely unchanged (Table~\ref{table:generation_comparison}).

\paragraph{Pretraining-source comparison.}
\label{sec:reconstruction_pretraining_control}
To test whether richer pretraining supervision benefits reconstruction,
we compare BrainViT trained against frozen OpenCLIP ViT-H/14 features with
BrainViT trained against BHA's fused visual targets
(Table~\ref{table:reconstruction_ablation_main}).
Both use the same reconstruction architecture, training schedules, data splits,
and model-selection criteria. Reconstruction heads read the pretrained
backbone directly. We evaluate ten subjects at $\alpha\in\{0,0.75,1\}$,
with the generation settings above and three generations per test image.

Fused pretraining improves SSIM at all three strengths, whereas OpenCLIP
has higher CLIP identification at $\alpha=0.75$ and $1$. Additional visual
supervision therefore benefits some reconstruction properties without
uniformly improving semantic identification.

\begin{table}[!t]
\centering
\caption{\textbf{Reconstruction: OpenCLIP vs.\ Fused VFM pretraining.} Ten-subject means over 200 test images, with three generations per image for $\alpha>0$ and deterministic decoding for $\alpha=0$. $\alpha=0/1$: spatial-/semantic-only conditions.}
\label{table:reconstruction_ablation_main}
\footnotesize
\setlength{\tabcolsep}{2pt}
\renewcommand{\arraystretch}{0.92}
\begin{tabular}{@{}cl*{8}{r}@{}}
\toprule
$\alpha$ & Pretraining & PixCorr$\uparrow$ & SSIM$\uparrow$ & Alex(2)$\uparrow$ & Alex(5)$\uparrow$ & Incep.$\uparrow$ & CLIP$\uparrow$ & EffNet$\downarrow$ & SwAV$\downarrow$ \\
\midrule
\multirow{2}{*}{0} & OpenCLIP & 0.333 & 0.580 & 0.604 & 0.565 & 0.507 & 0.514 & 0.985 & 0.689 \\
& Fused VFM & 0.335 & 0.583 & 0.599 & 0.561 & 0.507 & 0.515 & 0.984 & 0.687 \\
\addlinespace[1pt]
\multirow{2}{*}{0.75} & OpenCLIP & 0.274 & 0.448 & 0.873 & 0.922 & 0.798 & 0.841 & 0.835 & 0.516 \\
& \cellcolor{gray!10}Fused VFM & \cellcolor{gray!10}0.274 & \cellcolor{gray!10}0.459 & \cellcolor{gray!10}0.872 & \cellcolor{gray!10}0.921 & \cellcolor{gray!10}0.794 & \cellcolor{gray!10}0.831 & \cellcolor{gray!10}0.843 & \cellcolor{gray!10}0.515 \\
\addlinespace[1pt]
\multirow{2}{*}{1} & OpenCLIP & 0.189 & 0.413 & 0.849 & 0.915 & 0.799 & 0.839 & 0.837 & 0.519 \\
& Fused VFM & 0.185 & 0.427 & 0.845 & 0.910 & 0.790 & 0.828 & 0.844 & 0.520 \\
\bottomrule
\end{tabular}
\end{table}

\paragraph{Reconstruction strength.}
We examine the balance between preserving the predicted spatial latent
and allowing semantic denoising by varying
$\alpha\in\{0,0.25,0.5,0.75,1\}$ with the same predictors and random seeds
(Table~\ref{table:reconstruction_alpha_subjectwise}).
The corresponding numbers of denoising updates are $0,1,2,3,5$;
all other settings follow Appendix~\ref{sec:reconstruction_implementation}.
Avg and SD summarize ten subjects, and rankings use unrounded scores.
Lower strengths favor pixel and structural fidelity, whereas the default
$\alpha=0.75$ achieves the highest mean scores on most feature-based metrics.

\begingroup
\setlength{\LTleft}{\fill}
\setlength{\LTright}{\fill}
\setlength{\LTcapwidth}{\linewidth}
\setlength{\tabcolsep}{1pt}
\begin{longtable}{>{\footnotesize}l>{\footnotesize}l*{8}{>{\footnotesize}c}}
\caption{\textbf{Reconstruction strength sweep on THINGS-EEG2.}}
\label{table:reconstruction_alpha_subjectwise}\\
\toprule
\textbf{Subject} & $\alpha$ & \textbf{PixCorr}$\uparrow$ & \textbf{SSIM}$\uparrow$ & \textbf{AlexNet(2)}$\uparrow$ & \textbf{AlexNet(5)}$\uparrow$ & \textbf{InceptionV3}$\uparrow$ & \textbf{CLIP}$\uparrow$ & \textbf{EffNet-B}$\downarrow$ & \textbf{SwAV}$\downarrow$ \\
\midrule
\endfirsthead
\toprule
\textbf{Subject} & $\alpha$ & \textbf{PixCorr}$\uparrow$ & \textbf{SSIM}$\uparrow$ & \textbf{AlexNet(2)}$\uparrow$ & \textbf{AlexNet(5)}$\uparrow$ & \textbf{InceptionV3}$\uparrow$ & \textbf{CLIP}$\uparrow$ & \textbf{EffNet-B}$\downarrow$ & \textbf{SwAV}$\downarrow$ \\
\midrule
\endhead
\midrule
\multicolumn{10}{r}{Continued on next page} \\
\endfoot
\bottomrule
\endlastfoot
sub-01 & 0 & \underline{0.3514} & \textbf{0.5831} & 0.6086 & 0.5713 & 0.4969 & 0.5106 & 0.9864 & 0.6859 \\
sub-01 & 0.25 & \textbf{0.3889} & \underline{0.5739} & 0.7526 & 0.8240 & 0.7031 & 0.7322 & 0.8956 & 0.5794 \\
sub-01 & 0.5 & 0.3498 & 0.5226 & \underline{0.8693} & 0.9060 & 0.7589 & 0.8098 & 0.8559 & 0.5285 \\
sub-01 & \textbf{0.75 (Ours)} & 0.2983 & 0.4607 & \textbf{0.8919} & \textbf{0.9226} & \underline{0.7788} & \textbf{0.8229} & \underline{0.8425} & \textbf{0.5216} \\
sub-01 & 1 & 0.2073 & 0.4293 & 0.8612 & \underline{0.9119} & \textbf{0.7814} & \underline{0.8212} & \textbf{0.8392} & \underline{0.5228} \\
\midrule
sub-02 & 0 & \underline{0.3510} & \textbf{0.5846} & 0.5953 & 0.5582 & 0.5110 & 0.5106 & 0.9844 & 0.6885 \\
sub-02 & 0.25 & \textbf{0.3877} & \underline{0.5739} & 0.7633 & 0.8448 & 0.7440 & 0.7477 & 0.8887 & 0.5735 \\
sub-02 & 0.5 & 0.3304 & 0.5125 & \underline{0.8692} & 0.9220 & 0.7901 & \underline{0.8265} & 0.8619 & \underline{0.5210} \\
sub-02 & \textbf{0.75 (Ours)} & 0.2669 & 0.4393 & \textbf{0.8804} & \textbf{0.9319} & \underline{0.7922} & \textbf{0.8294} & \textbf{0.8576} & \textbf{0.5173} \\
sub-02 & 1 & 0.1866 & 0.4038 & 0.8610 & \underline{0.9241} & \textbf{0.8043} & 0.8251 & \underline{0.8578} & 0.5218 \\
\midrule
sub-03 & 0 & 0.3266 & \textbf{0.5801} & 0.5945 & 0.5601 & 0.5211 & 0.5163 & 0.9829 & 0.6865 \\
sub-03 & 0.25 & \textbf{0.3762} & \underline{0.5730} & 0.7519 & 0.8198 & 0.7182 & 0.7540 & 0.8917 & 0.5862 \\
sub-03 & 0.5 & \underline{0.3345} & 0.5200 & \underline{0.8520} & 0.9087 & 0.7850 & 0.8263 & 0.8498 & 0.5258 \\
sub-03 & \textbf{0.75 (Ours)} & 0.2766 & 0.4458 & \textbf{0.8640} & \textbf{0.9234} & \textbf{0.8092} & \textbf{0.8382} & \textbf{0.8427} & \textbf{0.5202} \\
sub-03 & 1 & 0.1841 & 0.4060 & 0.8351 & \underline{0.9088} & \underline{0.7879} & \underline{0.8299} & \underline{0.8490} & \underline{0.5220} \\
\midrule
sub-04 & 0 & 0.2902 & \textbf{0.5790} & 0.5697 & 0.5443 & 0.5089 & 0.5123 & 0.9853 & 0.6842 \\
sub-04 & 0.25 & \textbf{0.3386} & \underline{0.5679} & 0.7289 & 0.8315 & 0.7653 & 0.7799 & 0.8747 & 0.5654 \\
sub-04 & 0.5 & \underline{0.2967} & 0.5108 & \underline{0.8361} & 0.8960 & 0.7918 & 0.8365 & 0.8462 & 0.5170 \\
sub-04 & \textbf{0.75 (Ours)} & 0.2485 & 0.4381 & \textbf{0.8584} & \textbf{0.9172} & \underline{0.8021} & \underline{0.8509} & \underline{0.8395} & \textbf{0.5115} \\
sub-04 & 1 & 0.1766 & 0.4020 & 0.8276 & \underline{0.9123} & \textbf{0.8050} & \textbf{0.8570} & \textbf{0.8380} & \underline{0.5154} \\
\midrule
sub-05 & 0 & \underline{0.3226} & \textbf{0.5816} & 0.5821 & 0.5535 & 0.4978 & 0.5127 & 0.9857 & 0.6866 \\
sub-05 & 0.25 & \textbf{0.3627} & \underline{0.5697} & 0.7394 & 0.8006 & 0.7127 & 0.7326 & 0.8977 & 0.5817 \\
sub-05 & 0.5 & 0.3145 & 0.5100 & \underline{0.8332} & 0.8757 & \underline{0.7359} & 0.7885 & 0.8778 & 0.5381 \\
sub-05 & \textbf{0.75 (Ours)} & 0.2657 & 0.4420 & \textbf{0.8476} & \underline{0.8923} & 0.7260 & \textbf{0.7941} & \underline{0.8717} & \underline{0.5343} \\
sub-05 & 1 & 0.1815 & 0.4107 & 0.8273 & \textbf{0.8972} & \textbf{0.7417} & \underline{0.7918} & \textbf{0.8668} & \textbf{0.5331} \\
\midrule
sub-06 & 0 & \underline{0.3559} & \textbf{0.5829} & 0.6021 & 0.5702 & 0.5005 & 0.5137 & 0.9823 & 0.6857 \\
sub-06 & 0.25 & \textbf{0.3927} & \underline{0.5730} & 0.7707 & 0.8260 & 0.7221 & 0.7424 & 0.8883 & 0.5748 \\
sub-06 & 0.5 & 0.3407 & 0.5181 & \underline{0.8657} & 0.9109 & 0.7707 & 0.8114 & 0.8579 & 0.5279 \\
sub-06 & \textbf{0.75 (Ours)} & 0.2812 & 0.4443 & \textbf{0.8841} & \textbf{0.9225} & \textbf{0.7827} & \underline{0.8178} & \textbf{0.8528} & \textbf{0.5240} \\
sub-06 & 1 & 0.2138 & 0.4037 & 0.8484 & \underline{0.9114} & \underline{0.7757} & \textbf{0.8191} & \underline{0.8536} & \underline{0.5247} \\
\midrule
sub-07 & 0 & \underline{0.3183} & \textbf{0.5801} & 0.5969 & 0.5636 & 0.5063 & 0.5133 & 0.9826 & 0.6877 \\
sub-07 & 0.25 & \textbf{0.3590} & \underline{0.5722} & 0.7541 & 0.8231 & 0.7249 & 0.7417 & 0.8884 & 0.5803 \\
sub-07 & 0.5 & 0.3182 & 0.5237 & \underline{0.8559} & 0.8968 & \textbf{0.7957} & 0.8080 & 0.8519 & 0.5303 \\
sub-07 & \textbf{0.75 (Ours)} & 0.2614 & 0.4594 & \textbf{0.8709} & \textbf{0.9154} & 0.7886 & \underline{0.8114} & \textbf{0.8489} & \textbf{0.5230} \\
sub-07 & 1 & 0.1578 & 0.4206 & 0.8259 & \underline{0.9000} & \underline{0.7938} & \textbf{0.8142} & \underline{0.8506} & \underline{0.5266} \\
\midrule
sub-08 & 0 & \underline{0.3627} & \textbf{0.5846} & 0.6236 & 0.5875 & 0.5078 & 0.5219 & 0.9812 & 0.6854 \\
sub-08 & 0.25 & \textbf{0.3955} & \underline{0.5759} & 0.7814 & 0.8441 & 0.7488 & 0.7768 & 0.8756 & 0.5685 \\
sub-08 & 0.5 & 0.3551 & 0.5291 & \underline{0.8719} & 0.9225 & 0.7985 & \underline{0.8597} & 0.8281 & \underline{0.5041} \\
sub-08 & \textbf{0.75 (Ours)} & 0.3052 & 0.4757 & \textbf{0.8908} & \textbf{0.9379} & \underline{0.8157} & \textbf{0.8625} & \textbf{0.8157} & \textbf{0.4974} \\
sub-08 & 1 & 0.2174 & 0.4473 & 0.8557 & \underline{0.9237} & \textbf{0.8213} & 0.8532 & \underline{0.8222} & 0.5049 \\
\midrule
sub-09 & 0 & \underline{0.3705} & \textbf{0.5833} & 0.6243 & 0.5639 & 0.4930 & 0.5102 & 0.9827 & 0.6851 \\
sub-09 & 0.25 & \textbf{0.4105} & \underline{0.5759} & 0.7562 & 0.8217 & 0.7260 & 0.7287 & 0.8797 & 0.5713 \\
sub-09 & 0.5 & 0.3649 & 0.5314 & \underline{0.8578} & 0.9054 & 0.7755 & 0.8000 & 0.8573 & 0.5262 \\
sub-09 & \textbf{0.75 (Ours)} & 0.3008 & 0.4698 & \textbf{0.8738} & \textbf{0.9232} & \underline{0.7778} & \textbf{0.8052} & \underline{0.8542} & \textbf{0.5187} \\
sub-09 & 1 & 0.1919 & 0.4302 & 0.8382 & \underline{0.9090} & \textbf{0.7786} & \underline{0.8037} & \textbf{0.8533} & \underline{0.5235} \\
\midrule
sub-10 & 0 & \underline{0.3306} & \textbf{0.5827} & 0.6177 & 0.5727 & 0.5136 & 0.5158 & 0.9856 & 0.6919 \\
sub-10 & 0.25 & \textbf{0.3619} & \underline{0.5750} & 0.7448 & 0.8340 & 0.7687 & 0.7753 & 0.8747 & 0.5675 \\
sub-10 & 0.5 & 0.3220 & 0.5248 & \underline{0.8614} & 0.9129 & 0.8213 & \underline{0.8570} & 0.8291 & \underline{0.5032} \\
sub-10 & \textbf{0.75 (Ours)} & 0.2702 & 0.4666 & \textbf{0.8783} & \textbf{0.9324} & \underline{0.8351} & \textbf{0.8670} & \textbf{0.8217} & \textbf{0.4971} \\
sub-10 & 1 & 0.1762 & 0.4324 & 0.8548 & \underline{0.9203} & \textbf{0.8356} & 0.8562 & \underline{0.8281} & 0.5076 \\
\midrule
\textbf{Avg} & 0 & \underline{0.3380} & \textbf{0.5822} & 0.6015 & 0.5645 & 0.5057 & 0.5137 & 0.9839 & 0.6867 \\
\textit{SD} & & 0.0246 & 0.0019 & 0.0177 & 0.0118 & 0.0086 & 0.0035 & 0.0018 & 0.0022 \\
\textbf{Avg} & 0.25 & \textbf{0.3774} & \underline{0.5730} & 0.7543 & 0.8270 & 0.7334 & 0.7511 & 0.8855 & 0.5749 \\
\textit{SD} & & 0.0216 & 0.0026 & 0.0151 & 0.0128 & 0.0222 & 0.0196 & 0.0087 & 0.0069 \\
\textbf{Avg} & 0.5 & 0.3327 & 0.5203 & \underline{0.8573} & 0.9057 & 0.7823 & 0.8224 & 0.8516 & 0.5222 \\
\textit{SD} & & 0.0207 & 0.0074 & 0.0135 & 0.0138 & 0.0236 & 0.0235 & 0.0148 & 0.0112 \\
\textbf{Avg} & \textbf{0.75 (Ours)} & 0.2775 & 0.4542 & \textbf{0.8740} & \textbf{0.9219} & \underline{0.7908} & \textbf{0.8300} & \textbf{0.8447} & \textbf{0.5165} \\
\textit{SD} & & 0.0187 & 0.0139 & 0.0142 & 0.0125 & 0.0291 & 0.0245 & 0.0166 & 0.0117 \\
\textbf{Avg} & 1 & 0.1893 & 0.4186 & 0.8435 & \underline{0.9119} & \textbf{0.7925} & \underline{0.8271} & \underline{0.8459} & \underline{0.5202} \\
\textit{SD} & & 0.0187 & 0.0157 & 0.0143 & 0.0091 & 0.0262 & 0.0223 & 0.0138 & 0.0086 \\
\end{longtable}
\endgroup

\FloatBarrier
\subsection{\texorpdfstring{\protect\redhl{Transfer to fMRI: NSD}}{Transfer to fMRI: NSD}}
\label{sec:nsd_results}
\redhl{This subsection and the next test whether BHA transfers beyond EEG and MEG.}

\paragraph{\texorpdfstring{\protect\redhl{Setup.}}{Setup.}}
\redhl{We reuse the THINGS-MEG recipe, its hyperparameters, and the four
teachers, and change only the input tokenizer. A subject-specific linear layer
maps each subject's voxels to a $128\times32$ token grid, whose 4{,}096 values
match the width of MindEye2's shared latent~\citep{scotti2024mindeye2}, and the
grid enters the same BrainViT. Subject tokens are disabled, so this layer is the
only subject-specific module, and one model is trained on all eight subjects.
NSD has no concepts, so each image forms a class: a grouped batch holds the
three single trials of 85 images (255 rows), and the trial average of each image
forms the averaged auxiliary row. Validation holds out 10\% of each subject's
non-shared images with repetition-averaged queries; every stage selects its
checkpoint on this split. We report Raw
Top-1 for Brain$\rightarrow$Image and Image$\rightarrow$Brain retrieval,
300-way averaged over 30 random draws as in MindEye2 and 1{,}000-way over the
full test set. 2-way identification uses all 1{,}000 test images. Results are
means over three seeds and subjects 1, 2, 5, and 7.}

\redhl{\paragraph{Normalization and training setting.}
The joint model z-scores each voxel within each session over all trials of
that session, test trials included but not their labels. To separate this
choice and joint training from the results, we also follow the setup of
MindEye2's Table~1: pretraining on the other seven subjects, fine-tuning on the
target subject's 40 sessions, and z-scoring with training-trial statistics only.}

\begin{table}[!htbp]\redhlon
\centering
\caption{\textbf{Composition on NSD.} Raw Top-1 (\%), mean over subjects 1, 2,
5, and 7 and three seeds, joint model. \emph{Stage 1}: one fused space;
\emph{Shared}: the shared space only; \emph{BHA}: shared and teacher-specific
scores composed.}
\label{table:nsd_composition}
\small
\begin{tabular}{lcccc}
\toprule
Direction & Test & Stage 1 & Shared & BHA \\
\midrule
Brain$\rightarrow$Image & 300-way & 92.4 & 92.8 & \textbf{92.9} \\
Image$\rightarrow$Brain & 300-way & 93.0 & 93.6 & \textbf{94.1} \\
Brain$\rightarrow$Image & 1{,}000-way & 84.6 & 85.4 & \textbf{85.5} \\
Image$\rightarrow$Brain & 1{,}000-way & 85.3 & 86.3 & \textbf{87.2} \\
\bottomrule
\end{tabular}
\end{table}

\paragraph{\texorpdfstring{\protect\redhl{Comparison.}}{Comparison.}}
\redhl{Table~\ref{table:nsd_comparison} compares BHA with the methods in
MindEye2's Table~1. BHA gives the highest CLIP identification and ranks second on
retrieval, SSIM, and SwAV; MindEye2 leads the remaining metrics. The two BHA rows
differ by at most 0.3 points on retrieval and identification and by at most
0.004 on the other metrics, so the joint results depend neither on test-trial
normalization nor on joint training, and the gap to MindEye2 persists under its
own protocol. On the generator side, MindEye2 fine-tunes SDXL into an unCLIP
generator, whereas BHA trains no generator parameters. Subject 7 is the hardest
subject under both protocols, with about 87\% 300-way image retrieval against
94--96\% for the other three.}

\begin{table}[!htbp]\redhlon
\centering
\caption{\textbf{Comparison on NSD.} Subjects 1, 2, 5, and 7. Retrieval is
300-way Raw Top-1 (\%). \emph{Joint}: one model for all eight
subjects; \emph{MindEye2 protocol}: pretrained on the other seven subjects,
fine-tuned on the target subject, and normalized with training-trial statistics
only. Other rows are as reported by MindEye2~\citep{scotti2024mindeye2}:
\citet{takagi2023high}, Brain Diffuser~\citep{Ozcelik_2023}, and
MindEye~\citep{scotti2024reconstructing}.}
\label{table:nsd_comparison}
\small\setlength{\tabcolsep}{2pt}
\resizebox{\fitwidth}{!}{%
\begin{tabular}{lcccccccccc}
\toprule
Method & Image$\uparrow$ & Brain$\uparrow$ & PixCorr$\uparrow$ & SSIM$\uparrow$ & AlexNet(2)$\uparrow$ & AlexNet(5)$\uparrow$ & Inception$\uparrow$ & CLIP$\uparrow$ & EffNet-B$\downarrow$ & SwAV$\downarrow$ \\
\midrule
Takagi \& Nishimoto & -- & -- & 0.246 & 0.410 & 0.789 & 0.856 & 0.838 & 0.821 & 0.811 & 0.504 \\
Brain Diffuser & 18.8 & 26.3 & 0.273 & 0.365 & 0.944 & 0.966 & 0.913 & 0.909 & 0.728 & 0.421 \\
MindEye & 84.1 & 90.0 & 0.319 & 0.360 & 0.928 & 0.969 & 0.946 & 0.933 & 0.648 & 0.377 \\
MindEye2 & \textbf{98.3} & \textbf{98.8} & \textbf{0.322} & \textbf{0.431} & \textbf{0.961} & \textbf{0.986} & \textbf{0.954} & 0.930 & \textbf{0.619} & \textbf{0.344} \\
\midrule
\rowcolor{gray!10}
BHA, joint subjects & 92.9 & 94.1 & 0.239 & 0.419 & 0.883 & 0.959 & 0.941 & \textbf{0.947} & 0.656 & 0.373 \\
\rowcolor{gray!10}
BHA, MindEye2 protocol & 93.2 & 94.1 & 0.235 & 0.419 & 0.881 & 0.957 & 0.941 & 0.946 & 0.659 & 0.374 \\
\bottomrule
\end{tabular}}
\end{table}

\FloatBarrier
\subsection{\texorpdfstring{\protect\redhl{Transfer to Intracortical Recordings: TVSD}}{Transfer to Intracortical Recordings: TVSD}}
\label{sec:tvsd_results}

\paragraph{\texorpdfstring{\protect\redhl{Setup.}}{Setup.}}
\redhl{We reuse the THINGS-MEG recipe, its hyperparameters, and the four teachers
unchanged, and train one model per animal. The valid multi-unit channels form the
sensor axis, with five time bins each. Without a sensor montage, learned channel
embeddings replace the montage-based positional encoding. We follow the zero-shot
split of \citet{Song_2026}, so no test concept is seen in training. TVSD follows
the THINGS concept structure (1{,}754 training concepts of 12 images after the
split), so grouped batches, the concept-average auxiliary, and the 10\% held-out-concept
validation split match THINGS-MEG. Test queries average the 30 repetitions of each
test image, and test responses are z-scored with training statistics of the same
recording day. Reconstruction follows the THINGS-MEG recipe with the generator
frozen and three draws per test image; 2-way identification uses the 100 test
images. Results are three-seed means.}

\paragraph{\texorpdfstring{\protect\redhl{Results.}}{Results.}}
\redhl{We compare with FuzzyAlign~\citep{Song_2026}, a published MUA decoder
evaluated on the same zero-shot split with 100-way tests and repetition-averaged
queries. Retrieval stays near ceiling for every BHA stage
(Table~\ref{table:tvsd_retrieval}): Brain$\rightarrow$Image Top-1 is 100.0\%
against 62.8\% for FuzzyAlign, and Top-5 is 100.0\% against 87.7\%.
\citet{Ciferri_2026}, trained on the official split, report 69.3\% Top-1 and
93.6\% Top-5. TVSD thus
tests transfer to spiking data rather than the composition gain. With the same
reconstruction recipe and no generator training, BHA improves on both FuzzyAlign
pipelines in all eight metrics for both monkeys
(Table~\ref{table:tvsd_reconstruction}); CLIP identification is 0.862 against
0.814 for monkey N and 0.897 against 0.863 for monkey F. We also list Monkey
Perceptogram~\citep{Fei_2025} and \citet{Ciferri_2026}, which train on the official
split, where test concepts appear in training; this setting favours them. On the two-monkey mean that
\citet{Ciferri_2026} report, BHA is lower on Inception (0.864 against 0.868) and
equal on CLIP (0.879), and better on the other metrics. With THINGS-EEG2 and THINGS-MEG in the main results, BHA thus covers
EEG, MEG, fMRI, and intracortical recordings with one recipe.}

\begin{table}[!htb]\redhlon
\centering
\caption{\textbf{Retrieval on TVSD.} 100-way Raw Top-1 (\%), mean over monkeys
N and F. FuzzyAlign~\citep{Song_2026} and \citet{Ciferri_2026} are retrieval
results as reported (five seeds each); \citet{Ciferri_2026}$^\dagger$ trains on the
official split, in which test concepts appear in training, and the other columns
use the zero-shot split. The BHA columns are as in
Table~\ref{table:nsd_composition} (three seeds).}
\label{table:tvsd_retrieval}
\small
\begin{tabular}{lccccc}
\toprule
Direction & Ciferri et al.$^\dagger$ & FuzzyAlign & Stage 1 & Shared & BHA \\
\midrule
Brain$\rightarrow$Image & 69.3 & 62.8 & 98.2 & 99.7 & \textbf{100.0} \\
Image$\rightarrow$Brain & -- & -- & 98.8 & 99.0 & 99.0 \\
\bottomrule
\end{tabular}
\end{table}

\begin{table}[!htb]\redhlon
\centering
\caption{\textbf{Reconstruction on TVSD.} BHA and FuzzyAlign~\citep{Song_2026} use
the zero-shot split; methods marked $^\dagger$ train on the official split, in which
test concepts appear in training. Values of other methods are as reported
(Monkey Perceptogram~\citep{Fei_2025}; FuzzyAlign: high- and low-level pipelines,
five seeds; \citet{Ciferri_2026}: mean
over the two monkeys only). BHA uses three draws per test image and three seeds;
its generator is not trained. Best per group in bold.}
\label{table:tvsd_reconstruction}
\small\setlength{\tabcolsep}{2pt}
\resizebox{\fitwidth}{!}{%
\begin{tabular}{llcccccccc}
\toprule
Monkey & Method & PixCorr$\uparrow$ & SSIM$\uparrow$ & AlexNet(2)$\uparrow$ & AlexNet(5)$\uparrow$ & Inception$\uparrow$ & CLIP$\uparrow$ & EffNet-B$\downarrow$ & SwAV$\downarrow$ \\
\midrule
\multirow{4}{*}{N} & Monkey Perceptogram$^\dagger$ & 0.218 & 0.417 & 0.880 & 0.903 & 0.734 & 0.818 & 0.867 & 0.538 \\
 & FuzzyAlign (high) & 0.145 & 0.394 & 0.827 & 0.921 & 0.771 & 0.814 & 0.851 & 0.549 \\
 & FuzzyAlign (low) & 0.207 & 0.427 & 0.842 & 0.914 & 0.763 & 0.806 & 0.850 & 0.552 \\
 & BHA & \textbf{0.266} & \textbf{0.466} & \textbf{0.911} & \textbf{0.961} & \textbf{0.858} & \textbf{0.862} & \textbf{0.791} & \textbf{0.495} \\
\midrule
\multirow{4}{*}{F} & Monkey Perceptogram$^\dagger$ & 0.214 & 0.414 & 0.862 & 0.921 & 0.722 & 0.815 & 0.859 & 0.520 \\
 & FuzzyAlign (high) & 0.168 & 0.379 & 0.838 & 0.941 & 0.827 & 0.863 & 0.821 & 0.533 \\
 & FuzzyAlign (low) & 0.235 & 0.421 & 0.871 & 0.928 & 0.802 & 0.852 & 0.817 & 0.526 \\
 & BHA & \textbf{0.270} & \textbf{0.464} & \textbf{0.915} & \textbf{0.966} & \textbf{0.871} & \textbf{0.897} & \textbf{0.763} & \textbf{0.473} \\
\midrule
\multirow{4}{*}{Mean} & Ciferri et al.$^\dagger$ & 0.140 & 0.364 & -- & -- & \textbf{0.868} & \textbf{0.879} & 0.792 & 0.492 \\
 & FuzzyAlign (high) & 0.156 & 0.387 & 0.832 & 0.931 & 0.799 & 0.839 & 0.836 & 0.541 \\
 & FuzzyAlign (low) & 0.221 & 0.424 & 0.856 & 0.921 & 0.782 & 0.829 & 0.834 & 0.539 \\
 & BHA & \textbf{0.268} & \textbf{0.465} & \textbf{0.913} & \textbf{0.963} & 0.864 & \textbf{0.879} & \textbf{0.777} & \textbf{0.484} \\
\bottomrule
\end{tabular}}
\end{table}

\section{Supplementary Examples}
\label{sec:supplementary_examples}
These galleries complement the representation and reconstruction analyses
in Appendices~\ref{sec:extended_representation_analysis}
and~\ref{sec:comprehensive_reconstruction_performance}.

\subsection{Neuron Concept Associations}
\label{sec:concept_examples}
\begin{figure}[!hbp]
    \centering
    \includegraphics[width=\textwidth,height=0.68\textheight,keepaspectratio,pagebox=cropbox]{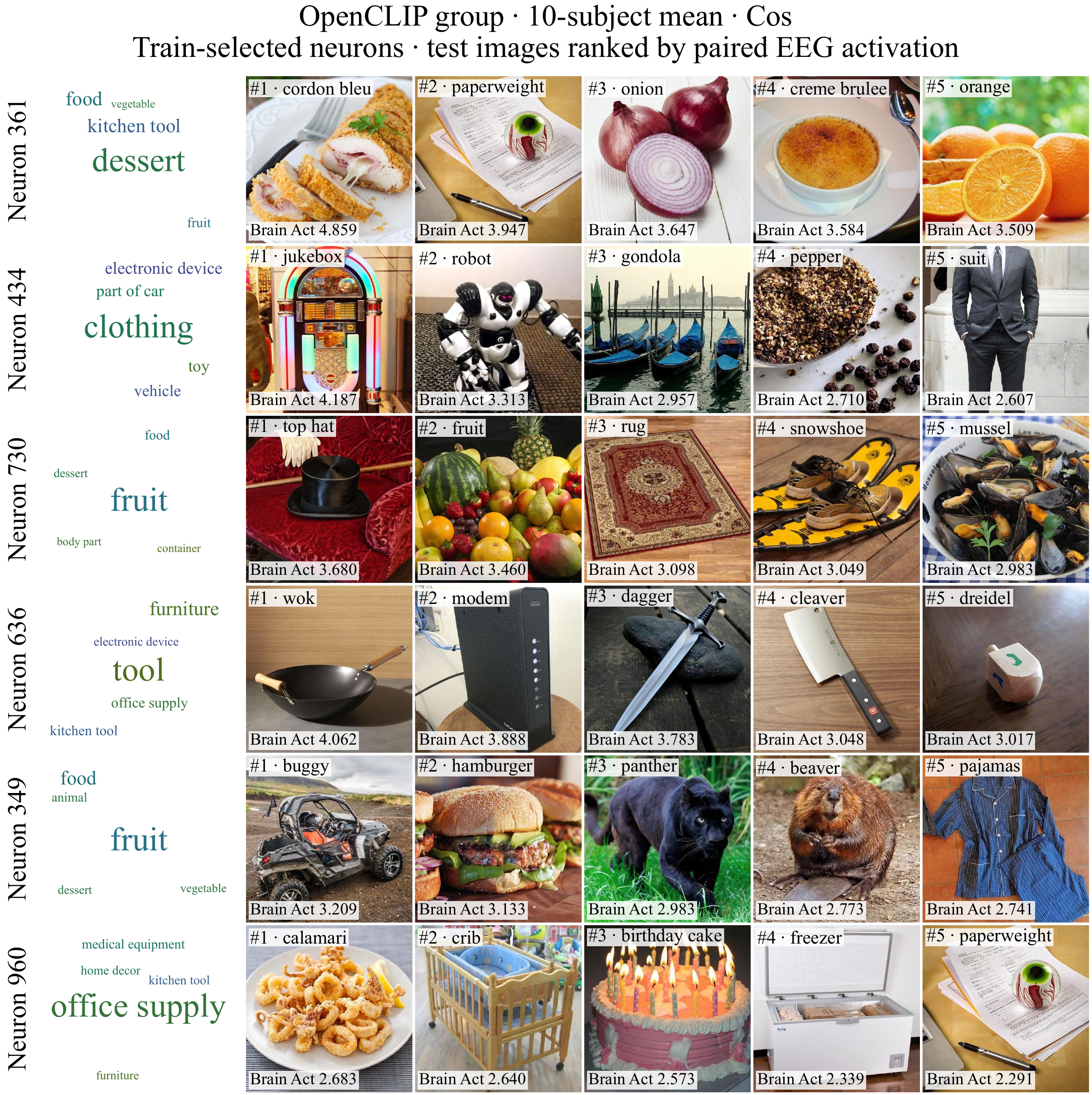}
    \caption{\textbf{OpenCLIP-aligned student expert: concept associations.}
Each row shows one of six training-selected neurons: its five highest-scoring
concepts (left) and five highest-activating test images (right).
Activation values are ten-subject means.
See Section~\ref{sec:concept_gallery_protocol} for more details.}
    \label{fig:brain_cosine_openclip}
\end{figure}

\begin{figure}[!t]
    \centering
    \includegraphics[width=\textwidth,height=0.68\textheight,keepaspectratio,pagebox=cropbox]{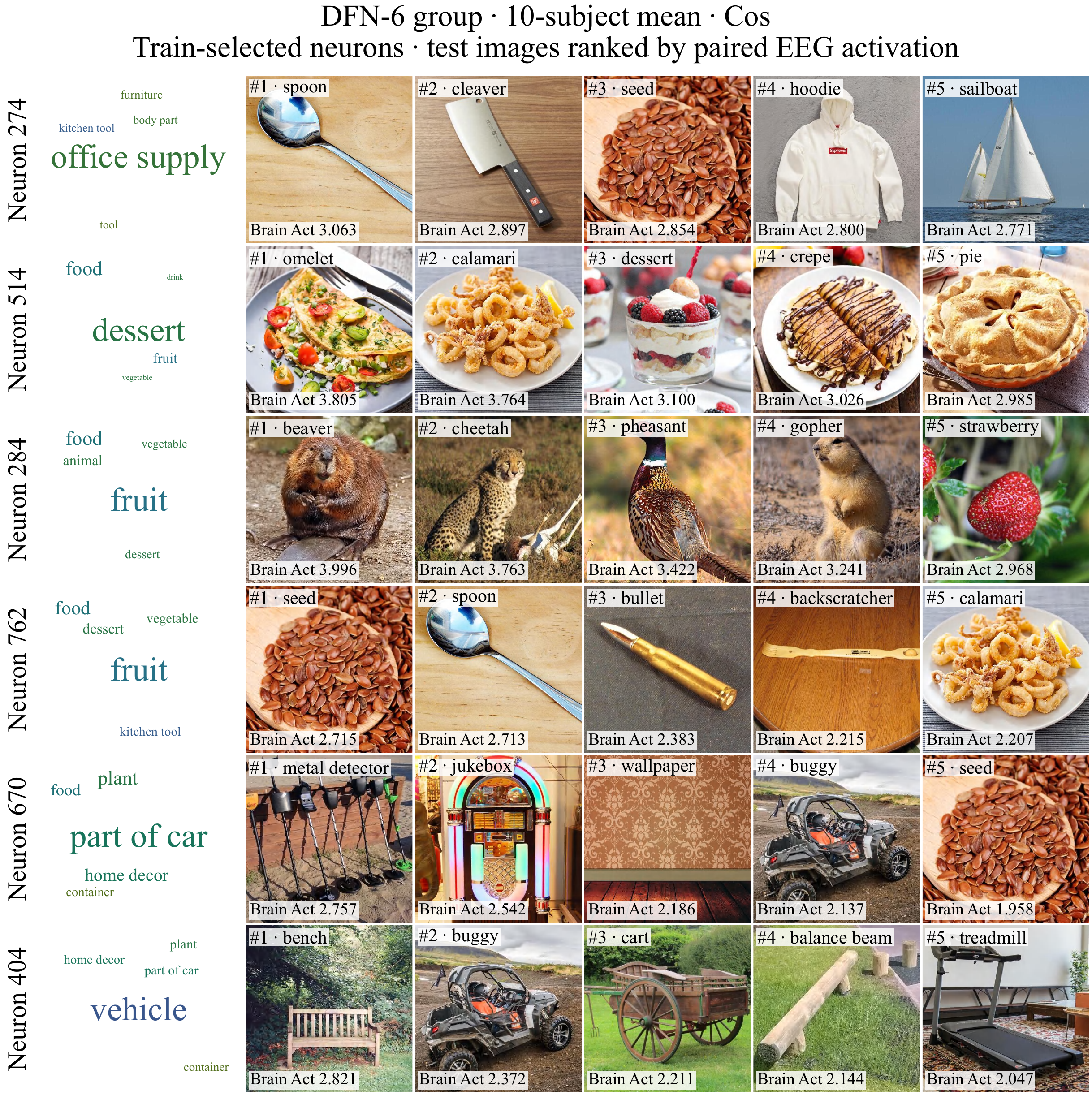}
    \caption{\textbf{DFN-6-aligned student expert: concept associations.}
Each row shows one of six training-selected neurons: its five highest-scoring
concepts (left) and five highest-activating test images (right).
Activation values are ten-subject means.
See Section~\ref{sec:concept_gallery_protocol} for more details.}
    \label{fig:brain_cosine_dfn}
\end{figure}

\begin{figure}[!t]
    \centering
    \includegraphics[width=\textwidth,height=0.68\textheight,keepaspectratio,pagebox=cropbox]{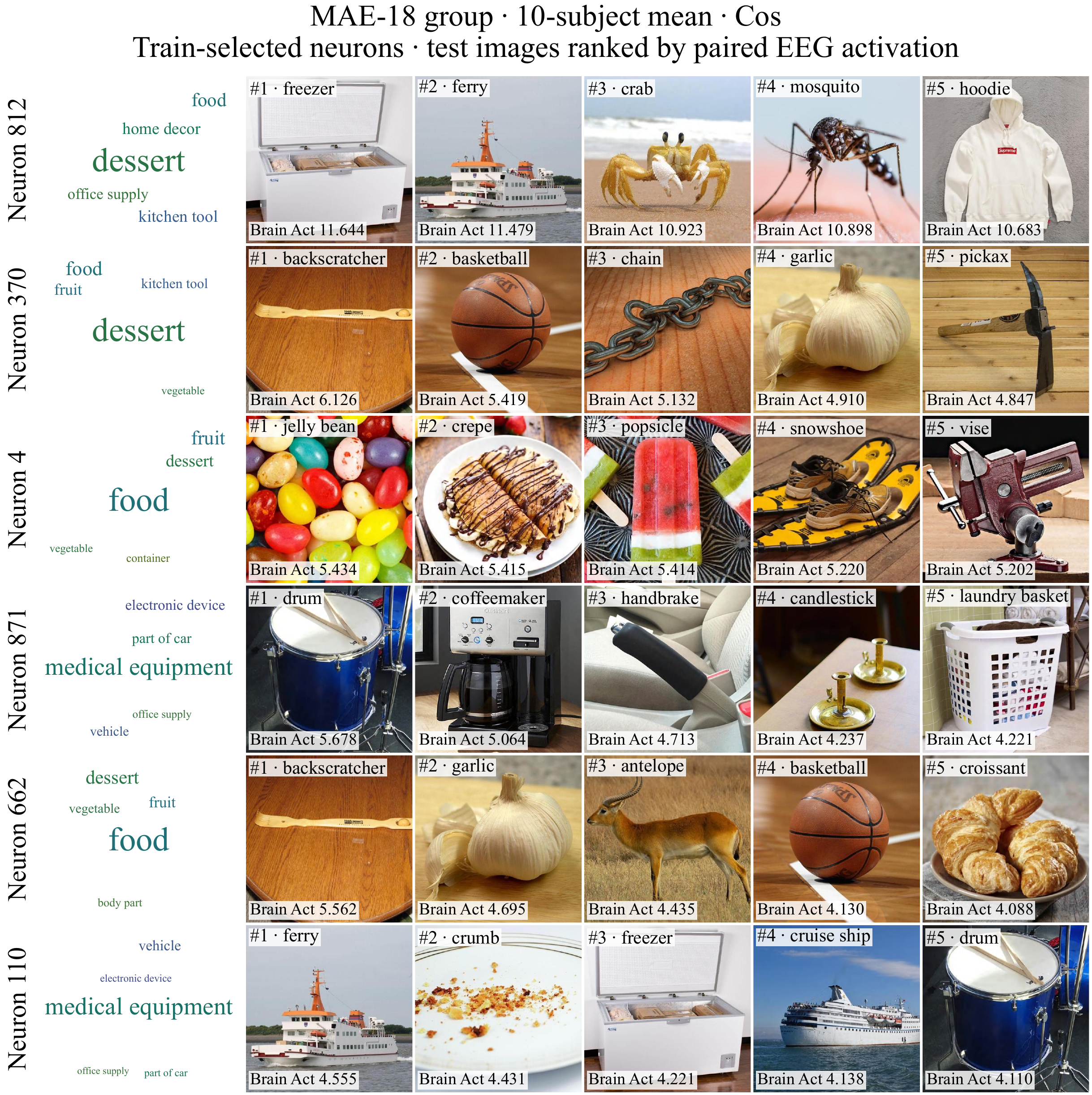}
    \caption{\textbf{MAE-18-aligned student expert: concept associations.}
Each row shows one of six training-selected neurons: its five highest-scoring
concepts (left) and five highest-activating test images (right).
Activation values are ten-subject means.
See Section~\ref{sec:concept_gallery_protocol} for more details.}
    \label{fig:brain_cosine_mae}
\end{figure}

\begin{figure}[!t]
    \centering
    \includegraphics[width=\textwidth,height=0.68\textheight,keepaspectratio,pagebox=cropbox]{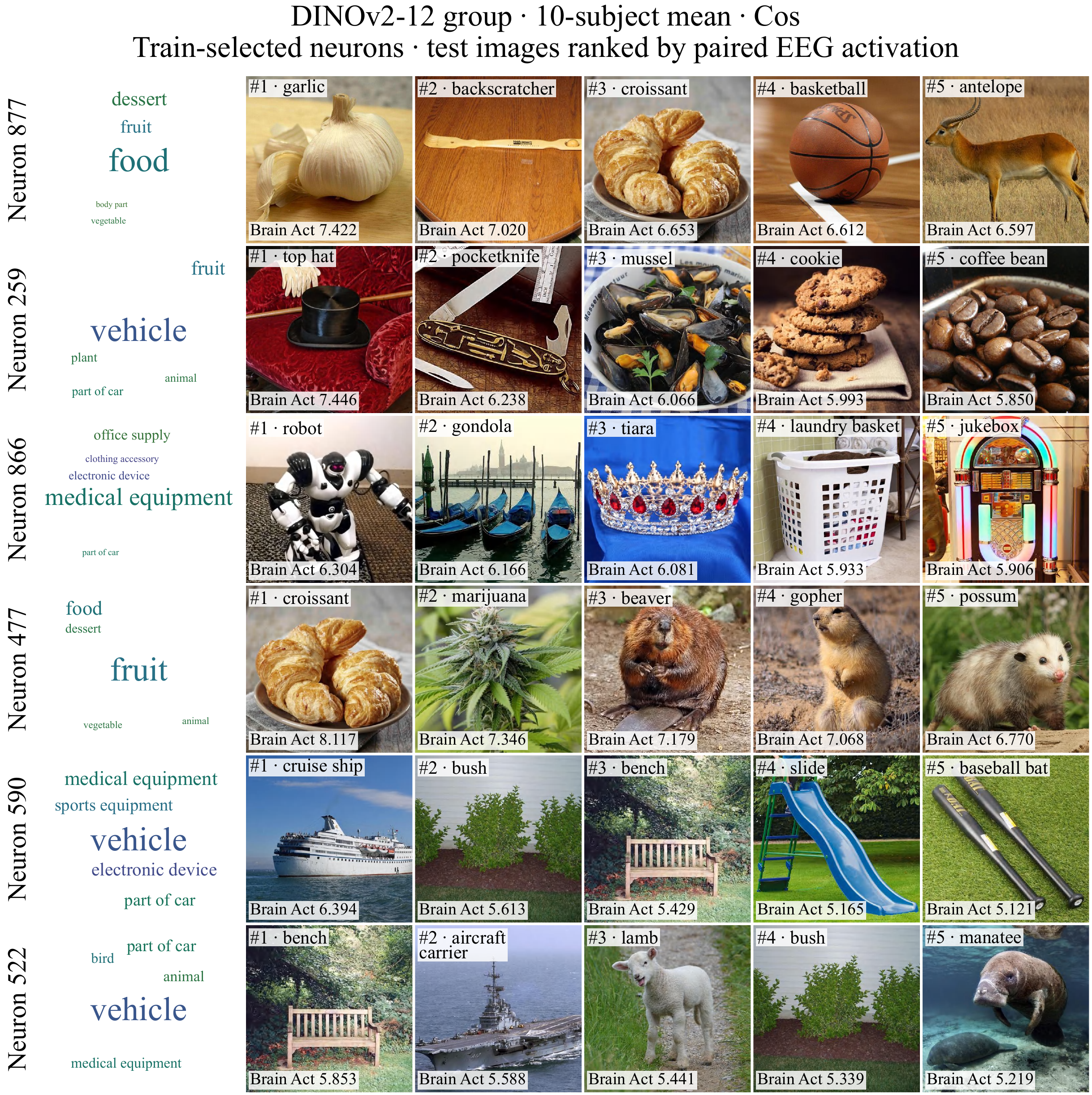}
    \caption{\textbf{DINOv2-12-aligned student expert: concept associations.}
Each row shows one of six training-selected neurons: its five highest-scoring
concepts (left) and five highest-activating test images (right).
Activation values are ten-subject means.
See Section~\ref{sec:concept_gallery_protocol} for more details.}
    \label{fig:brain_cosine_dinov2l}
\end{figure}

\begin{figure}[!t]
    \centering
    \includegraphics[width=\textwidth,height=0.68\textheight,keepaspectratio,pagebox=cropbox]{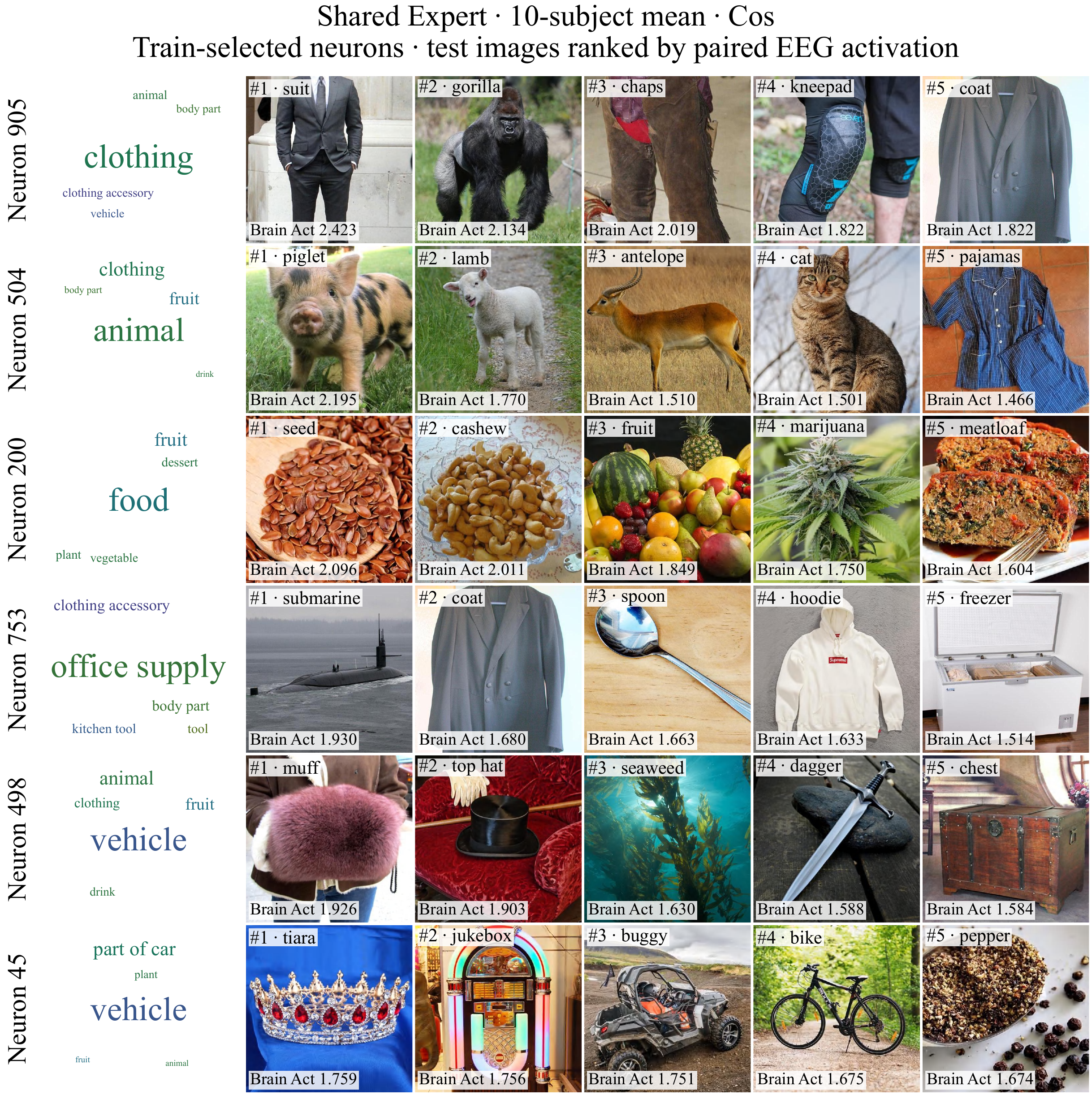}
    \caption{\textbf{Shared student expert: concept associations.}
Each row shows one of six training-selected neurons: its five highest-scoring
concepts (left) and five highest-activating test images (right).
Activation values are ten-subject means.
See Section~\ref{sec:concept_gallery_protocol} for more details.}
    \label{fig:brain_cosine_shared_fusion}
\end{figure}

\begin{figure}[!t]
    \centering
    \includegraphics[width=\textwidth,height=0.68\textheight,keepaspectratio,pagebox=cropbox]{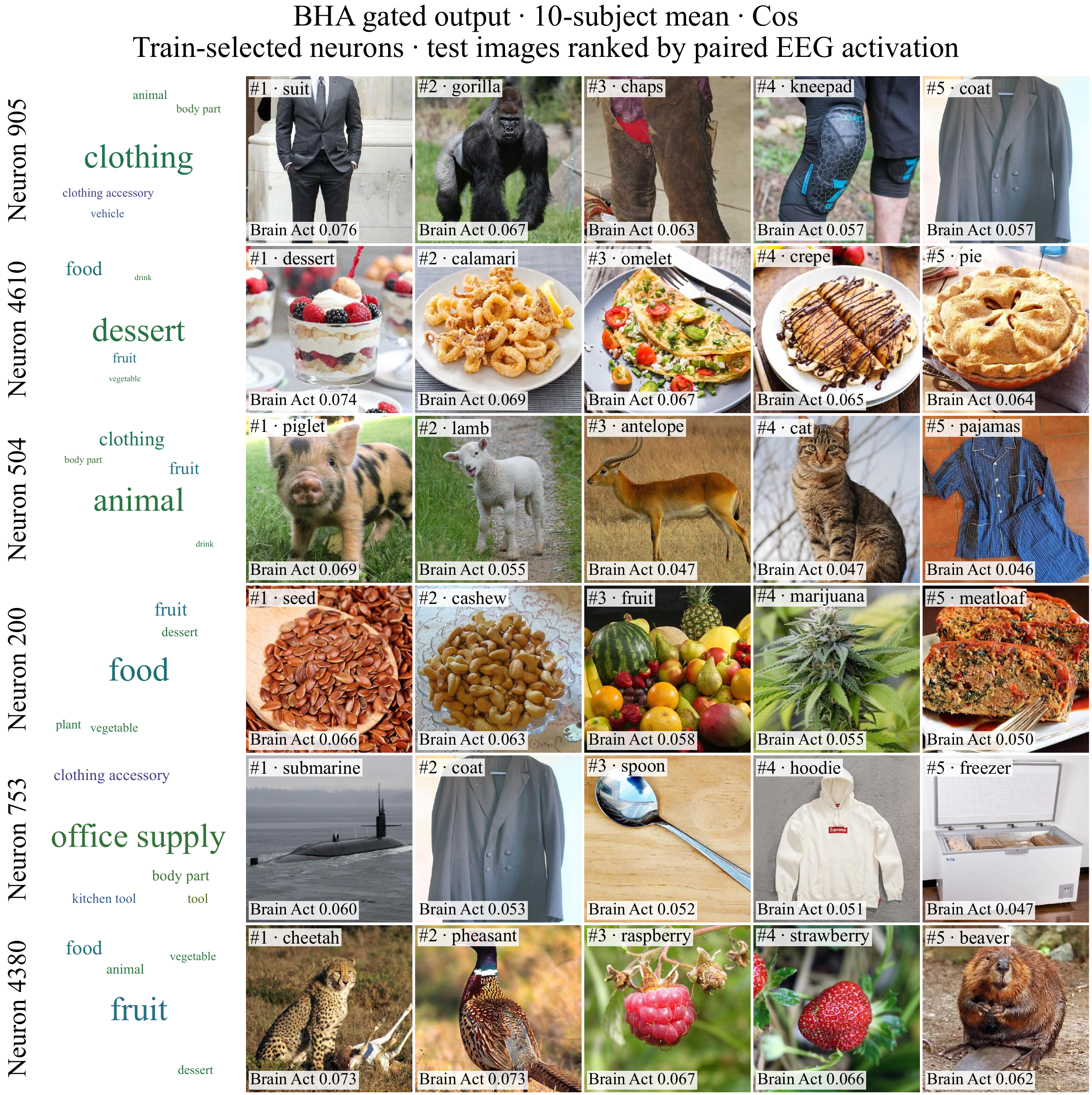}
    \caption{\textbf{BHA gated student output: concept associations.}
Each row shows one of six training-selected neurons: its five highest-scoring
concepts (left) and five highest-activating test images (right).
Activation values are ten-subject means.
See Section~\ref{sec:concept_gallery_protocol} for more details.}
    \label{fig:brain_cosine_bha_gated}
\end{figure}

Figures~\ref{fig:brain_cosine_openclip}--\ref{fig:brain_cosine_bha_gated}
show six training-selected neurons for each teacher-specific expert, the
shared expert, and the gated BHA output. Each row pairs a five-concept
word cloud with the five test images that elicit the highest mean activation
across ten subjects. Neuron and concept selection use training images only;
word sizes express relative scores within a neuron.
These pooled associations do not establish correspondence between neurons
across subjects. Section~\ref{sec:concept_gallery_protocol} specifies the
selection and pooling procedure; subject-wise quantitative scores are
reported in Section~\ref{sec:concept_score_protocol}.

\clearpage
\subsection{Feature Inversion Examples}
\label{sec:feature_inversion_examples}
\begin{figure}[!htbp]
    \centering
    \includegraphics[width=\textwidth,pagebox=cropbox]{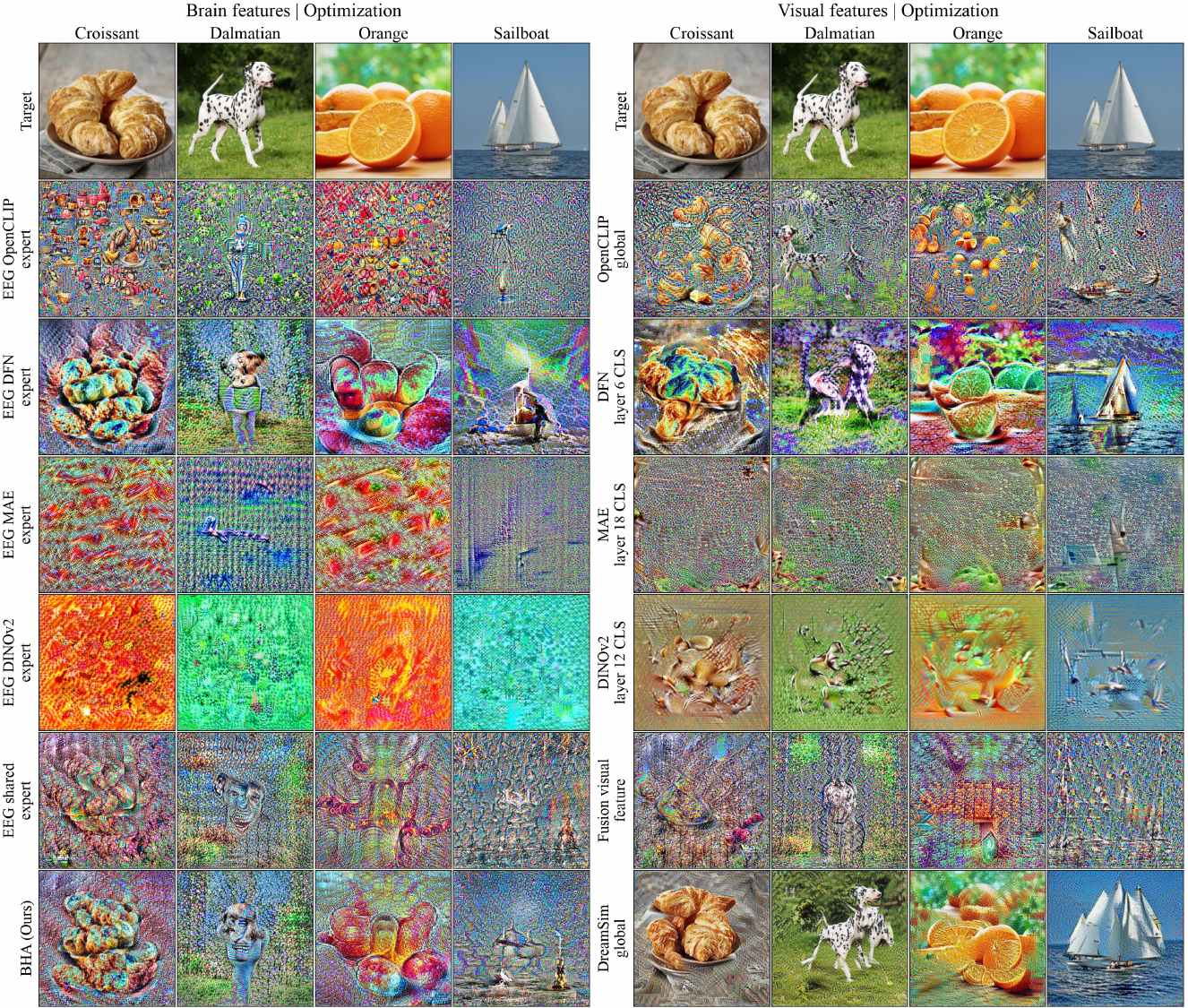}
    \caption{\textbf{Direct pixel inversion.}
Left: EEG-predicted features from subject~08 (80-trial averages before
encoding); right: GT-image features. Columns: \emph{croissant, dalmatian,
orange, and sailboat}, with GT images above. Bottom rows: gated BHA
(left) and an independent image-derived DreamSim reference (right).
See Section~\ref{sec:feature_inversion} for more details.}
    \label{fig:feature_inversion_optimization}
\end{figure}

\begin{figure}[!t]
    \centering
    \includegraphics[width=\textwidth,pagebox=cropbox]{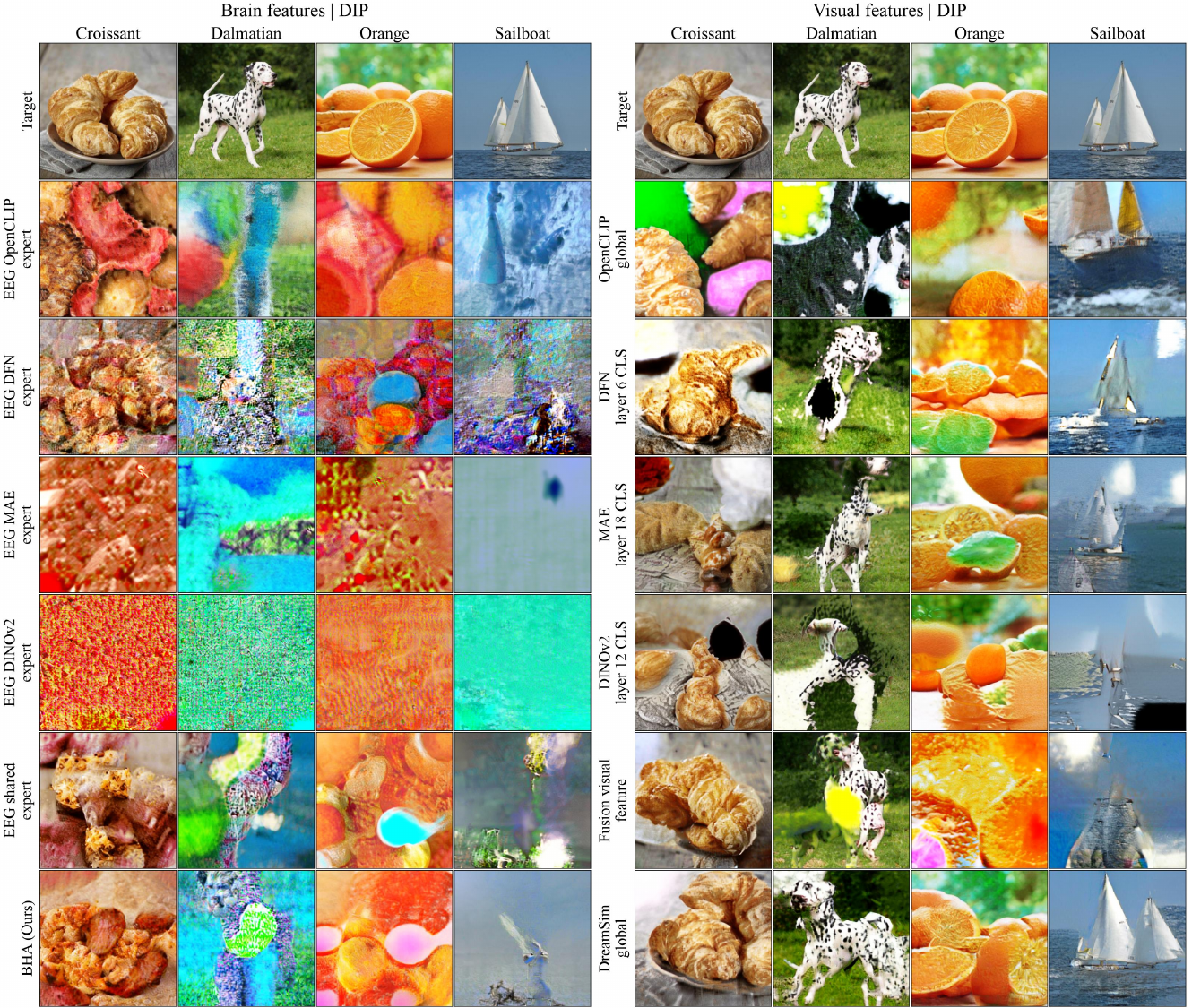}
    \caption{\textbf{Feature inversion with a deep image prior.}
A randomly initialized generator is fitted to each feature target.
Left: EEG-predicted features from subject~08 (80-trial averages before
encoding); right: GT-image features. Columns: \emph{croissant, dalmatian,
orange, and sailboat}, with GT images above. Bottom rows: gated BHA
(left) and the image-derived DreamSim reference (right).
See Section~\ref{sec:feature_inversion} for more details.}
    \label{fig:feature_inversion_dip}
\end{figure}

\begin{figure}[!t]
    \centering
    \includegraphics[width=\textwidth,pagebox=cropbox]{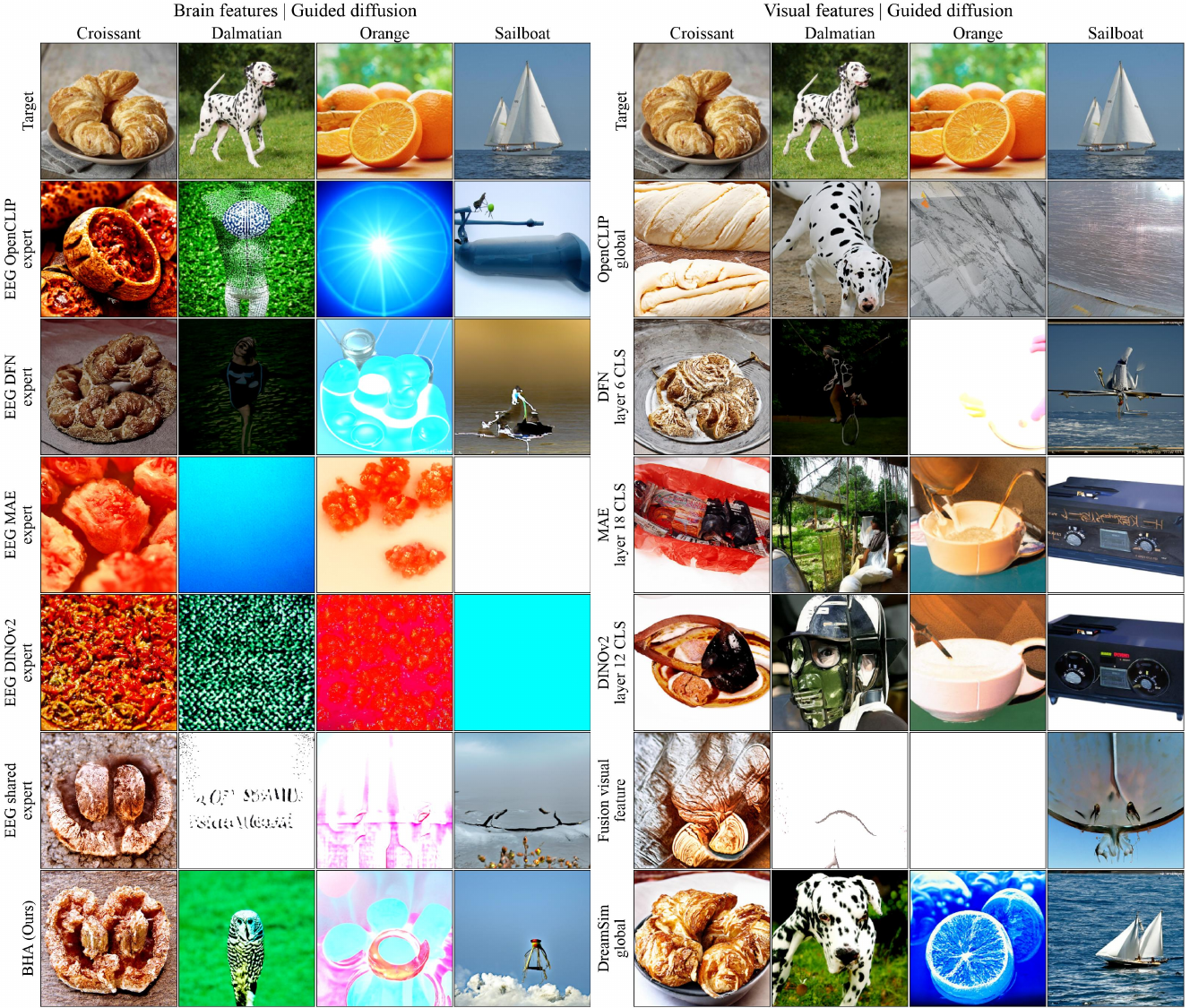}
    \caption{\textbf{Feature inversion with a pretrained diffusion prior.}
A frozen unconditional ADM supplies the image prior.
Left: EEG-predicted features from subject~08 (80-trial averages before
encoding); right: GT-image features. Columns: \emph{croissant, dalmatian,
orange, and sailboat}, with GT images above. Bottom rows: gated BHA
(left) and the image-derived DreamSim reference (right).
See Section~\ref{sec:feature_inversion} for more details.}
    \label{fig:feature_inversion_diffusion}
\end{figure}

Figures~\ref{fig:feature_inversion_optimization}--\ref{fig:feature_inversion_diffusion}
compare direct pixel optimization, a deep image prior, and guided diffusion
for the same EEG-predicted and image-derived targets. They illustrate how
the image prior affects visual plausibility and stimulus fidelity;
Section~\ref{sec:feature_inversion} defines the objectives and example selection.

\clearpage

\subsection{Reconstruction Examples}
\label{sec:add_reconst_display}
\begin{figure}[!hbp]
    \centering
    \includegraphics[width=\textwidth,height=0.72\textheight,keepaspectratio]{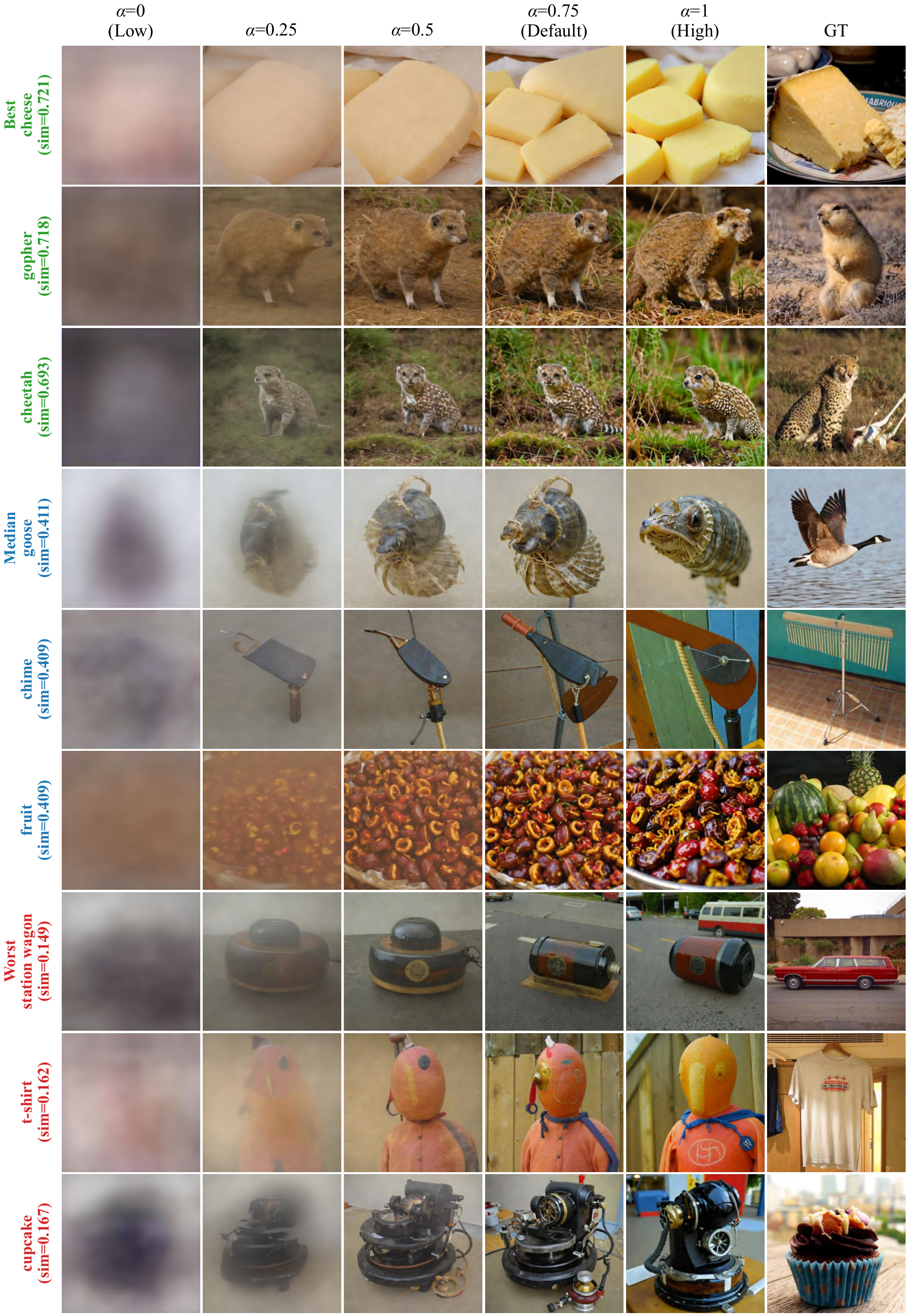}
    \caption{\textbf{Subject-01: reconstruction examples.}
Columns: $\alpha=0,0.25,0.5,0.75,1$ and target.
Rows: score-selected Best/Median/Worst examples
(Appendix~\ref{sec:add_reconst_display}).}
    \label{fig:reconst_alpha_sweep_sub01}
\end{figure}

\begin{figure}[!t]
    \centering
    \includegraphics[width=\textwidth,height=0.86\textheight,keepaspectratio]{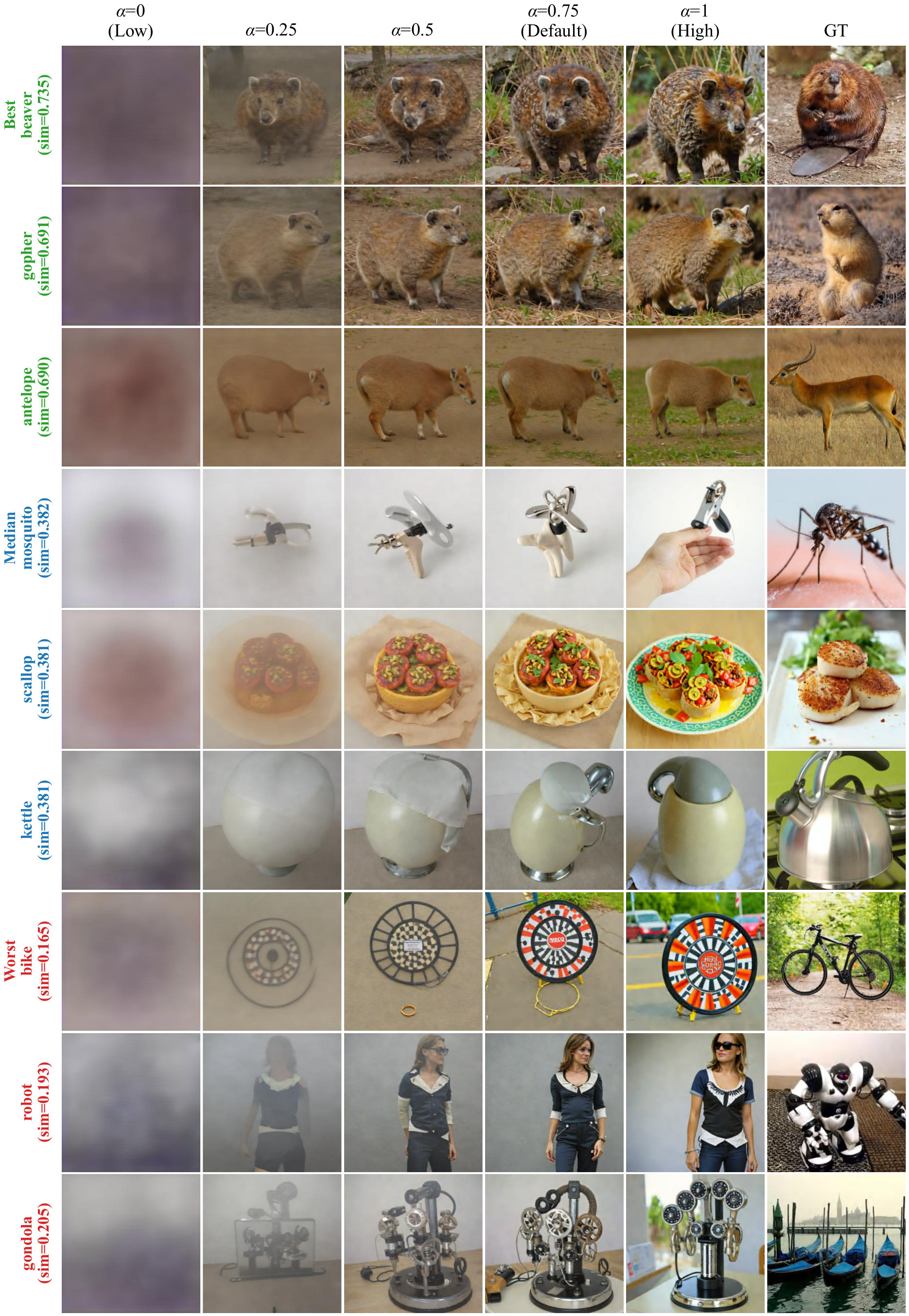}
    \caption{\textbf{Subject-02: reconstruction examples.}
Columns: $\alpha=0,0.25,0.5,0.75,1$ and target.
Rows: score-selected Best/Median/Worst examples
(Appendix~\ref{sec:add_reconst_display}).}
    \label{fig:reconst_alpha_sweep_sub02}
\end{figure}

\begin{figure}[!t]
    \centering
    \includegraphics[width=\textwidth,height=0.86\textheight,keepaspectratio]{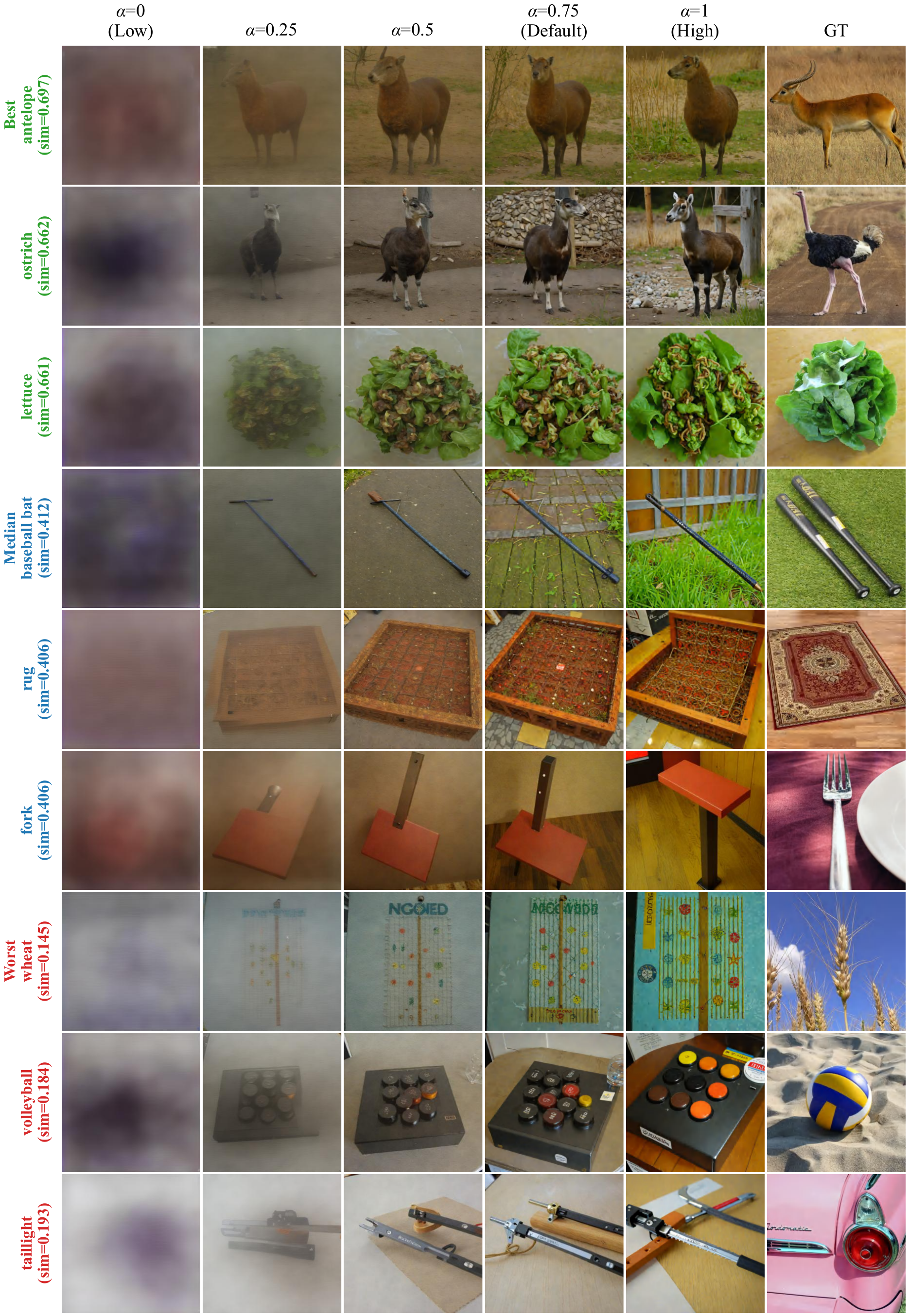}
    \caption{\textbf{Subject-03: reconstruction examples.}
Columns: $\alpha=0,0.25,0.5,0.75,1$ and target.
Rows: score-selected Best/Median/Worst examples
(Appendix~\ref{sec:add_reconst_display}).}
    \label{fig:reconst_alpha_sweep_sub03}
\end{figure}

\begin{figure}[!t]
    \centering
    \includegraphics[width=\textwidth,height=0.86\textheight,keepaspectratio]{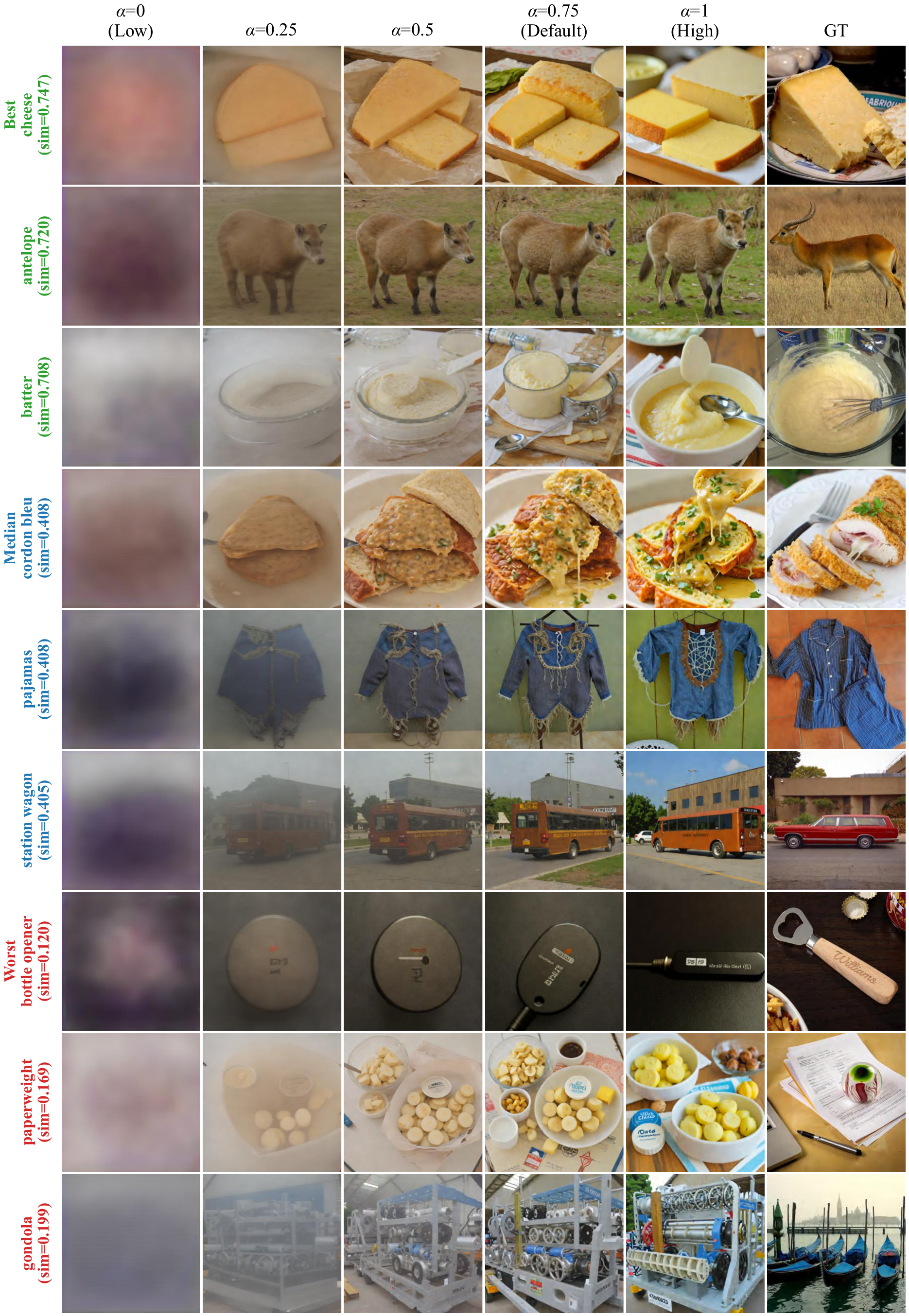}
    \caption{\textbf{Subject-04: reconstruction examples.}
Columns: $\alpha=0,0.25,0.5,0.75,1$ and target.
Rows: score-selected Best/Median/Worst examples
(Appendix~\ref{sec:add_reconst_display}).}
    \label{fig:reconst_alpha_sweep_sub04}
\end{figure}

\begin{figure}[!t]
    \centering
    \includegraphics[width=\textwidth,height=0.86\textheight,keepaspectratio]{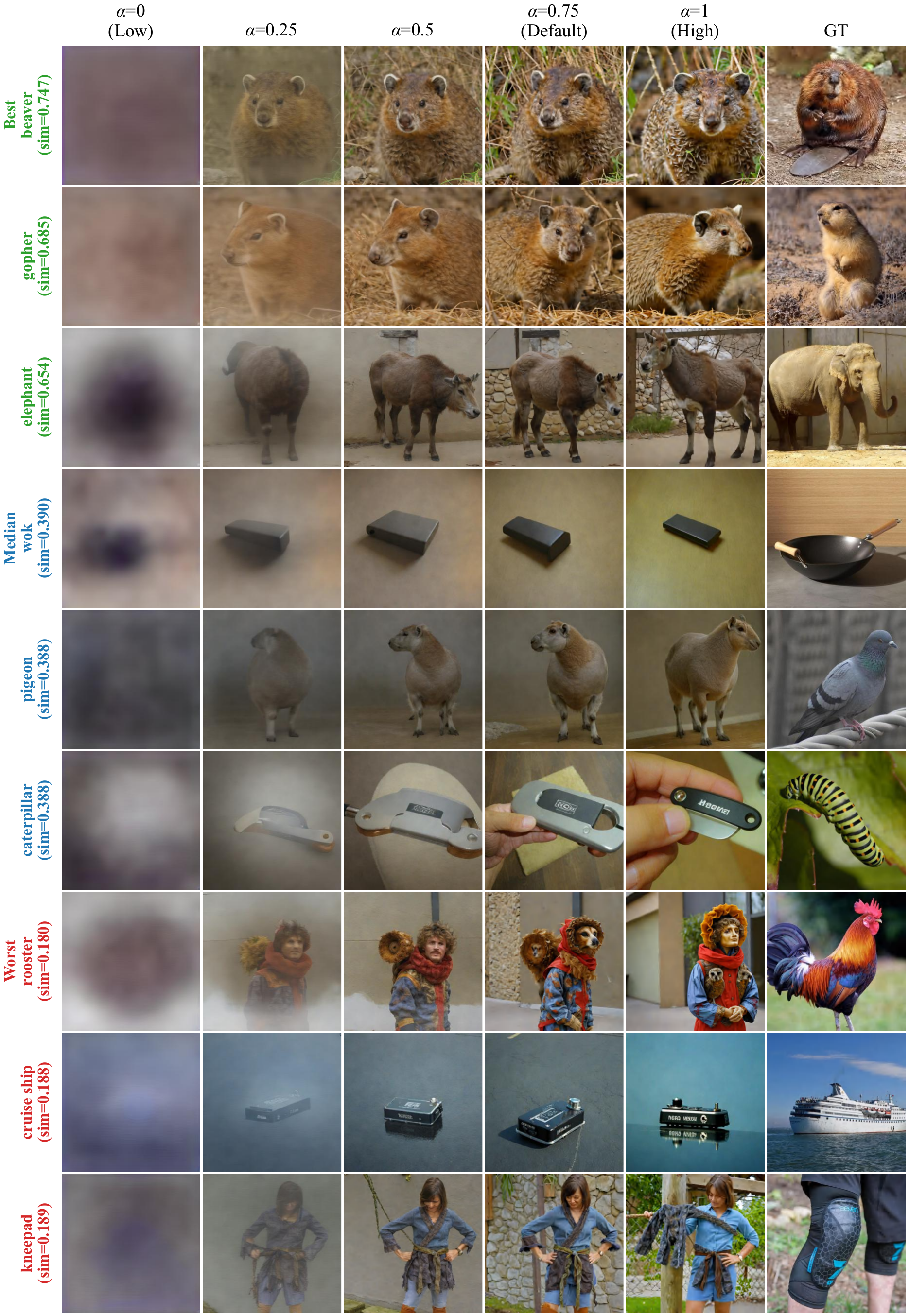}
    \caption{\textbf{Subject-05: reconstruction examples.}
Columns: $\alpha=0,0.25,0.5,0.75,1$ and target.
Rows: score-selected Best/Median/Worst examples
(Appendix~\ref{sec:add_reconst_display}).}
    \label{fig:reconst_alpha_sweep_sub05}
\end{figure}

\begin{figure}[!t]
    \centering
    \includegraphics[width=\textwidth,height=0.86\textheight,keepaspectratio]{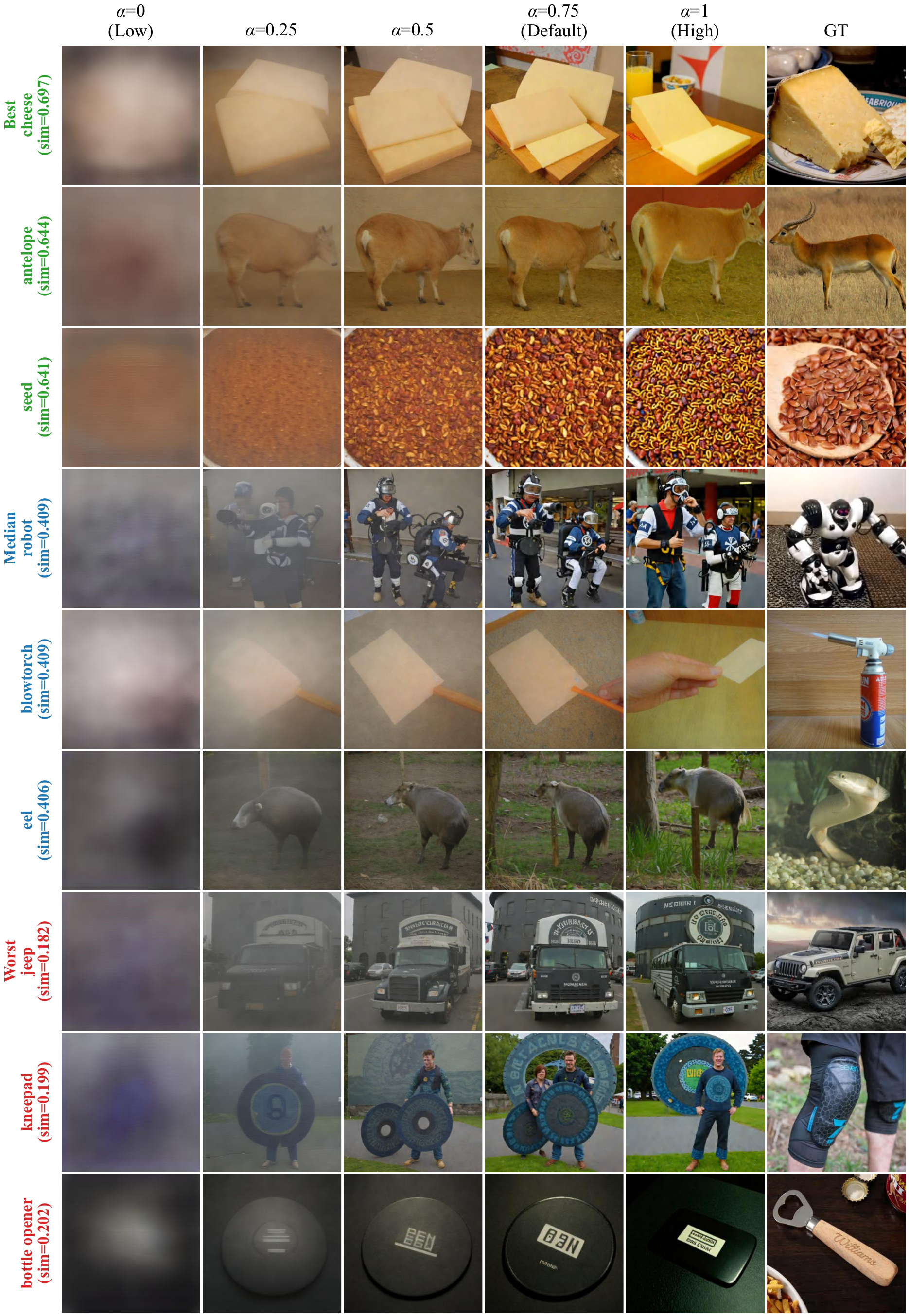}
    \caption{\textbf{Subject-06: reconstruction examples.}
Columns: $\alpha=0,0.25,0.5,0.75,1$ and target.
Rows: score-selected Best/Median/Worst examples
(Appendix~\ref{sec:add_reconst_display}).}
    \label{fig:reconst_alpha_sweep_sub06}
\end{figure}

\begin{figure}[!t]
    \centering
    \includegraphics[width=\textwidth,height=0.86\textheight,keepaspectratio]{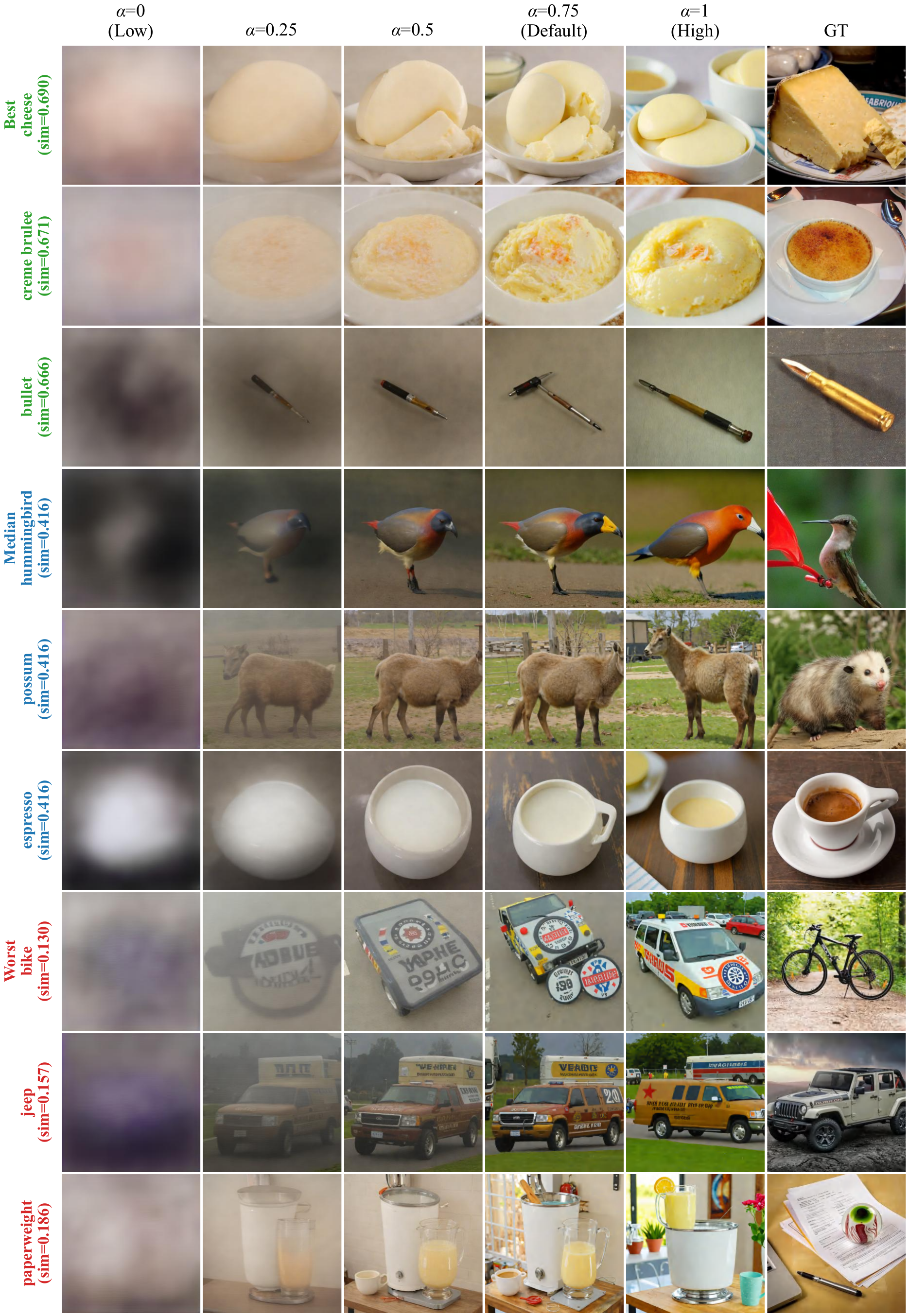}
    \caption{\textbf{Subject-07: reconstruction examples.}
Columns: $\alpha=0,0.25,0.5,0.75,1$ and target.
Rows: score-selected Best/Median/Worst examples
(Appendix~\ref{sec:add_reconst_display}).}
    \label{fig:reconst_alpha_sweep_sub07}
\end{figure}

\begin{figure}[!t]
    \centering
    \includegraphics[width=\textwidth,height=0.86\textheight,keepaspectratio]{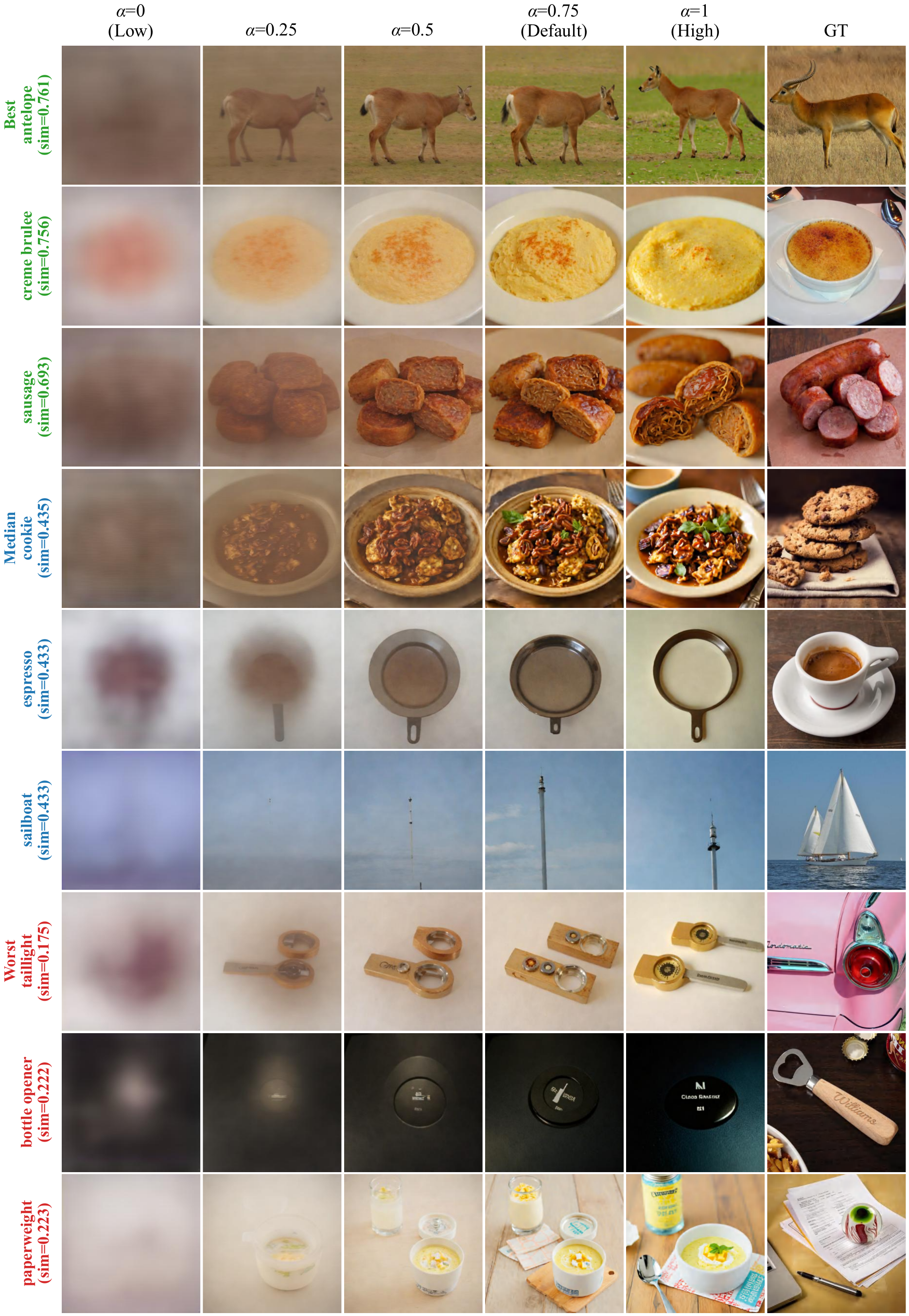}
    \caption{\textbf{Subject-08: reconstruction examples.}
Columns: $\alpha=0,0.25,0.5,0.75,1$ and target.
Rows: score-selected Best/Median/Worst examples
(Appendix~\ref{sec:add_reconst_display}).}
    \label{fig:reconst_alpha_sweep_sub08}
\end{figure}

\begin{figure}[!t]
    \centering
    \includegraphics[width=\textwidth,height=0.86\textheight,keepaspectratio]{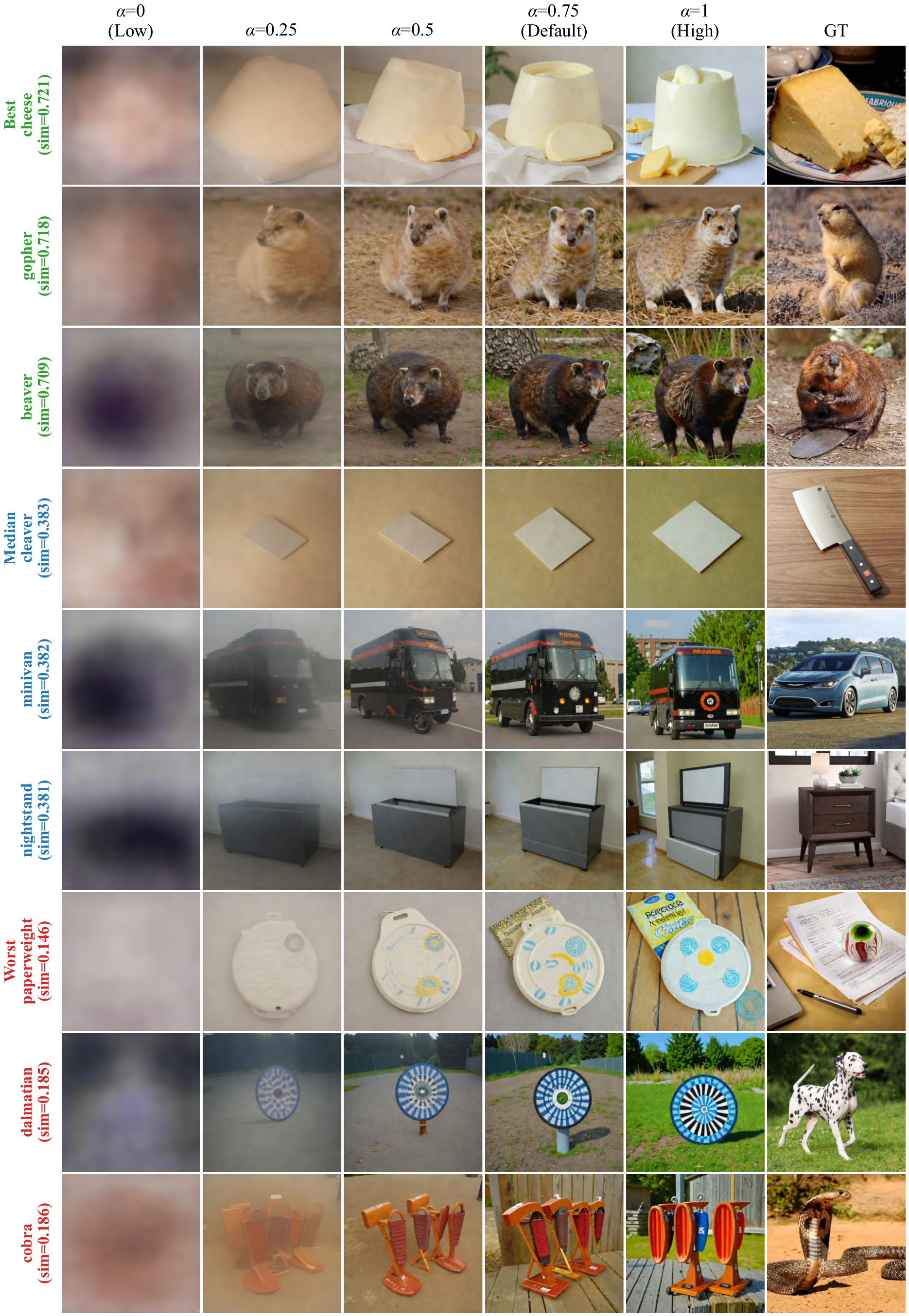}
    \caption{\textbf{Subject-09: reconstruction examples.}
Columns: $\alpha=0,0.25,0.5,0.75,1$ and target.
Rows: score-selected Best/Median/Worst examples
(Appendix~\ref{sec:add_reconst_display}).}
    \label{fig:reconst_alpha_sweep_sub09}
\end{figure}

\begin{figure}[!t]
    \centering
    \includegraphics[width=\textwidth,height=0.86\textheight,keepaspectratio]{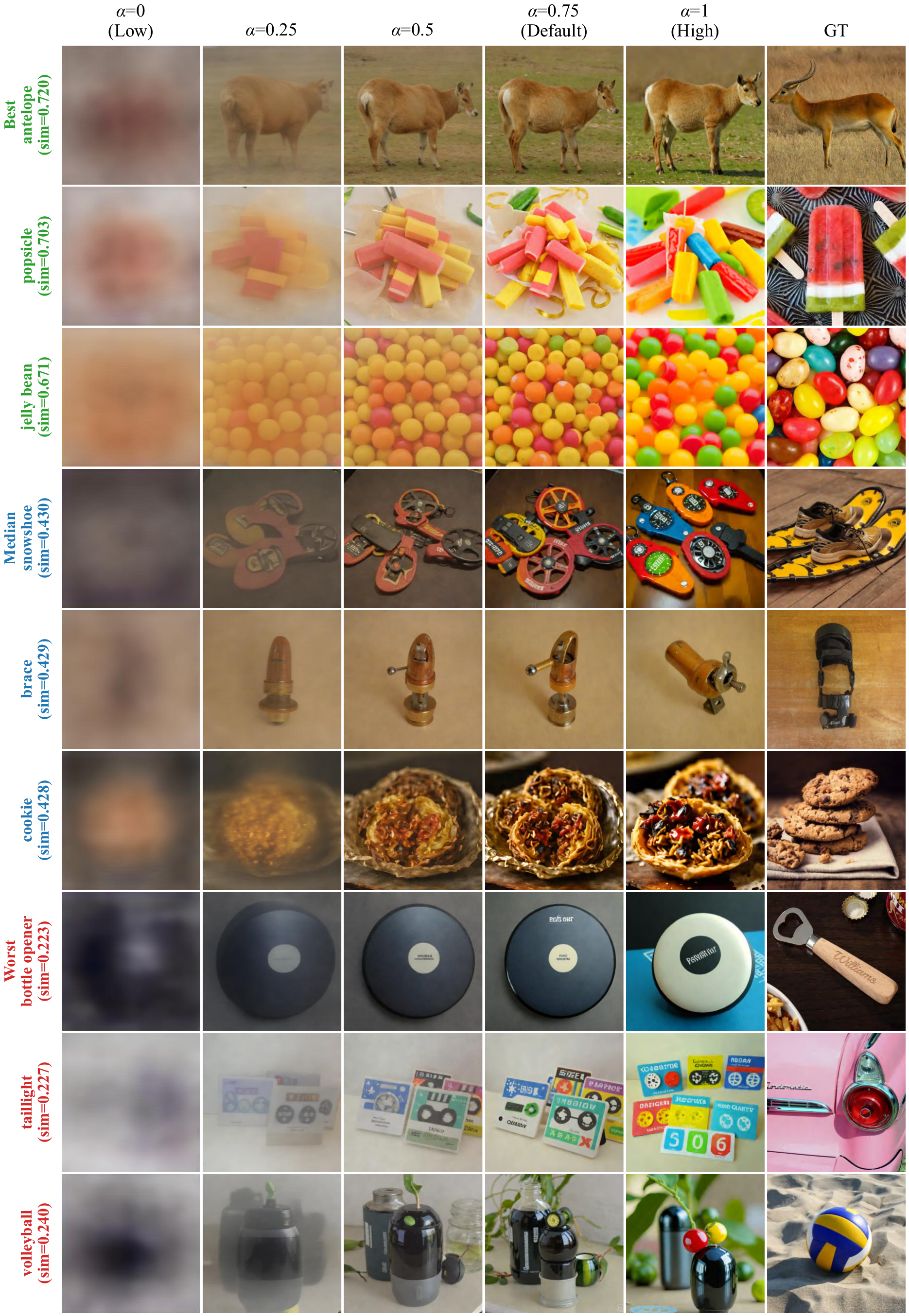}
    \caption{\textbf{Subject-10: reconstruction examples.}
Columns: $\alpha=0,0.25,0.5,0.75,1$ and target.
Rows: score-selected Best/Median/Worst examples
(Appendix~\ref{sec:add_reconst_display}).}
    \label{fig:reconst_alpha_sweep_sub10}
\end{figure}

Figures~\ref{fig:reconst_alpha_sweep_sub01}--\ref{fig:reconst_alpha_sweep_sub10}
show reconstructions across mixing strengths for all ten subjects.
Columns correspond to $\alpha\in\{0,0.25,0.5,0.75,1\}$ and the target;
$\alpha=0.75$ is the default.

For each subject, we rank 200 test images by paired OpenCLIP ViT-H/14
cosine similarity, averaged equally across the five strengths using
generation~0. The top 30\%, middle 40\%, and bottom 30\% define the
Best, Median, and Worst partitions. We display the three highest-scoring
images, the three middle-partition images closest to the overall median,
and the three lowest-scoring images. This selection illustrates variation
in reconstruction fidelity rather than estimating average test performance.

\end{document}